\documentclass[11pt]{article}

\usepackage[preprint]{acl}

\usepackage{times}
\usepackage{latexsym}

\usepackage[T1]{fontenc}

\usepackage[utf8]{inputenc}

\usepackage{microtype}

\usepackage{inconsolata}

\usepackage{graphicx}

\usepackage[utf8]{inputenc}
\usepackage[T1]{fontenc}
\usepackage{microtype}
\usepackage{hyperref}
\usepackage{url}
\usepackage{booktabs}
\usepackage{amsfonts}
\usepackage{nicefrac}

\usepackage{pdflscape}

\usepackage{float}
\usepackage{graphicx}
\usepackage{pgfplots}
\pgfplotsset{compat=1.18}

\usepackage{subcaption}
\usepackage{array}
\usepackage{makecell}
\usepackage{adjustbox}
\usepackage{pifont}
\usepackage{lineno}
\usepackage{amsmath}
\usepackage{xspace}
\usepackage{longtable}
\usepackage{multirow}

\definecolor{reactgreen}{HTML}{B8D98A}
\definecolor{cotblue}{HTML}{89D0F2}
\definecolor{psorange}{HTML}{F6C58F}
\definecolor{madblue}{HTML}{8CAEE5}
\definecolor{refpink}{HTML}{E8B3CF}
\definecolor{directpink}{HTML}{F4A3A3}
\definecolor{darkblue}{rgb}{0, 0, 0.5}

\usepackage{cuted}
\usepackage{capt-of}
\usepackage{tabularx}  
\usepackage{booktabs}

\newcommand{\cmark}{\textcolor{green!55!black}{\ding{51}}}
\newcommand{\xmark}{\textcolor{red!70!black}{\ding{55}}}

\newcommand{\ours}[0]{\textsc{AgentSpec}\xspace}

\title{AgentSpec: Understanding Embodied Agent Scaffolds Through Controlled Composition}

\author{
\begin{tabular}{@{}c@{}}
{\small\bfseries
Jixuan Chen\textsuperscript{1}\quad
Jianzhi Shen\textsuperscript{2}\quad
Haoqiang Kang\textsuperscript{1}\quad
Zhi Hong\textsuperscript{1}\quad
Qingyi Jiang\textsuperscript{1}\quad
Soham Bose\textsuperscript{1}
}
\\[-0.1em]
{\small\bfseries
Yiming Zhang\textsuperscript{1}\quad
Leon Leng\textsuperscript{3}\quad
Amit Vyas\textsuperscript{1}\quad
Lingjun Mao\textsuperscript{1}\quad
Siru Ouyang\textsuperscript{4}\quad
Kun Zhou\textsuperscript{1}\quad
Lianhui Qin\textsuperscript{1}
}
\\[0.45em]
{\normalfont\small
\textsuperscript{1}University of California, San Diego\quad
\textsuperscript{2}Johns Hopkins University
}
\\[-0.1em]
{\normalfont\small
\textsuperscript{3}University of Washington\quad
\textsuperscript{4}University of Illinois Urbana-Champaign
}
\end{tabular}
}

\begin{document}

\maketitle

\begin{abstract}
LLM agents are increasingly built not as single model calls, but as scaffolded systems that combine reasoning, memory, reflection, action execution, and learning. While such scaffolds often improve performance, they are often embedded in tightly coupled pipelines, making it difficult to isolate component contributions, compare alternative designs, or understand how module interactions shape agent behavior. We introduce \textsc{AgentSpec}, a modular specification framework that represents embodied agents as typed compositions of reusable policy components with standardized interfaces. \textsc{AgentSpec} standardizes the interfaces among perception, memory, reasoning, reflection, action, and optional learning, enabling components to be swapped and recombined under controlled conditions. We instantiate this framework across DeliveryBench, ALFRED, MiniGrid, and RoboTHOR, and analyze reasoning, memory, reflection, and reinforcement-learning modules across model backbones. Our results show that agent performance is governed by scaffold compatibility and interaction effects rather than isolated module strength. In particular, structured multi-granularity memory improves long-horizon state tracking, reasoning and memory interact non-uniformly across environments, reflection trades off correction and cost, and RL-trained policies compose best when optimized with deployment-time scaffold structure. \textsc{AgentSpec} provides a controlled foundation for studying, comparing, and designing composable LLM agents. Our code, baselines and interactive playground are publicly available at \url{https://agentspec-embodied.github.io}.

\end{abstract}
\section{Introduction}

\begin{figure*}[t]
\vspace{-2pt}
    \begin{center}
    \includegraphics[width=\textwidth, trim=0 100 0 0, clip]{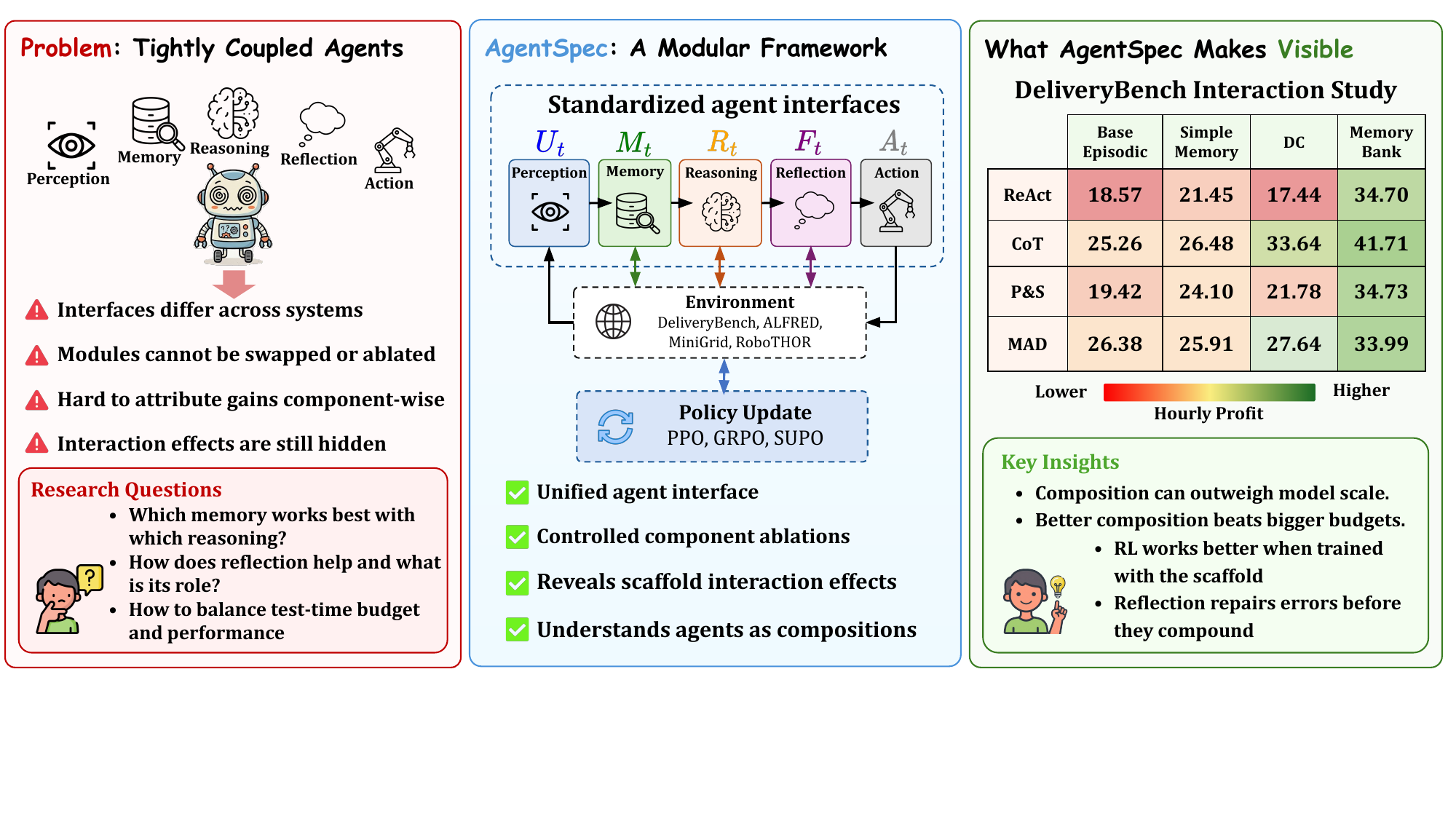}
    \end{center}

    \caption{\ours turns tightly coupled embodied-agent pipelines into a controlled modular design space with fixed typed interfaces, enabling systematic module composition and revealing interaction effects between reasoning, memory, reflection, action, and learning.}
    \label{fig:teaser}
    \vspace{-0.5cm}
\end{figure*}

Recent advances in large language models (LLMs) have substantially improved end-to-end reasoning capability. However, solving complex real-world tasks as agents, especially long-horizon decision-making in embodied environments, requires more than stronger next-step prediction~\citep{ahn2022can,huang2022inner,wang2023voyager}. Success depends on aligning perception, memory, reasoning, and action across many rounds of interaction. Recent agent frameworks such as OpenClaw\footnote{\url{https://github.com/openclaw/openclaw}} illustrate this shift: rather than relying on a single model invocation, they augment a base LLM (e.g., GPT-5) with tool execution, state tracking, and persistent memory. Their capability therefore comes not from the model alone, but from composing these components into a coherent decision-making system.

Yet despite their growing capability, most agent systems remain tightly coupled pipelines. Recent modular agent frameworks and cognitive architectures, such as CoALA~\citep{sumers2023cognitive}, AgentSquare~\citep{shang2024agentsquare}, AgentGym~\citep{xi2025agentgym}, Voyager~\citep{wang2023voyager}, and OpenClaw, expose reasoning, memory, tool use, and action execution as reusable building blocks. However, they are typically designed as complete systems or optimized for high-performing configurations, rather than controlled platforms for attributing component-level and interaction-level effects. Consequently, when reasoning, memory, reflection, and reinforcement learning are intertwined, improvements remain difficult to isolate and generalize. The field still lacks principled answers to basic design questions: which reasoning strategies help in which settings; when memory is useful and what form it should take; when reinforcement learning composes well with reasoning strategies; and when reflection improves decisions rather than merely increasing cost.

We address this gap with \ours, a modular framework that makes agent composition explicit. It represents an agent as a Perception--Memory--Reasoning--Reflection--Action loop, with reinforcement learning as an optional module for further optimizing behavior. Perception converts raw observations into a standardized state representation; memory retrieves relevant history and knowledge; reasoning proposes a decision; reflection critiques or revises it; and action executes it in the environment. By standardizing interfaces, \ours turns many existing agent designs~\citep{packer2023memgpt,park2023generative,li2025metaagents} into special cases within a shared design space, allowing components to be swapped, recombined, and studied without rebuilding the full system. This enables stronger agents and clearer scientific analysis, but requires evaluation settings where module interactions are observable rather than hidden inside one-shot outputs.

We use embodied agents as a diagnostic setting for modular agent design because embodied tasks are closed-loop: each action changes the agent's future observations, available choices, and accumulated history. Performance therefore depends not only on individual module quality, but also on compatibility within the full decision loop. For instance, detailed trajectory memory may help long-horizon state tracking, but distract a planning-oriented reasoner if the retrieved context is too low-level; conversely, strong reasoning may still fail when memory does not preserve the task state needed for later decisions. These interaction effects are central to the questions \ours is designed to study.

We evaluate \ours across four embodied benchmarks that stress complementary aspects of modular decision-making: DeliveryBench~\citep{mao2025deliverybench} emphasizes long-horizon planning under resource and deadline constraints; ALFRED~\citep{shridhar2020alfred} requires compositional household manipulation and persistent task-state tracking; MiniGrid~\citep{chevalier2023minigrid} isolates symbolic navigation and partial observability; and RoboTHOR~\citep{deitke2020robothor} tests first-person navigation in realistic 3D scenes. Together, they vary horizon length, observation modality, realism, and control difficulty, allowing us to study when module combinations help, when they hurt, and which design principles transfer across settings.

Our experiments reveal three general principles. First, module compatibility matters as much as module strength: reasoning structures local decisions, while memory preserves task state across long horizons, but memory helps only when its representation matches the reasoning strategy. Second, the best composition is environment-dependent. Shorter or more symbolic tasks rely more on per-step reasoning, whereas long-horizon embodied tasks are bottlenecked by state tracking and trajectory coherence. Third, effectiveness must be evaluated together with efficiency: stronger performance does not simply come from more tokens or deeper deliberation, and lightweight but well-matched compositions often achieve better performance--cost trade-offs than heavier misaligned ones.

Overall, these findings suggest that modular agent design should be treated as a structured and analyzable design space rather than a collection of interchangeable heuristics. \ours provides a controlled framework for composing reasoning, memory, reflection, and learning modules under shared interfaces, enabling systematic comparisons across backbones, tasks, and efficiency constraints. Beyond improving benchmark performance, our results expose reusable design principles: modules should be selected based on task horizon, state-tracking demands, representation compatibility, and inference cost. This also highlights an important future direction: instead of attaching reasoning or memory only at inference time, modular components may need to be jointly optimized with the policy so learned agents remain compatible with their deployment-time scaffolds.

In general, our contributions are threefold. First, we introduce \ours, a typed modular specification for embodied LLM agents that separates perception, memory, reasoning, reflection, action execution, and optional learning into interchangeable components with shared interfaces. Second, we instantiate this specification across four embodied benchmarks and multiple model backbones, enabling controlled comparisons of module choices that are usually entangled inside complete agent pipelines. Third, we use this controlled design space to identify reusable principles for scaffolded agent design, showing that memory is useful only when its representation matches the downstream reasoner, multi-granularity memory is a robust default for long-horizon tasks, reflection is most valuable when it repairs local execution errors, and RL-trained policies should be optimized together with the scaffolds they will use at deployment time.

\section{Related Work}

\textbf{LLM-Based Agent Systems.}
Modern LLM agents are often built as multi-step pipelines that integrate reasoning, memory, tool use, reflection, and action execution~\citep{park2023generative,hong2023metagpt,chen2023agentverse,wu2024autogen,li2023camel}. Cognitive-inspired frameworks such as CoALA~\citep{sumers2023cognitive} formalize agents as compositions of functional modules, while systems such as Voyager and AgentGym show that agents can accumulate skills or improve across environments~\citep{wang2023voyager,xi2025agentgym,lin2025seagent,huang2025cascade}. However, most existing systems are proposed as complete end-to-end designs, with reasoning, memory, perception, and action components tightly coupled to task-specific prompts, control logic, or environment interfaces, especially in long-horizon embodied settings~\citep{deitke2020robothor,mao2025deliverybench}. This makes it difficult to isolate components, replace them with alternatives, or systematically study how module interactions affect performance. In contrast, \ours treats agents as explicit compositions of reusable policy components with standardized interfaces, enabling controlled replacement, recombination, and analysis.

\textbf{Agent Design Space.}
Prior work has explored a wide range of agent components, including reasoning strategies such as chain-of-thought prompting~\citep{kojima2022large,wang2022self}, search-based planning~\citep{yao2023tree,zhou2023language}, and self-correction~\citep{madaan2023selfrefine,shinn2023reflexion,kumar2024training}, as well as memory mechanisms such as flat buffers~\citep{zhong2024memorybank}, tiered stores~\citep{packer2023memgpt,chhikara2025mem0}, graph or hierarchical memories~\citep{li2025cam,rasmussen2025zep,anokhin2024arigraph,zhang2025g}, retrieval-augmented memories~\citep{qian2025memorag,fang2025lightmem,liu2026simplemem}, and procedural or self-organizing memories~\citep{wang2024agent_workflow,zheng2023synapse,hu2026evermemos,nan2025nemori}. Recent frameworks further automate architecture search~\citep{hu2024automated,zhang2024aflow,li2026agentswift}, and AgentSquare~\citep{shang2024agentsquare} standardizes modules for automatic recombination. Yet these methods mainly aim to discover high-performing configurations under a target metric, offering limited insight into why a configuration works, how much each module contributes, or when modules interact constructively or destructively. Rather than only searching for the best agent, \ours exposes the agent design space as a controlled platform for analyzing component-level and interaction-level effects across tasks and backbones.
\section{Modular Design}

\begin{figure*}[t]
\begin{center}
\includegraphics[width=\textwidth]{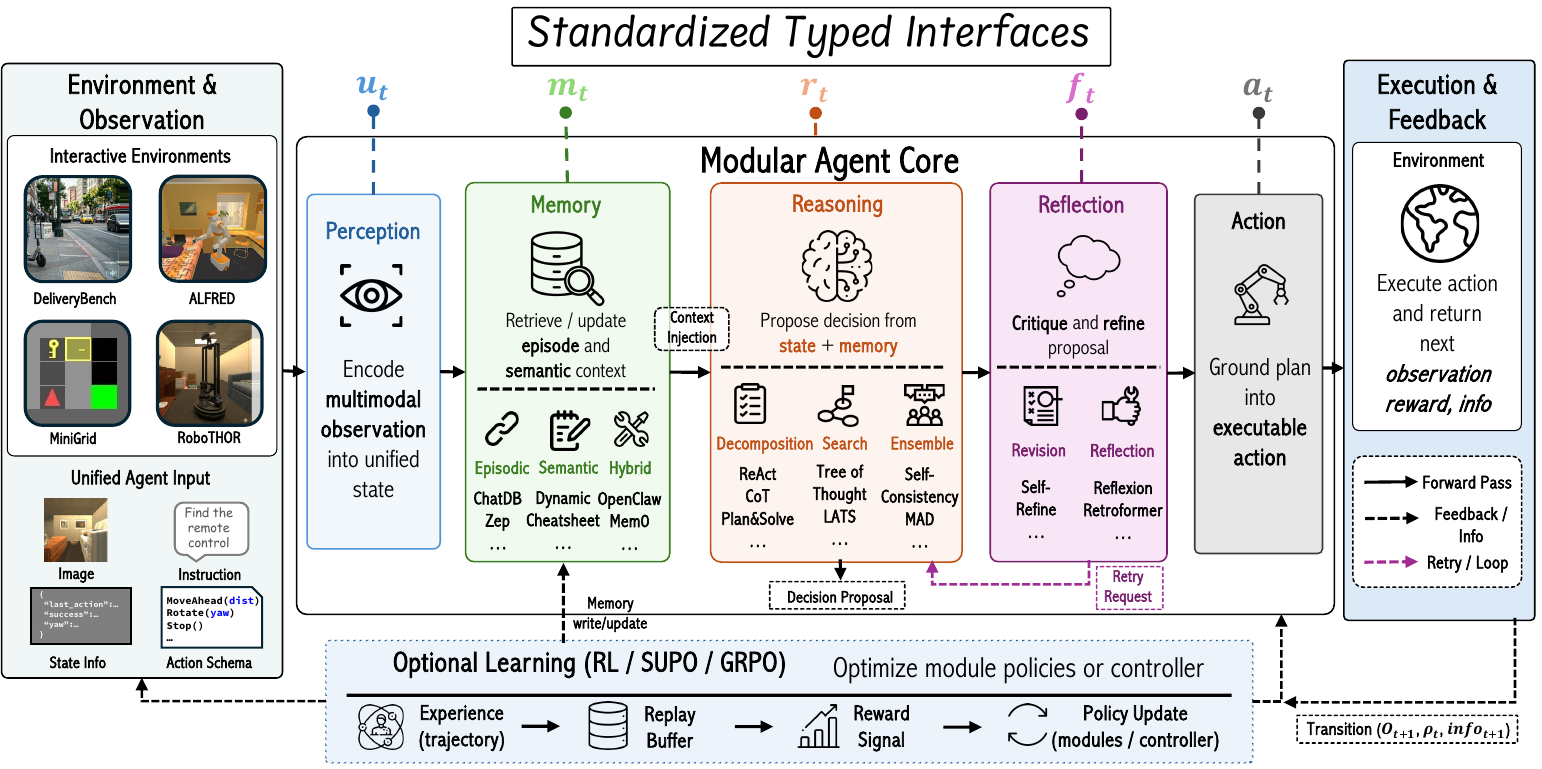}
\end{center}
\vspace{-4pt}
\caption{\textbf{Framework overview of \ours.}
\ours decomposes embodied decision-making into a typed Perception--Memory--Reasoning--Reflection--Action loop, where transition feedback updates memory and optional learning optimizes module policies or the controller.
}
\vspace{-0.5cm}
\label{fig:agentspec_main}
\end{figure*}

We instantiate \textsc{AgentSpec} as a Gym-compatible agent wrapper organized around a modular \textbf{Perception--Memory--Reasoning--Reflection--Action} loop, as shown in Figure~\ref{fig:agentspec_main}. The key design choice is not simply modularization but interface control, which indicates that every module receives and emits a typed intermediate object, so changing one component does not require rewriting the rest of the agent. The interaction can be viewed as a partially observable sequential decision-making problem, where the environment is modeled as $(\mathcal{S}, \mathcal{O}, \mathcal{A}, T, \rho)$ with \textit{latent states}, \textit{observations}, \textit{actions}, \textit{transition dynamics}, and \textit{rewards}. This design separates capabilities that are conceptually distinct but often entangled in embodied-agent systems: interpreting heterogeneous observations, retaining task-relevant history, reasoning over actions, revising decisions, and executing valid environment actions.

At each time step $t$, the agent receives a task description $d$ and raw observation $o_t \in \mathcal{O}$. The perception module abstracts them into a unified state representation $u_t = \mathcal{P}(d, o_t)$; memory retrieves relevant historical context $m_t = \mathcal{M}(h_{<t})$; reasoning produces an initial decision $r_t = \mathcal{R}(u_t, m_t)$; and reflection refines it into $\hat{r}_t = \mathcal{F}(r_t)$, which is converted into an executable action $a_t \in \mathcal{A}$. The environment returns the next observation $o_{t+1}$, reward $\rho_t$, and termination signal $\textit{done}_t$; the transition is then written back to memory, with optional reinforcement-learning updates. By fixing interface-level computation while varying module implementations, \textsc{AgentSpec} enables controlled studies of how agent components and their interactions affect performance.

\paragraph{Perception.}
The perception module $\mathcal{P}$ converts heterogeneous observations, such as symbolic states, sensor readings, RGB frames, and textual feedback, into a standardized representation for downstream modules. It normalizes raw inputs into structured JSON-like fields and concise textual summaries consumed by language-model-based components. By decoupling environment-specific observations from downstream control, perception allows the same memory, reasoning, and reflection modules to operate across environments while preserving task-relevant structure.

\paragraph{Memory.}
The memory module $\mathcal{M}$ stores and retrieves information beyond the current context, allowing the agent to accumulate experience and reuse task-relevant knowledge. In \ours, memory includes both \emph{episodic} and \emph{semantic} forms: episodic memory records concrete trajectories, action sequences, failures, and successes as raw logs, summaries, or selectively retained experiences, while semantic memory stores reusable knowledge such as maps, domain constraints, strategies, and heuristics.

\ours supports and compares multiple memory paradigms, including retrieval-based methods, persistent guidance such as dynamic cheatsheets~\citep{suzgun2026dynamic} or writable notebooks, and broader agent-centric memory engineering approaches such as Agent Context Engineering (ACE)~\citep{zhang2025agentic}. It also accommodates recent memory systems including CAM~\citep{li2025cam}, Zep~\citep{rasmussen2025zep}, MemGPT~\citep{packer2023memgpt}, Mem0~\citep{chhikara2025mem0}, SimpleMem~\citep{liu2026simplemem}, and OpenClaw context management, where external knowledge and semantic memory are represented as maintainable agent skills.

\paragraph{Reasoning.}
The reasoning module $\mathcal{R}$ maps the current state representation and retrieved memory to an action proposal with supporting rationale. Under a shared interface, \textsc{AgentSpec} supports direct methods such as Chain-of-Thought (CoT)~\citep{wei2022chain}, interactive methods such as ReAct~\citep{yao2022react}, and search-based methods such as Reasoning via Planning (RAP)~\citep{hao2023reasoning} and Tree of Thoughts (ToT)~\citep{yao2023tree}. This modularization allows us to study not only which reasoning strategy performs best, but also how reasoning interacts with memory, reflection, and computational cost.

\paragraph{Reflection.}
The reflection module $\mathcal{F}$ critiques or revises intermediate decisions before execution and can reuse feedback from prior failures. Under a shared interface, \textsc{AgentSpec} supports step-level reflection such as Self-Refine~\citep{madaan2023selfrefine}, trajectory-level verbal feedback such as Reflexion~\citep{shinn2023reflexion}, and retrospective trajectory analysis such as Retroformer~\citep{yao2023retroformer}. This enables controlled study of when explicit revision improves decisions and when it mainly increases inference cost.

\paragraph{Reinforcement Learning.}
\ours supports reinforcement learning as an optional module for improving agent policies from environment feedback. The framework is compatible with policy optimization methods such as \textsc{GRPO}~\citep{guo2025deepseek,deepseekmath} and exposes a unified interface for integrating learning with reasoning, memory, and action. The RL module is task-agnostic and can be applied across environments supported by \ours, including DeliveryBench and AI2-THOR, enabling analysis of both learning itself and its interaction with different reasoning and memory modules.
\section{Experiments}\label{others}

\subsection{Evaluation Benchmarks}
\label{sec:evaluation_benchmarks}

We evaluate \ours on four embodied-agent benchmarks that require multi-step interaction with an environment, covering complementary dimensions of embodied intelligence: DeliveryBench~\citep{mao2025deliverybench} for long-horizon planning under resource constraints, ALFRED~\citep{shridhar2020alfred} for compositional household instruction following, MiniGrid~\citep{chevalier2023minigrid} for symbolic navigation under partial observability, and RoboTHOR~\citep{deitke2020robothor} for object navigation in photorealistic 3D scenes. These benchmarks vary in observation modality, environment realism, task horizon, and required skills, allowing us to assess whether \ours generalizes across diverse embodied settings with minimal environment-specific adaptation. More details are provided in Appendix~\ref{app:evaluation_settings}.

\textbf{DeliveryBench} is a city-scale delivery benchmark across 9 urban maps, testing long-horizon planning under consumable resource constraints with structured and natural-language observations. Performance is measured by hourly profit. \textbf{ALFRED} evaluates long-horizon household instruction following in 3D environments, requiring object manipulation and state-dependent reasoning across seven task types. We report Success Rate (SR) and Success-weighted Path Length (SPL). \textbf{MiniGrid} is a 2D gridworld benchmark with pixel observations, evaluating navigation, object interaction, and reasoning under partial observability across 10 tasks, using SR and an SPL-style efficiency reward. \textbf{RoboTHOR} evaluates first-person object navigation in photorealistic indoor scenes, where agents navigate to a target object category within a limited step budget; we report SR and SPL.

\subsection{Overall Performance Across Tasks}

\paragraph{Experimental Setup.}
We conduct the main experiments on DeliveryBench under the \textbf{1-hour setting}, a long-horizon and resource-constrained testbed for evaluating how agent modules affect embodied decision-making. We compare configurations along three axes---reasoning, memory, and reflection---across both open-source and closed-source backbones. For each backbone, we start from a lightweight base agent with simple memory and no additional reasoning or reflection, then swap in different module variants under a unified protocol. We also evaluate representative configurations on MiniGrid, ALFRED, and RoboTHOR to test whether the trends transfer across environments with different horizons, observation modalities, and task structures.


\begin{figure*}[t]
    \centering
    \includegraphics[width=0.9\textwidth]{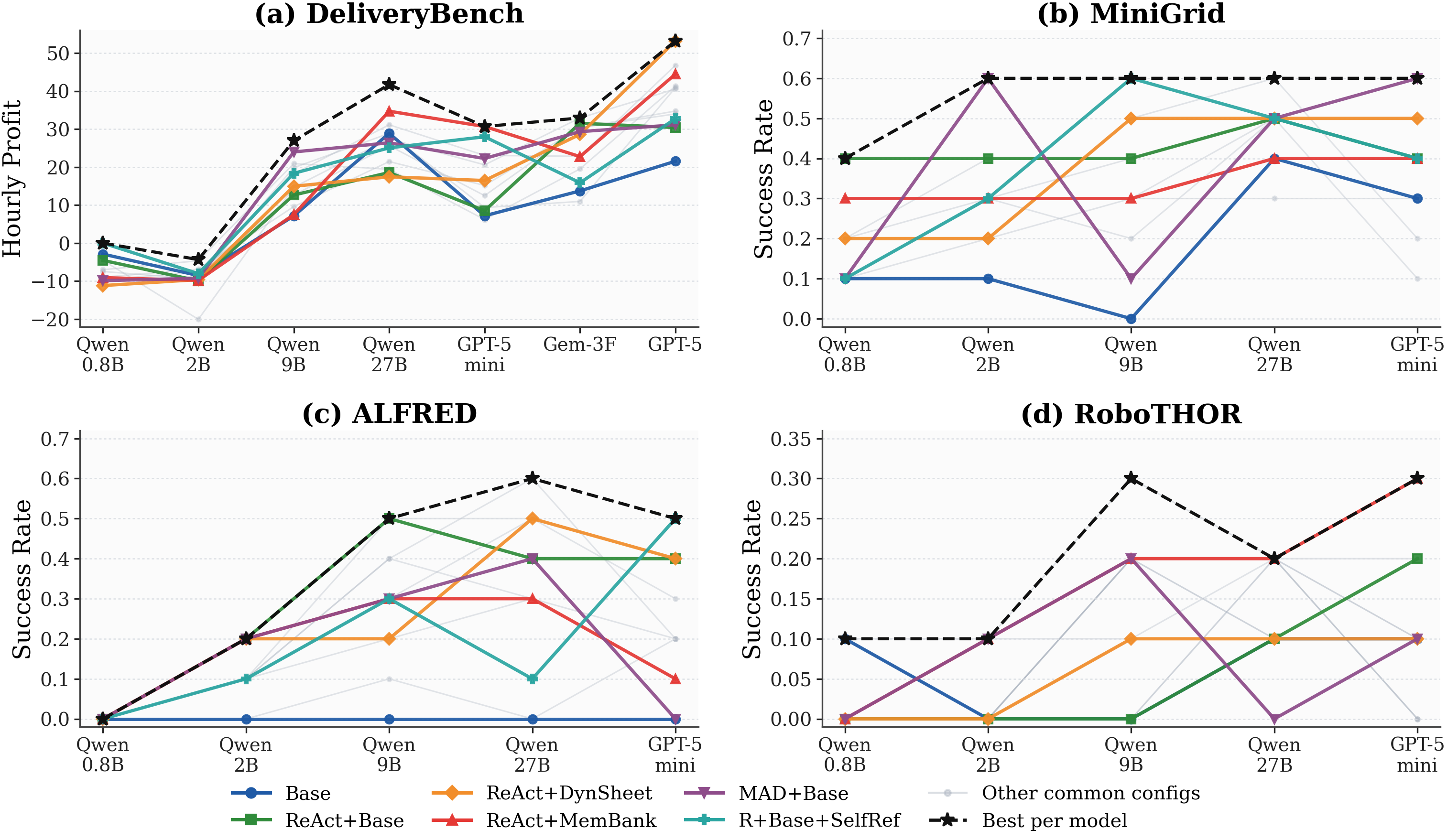}
    \caption{\textbf{
    Main results across four interactive-agent benchmarks.} Colored curves denote representative modular configurations, gray curves denote additional common configurations, and the dashed star-marked curve shows the best configuration for each backbone. DeliveryBench reports hourly profit; the other benchmarks report success rate. Higher is better. Full results are provided in Appendix~\ref{sec:detailed_experiment_analysis}. }
    \label{fig:main_four_benchmarks}
\end{figure*}

\paragraph{Scaffolding narrows the backbone gap.}
As shown in Figure~\ref{fig:main_four_benchmarks}(a), stronger backbones generally achieve higher best-case DeliveryBench performance, with Qwen-27B and closed-source models outperforming smaller Qwen variants. However, the gap is not determined by scale alone. Well-matched modular configurations allow smaller or open-source models to approach stronger models: Qwen-27B with ReAct+MemoryBank performs strongly, and Qwen-9B also benefits from memory- and reasoning-augmented variants. This suggests that external scaffolding can compensate for weaker native long-horizon reasoning, especially when it provides structured state tracking or reusable task guidance.

\paragraph{Memory helps only when it is structured.}
Figure~\ref{fig:main_four_benchmarks}(a) shows that memory can substantially improve DeliveryBench performance, but its benefit depends on how information is organized and retrieved. Structured, action-oriented memories such as DynamicCheatsheet and MemoryBank provide compact guidance useful for the next decision, while less selective memory may introduce stale or weakly relevant context. Thus, effective memory is not simply longer context; it must preserve task-relevant experience in a concise and controllable form. As shown in Figure~\ref{fig:modular_combination_gpt5mini}, on DeliveryBench with GPT-5 mini, replacing Base memory with MemoryBank increases ReAct from 8.54 to 30.67 and Plan-and-Solve from 6.18 to 26.78, while CoT shows a weaker and less monotonic pattern. This indicates that memory gains are mediated by the reasoner that consumes them.

\paragraph{The best module depends on the environment.}
The cross-benchmark results in Figure~\ref{fig:main_four_benchmarks}(b--d) show that module effectiveness varies across environments. In MiniGrid, where tasks are shorter and more symbolic, reasoning-oriented configurations often match the best envelope, while memory gains are less consistent. In ALFRED, both memory and reasoning matter because agents must maintain coherence over multi-step household instructions. In RoboTHOR, success remains lower and the best configuration varies across backbones, suggesting additional bottlenecks from perception, navigation, and long-horizon recovery. Overall, modules should be selected based on the dominant failure mode: reasoning improves local decision structure, memory supports long-horizon state tracking, and reflection helps when errors can be corrected through explicit revision.

\subsection{How Modules Interact}

\begin{figure}[t] 
    \centering
    \begin{minipage}[t]{0.48\textwidth}
        \centering
        \centering
\begin{tikzpicture}
\begin{axis}[
    width=1\linewidth,
    height=0.635\linewidth,
    enlargelimits=false,
    xmin=-0.5, xmax=4.5,
    ymin=-0.5, ymax=3.5,
    y dir=reverse,
    axis line style={draw=none},
    tick style={draw=none},
    xtick={0,1,2,3,4},
    xticklabels={Base, SMem, DC, CDB, MB},
    ytick={0,1,2,3},
    yticklabels={R, CoT, P\&S, MAD},
    xticklabel style={font=\tiny},
    yticklabel style={font=\tiny},
    colorbar,
    colorbar style={
        width=0.16cm,
        yticklabel style={font=\scriptsize},
        ytick align=outside,
    },
    point meta min=5,
    point meta max=31,
    colormap={softblue}{
        rgb255=(247,251,255)
        rgb255=(230,240,250)
        rgb255=(210,228,244)
        rgb255=(186,215,236)
        rgb255=(150,196,224)
        rgb255=(115,171,207)
        rgb255=(74,144,189)
        rgb255=(44,109,168)
    }
]
\addplot[
    matrix plot*,
    mesh/cols=5,
    point meta=explicit,
    draw=white,
    line width=0.8pt,
] coordinates {
    (0,0) [8.54400767]
    (1,0) [15.16442767]
    (2,0) [16.45667044]
    (3,0) [9.348806333]
    (4,0) [30.66717333]

    (0,1) [12.480542]
    (1,1) [7.69127667]
    (2,1) [13.18834633]
    (3,1) [12.72900195]
    (4,1) [13.37472467]

    (0,2) [6.184020333]
    (1,2) [17.43355333]
    (2,2) [21.91184667]
    (3,2) [11.02592933]
    (4,2) [26.77506333]

    (0,3) [22.25709]
    (1,3) [24.72145333]
    (2,3) [16.62626333]
    (3,3) [18.47167333]
    (4,3) [29.46289]
};

\node[font=\bfseries\scriptsize, text=black] at (axis cs:0,0) {8.54};
\node[font=\bfseries\scriptsize, text=black] at (axis cs:1,0) {15.16};
\node[font=\bfseries\scriptsize, text=black] at (axis cs:2,0) {16.46};
\node[font=\bfseries\scriptsize, text=black] at (axis cs:3,0) {9.35};
\node[font=\bfseries\scriptsize, text=black] at (axis cs:4,0) {30.67};

\node[font=\bfseries\scriptsize, text=black] at (axis cs:0,1) {12.48};
\node[font=\bfseries\scriptsize, text=black] at (axis cs:1,1) {7.69};
\node[font=\bfseries\scriptsize, text=black] at (axis cs:2,1) {13.19};
\node[font=\bfseries\scriptsize, text=black] at (axis cs:3,1) {12.73};
\node[font=\bfseries\scriptsize, text=black] at (axis cs:4,1) {13.37};

\node[font=\bfseries\scriptsize, text=black] at (axis cs:0,2) {6.18};
\node[font=\bfseries\scriptsize, text=black] at (axis cs:1,2) {17.43};
\node[font=\bfseries\scriptsize, text=black] at (axis cs:2,2) {21.91};
\node[font=\bfseries\scriptsize, text=black] at (axis cs:3,2) {11.03};
\node[font=\bfseries\scriptsize, text=black] at (axis cs:4,2) {26.78};

\node[font=\bfseries\scriptsize, text=black] at (axis cs:0,3) {22.26};
\node[font=\bfseries\scriptsize, text=black] at (axis cs:1,3) {24.72};
\node[font=\bfseries\scriptsize, text=black] at (axis cs:2,3) {16.63};
\node[font=\bfseries\scriptsize, text=black] at (axis cs:3,3) {18.47};
\node[font=\bfseries\scriptsize, text=black] at (axis cs:4,3) {29.46};

\end{axis}
\end{tikzpicture}
        
        \caption{Modular combination performance (mean hourly profit over three runs) on DeliveryBench using GPT-5 mini.\protect\footnotemark}
\label{fig:modular_combination_gpt5mini}
    \end{minipage}
    \hfill 
    \begin{minipage}[t]{0.48\textwidth}
        \centering
        \centering
\begin{tikzpicture}

\definecolor{myBlueEmbed}{RGB}{70,130,180}   
\definecolor{myBlueBorder}{RGB}{50,100,150}  

\definecolor{myRedEmbed}{RGB}{240,128,128}   
\definecolor{myRedBorder}{RGB}{200,100,100}   



\begin{axis}[
    ybar,
    bar width=0.25cm,
    width=1\linewidth,
    height=0.6\linewidth,
    enlarge x limits=0.2,
    symbolic x coords={P\&S+Base, CoT+Base, R+Base, CoT+SMem, MAD+MB},
    xtick=data,
    xticklabel style={font=\fontsize{5}{6}\selectfont, anchor=north},   
    ymin=0, ymax=50, 
    ytick={0,10,20,30,40,50},
    yticklabel style={font=\tiny},
    nodes near coords,
    nodes near coords style={
        font=\tiny,
        /pgf/number format/fixed,
        /pgf/number format/precision=1
    },
    axis lines=box,
    clip=false,           
    axis line style={thin},  
    xtick pos=bottom,
    ytick pos=left,
    xtick align=outside,
    ytick align=inside,
    legend style={
        at={(0.02,0.96)}, 
        anchor=north west,
        font=\tiny,
        fill opacity=0.8,
        draw=black!50,
        inner sep=2pt
    },
    legend image code/.code={
        \draw[#1, fill=#1] (0cm,-0.05cm) rectangle (0.15cm,0.05cm);
    },
]

\addplot[draw=myBlueBorder, fill=myBlueEmbed] coordinates {
    (P\&S+Base, 6.18) (CoT+Base, 12.48) (R+Base, 8.54) (CoT+SMem, 7.69) (MAD+MB, 29.46)
};
\addlegendentry{w/o refine}

\addplot[draw=myRedBorder, fill=myRedEmbed] coordinates {
    (P\&S+Base, 18.64) (CoT+Base, 18.97) (R+Base, 27.97) (CoT+SMem, 24.17) (MAD+MB, 35.44)
};
\addlegendentry{w/ refine}

\end{axis}
\end{tikzpicture}
        \caption{Mean hourly profit of Self-Refine on selected modular combinations (\textsc{DeliveryBench}, GPT-5 mini).}
        \label{fig:self_refine_comparison}
    \end{minipage}
    \vspace{-0.4cm}
\end{figure}
\footnotetext{Abbreviations used throughout the paper: R=ReAct, CDB=ChatDB, SMem=SimpleMem, MB=MemoryBank, DC=DynamicCheatsheet, OC=OpenClaw, P\&S=Plan-and-Solve, SR=SelfRefine, Rfx=Reflexion.}

\paragraph{Experimental Setup.}
We further analyze module interactions on DeliveryBench. Since module effects in long-horizon embodied tasks are rarely independent, we compare reasoning, memory, and reflection combinations under a unified setting to identify helpful pairings and harmful mismatches.

\paragraph{Planning benefits from abstract memory.}
As shown in Figure~\ref{fig:modular_combination_gpt5mini}, ReAct and Plan-and-Solve gain only modestly from low-level memories such as Base and ChatDB, which mainly return raw observations or trajectory fragments. They improve substantially with abstract memories such as SimpleMem, DynamicCheatsheet, and MemoryBank, suggesting that planning-oriented reasoning benefits more from distilled strategies than exact historical states. Raw observations can support local control, but summarized memory better matches long-horizon plan-then-act reasoning (see Appendix~\ref{app:case_planning_memory} and Appendix~\ref{app:case_dc_memory_triage}).

\paragraph{Multi-granularity memory is the safest default.}
MemoryBank performs strongest across reasoning methods because it combines raw trajectories, experience summaries, and higher-level environmental insights. This lets each reasoning strategy use memory at the granularity it needs, making the module robust to different downstream reasoning styles (see Appendix~\ref{app:case_memorybank_charging} and Appendix~\ref{app:case_openclaw_waittime_charging}).

\paragraph{Multi-agent reasoning tolerates weaker memory.}
Across memory settings, multi-agent reasoning remains effective even with low-level memory, showing lower sensitivity to memory quality than single-pass reasoning. Its robustness likely comes from error correction and complementary perspectives, which help maintain decision quality when retrieved memory is coarse or partially relevant (see Appendix~\ref{app:case_mad_debate}).

\paragraph{Reflection is a general-purpose correction layer.}
As Figure~\ref{fig:self_refine_comparison} shows, reflection yields large gains for weak reasoning--memory pairs and smaller but consistent gains for strong ones. This suggests that reflection is broadly compatible: by re-evaluating candidate actions against state feedback and recent failures, it reduces local execution errors before they accumulate over long horizons (see Appendix~\ref{app:case_selfrefine_bag_recovery}).

\section{Analysis}

\subsection{Efficiency--Performance Trade-off}

\begin{figure}[t]
    \centering
    \includegraphics[width=\columnwidth]{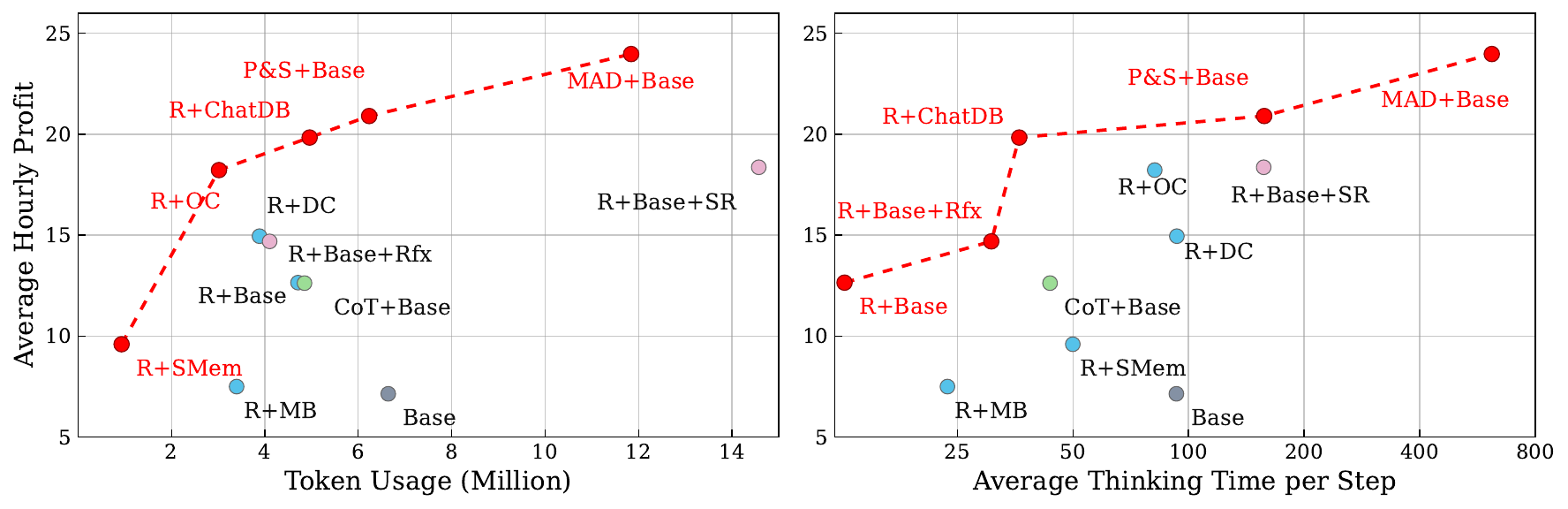}
    \caption{Pareto frontier of Qwen3.5-9B across different module combinations, showing the trade-off between (a) Profit vs. Token Usage and (b) Profit vs. Thinking Time (in log axis).}
    \label{fig:pareto}
    \vspace{-0.5cm}
\end{figure}

\paragraph{Motivation.}
Performance alone does not fully characterize embodied agents, since stronger module combinations may introduce substantial inference overhead over long-horizon interactions. We therefore evaluate each configuration by both effectiveness and efficiency, using \textit{total token usage} and \textit{average thinking time per step} as the main efficiency metrics.

\paragraph{More computation is not always better.}
As shown in Figure~\ref{fig:pareto}a, configurations with similar token budgets can achieve very different profits, while some higher-cost methods offer limited additional gains. Several ReAct-based variants lie on the Pareto frontier, suggesting that strong performance can often come from better module composition rather than simply larger inference budgets. Thus, token efficiency depends not only on reasoning strength, but also on whether reasoning and memory are well matched.

\paragraph{Latency depends on alignment, not just deliberation.}
A similar pattern appears in Figure~\ref{fig:pareto}b. Higher-profit configurations do not always require proportionally longer thinking time: some achieve strong returns with moderate latency, whereas others spend more time without comparable gains. This shows that useful computation is task- and module-aligned computation, not merely more deliberation.

\subsection{Case Study and Error Analysis}

\begin{figure}[t]
    \centering
    \includegraphics[
        width=\columnwidth
    ]{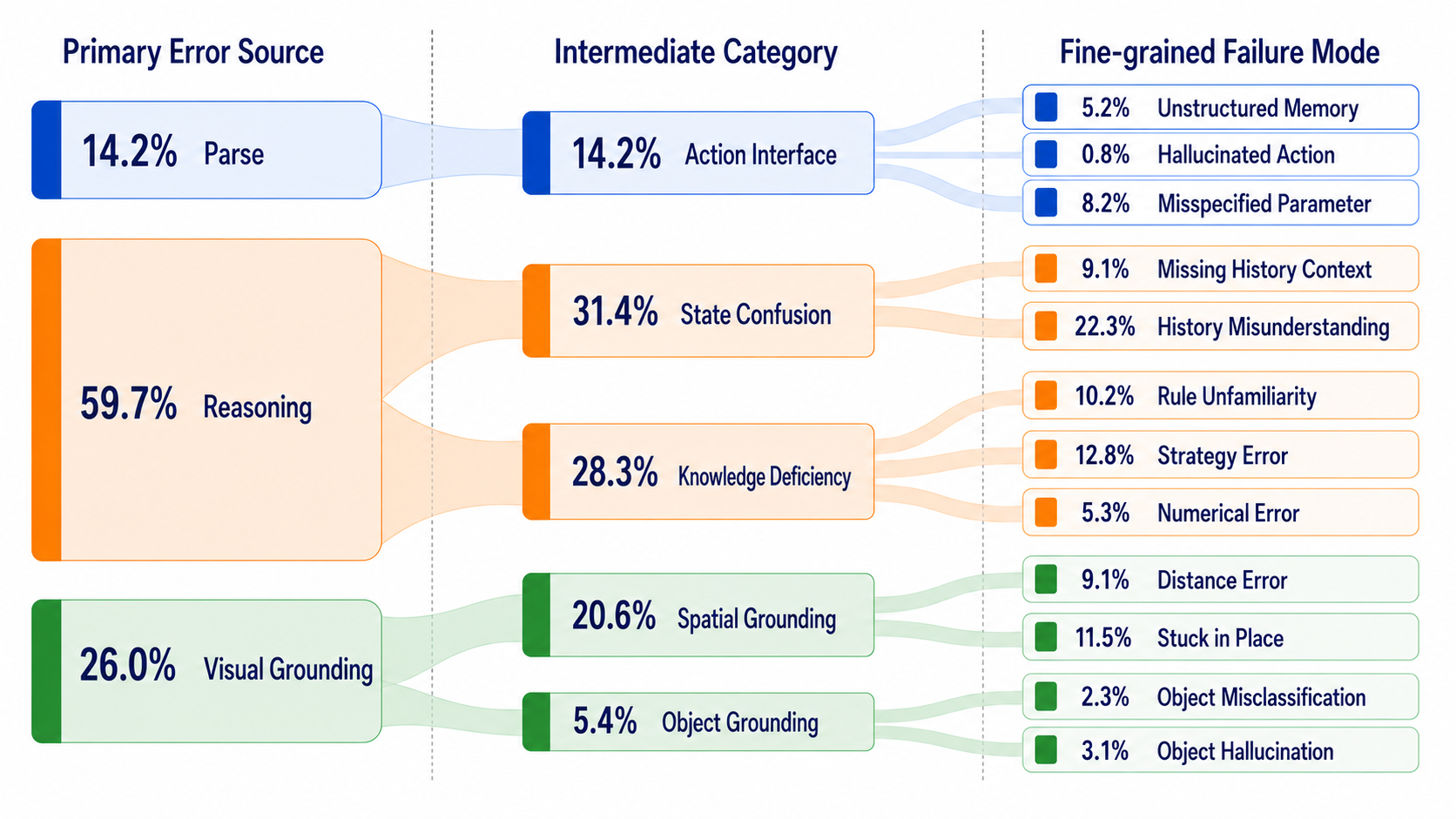}
\vspace{-0.7cm}
    \caption{Overall failure taxonomy across benchmarks.}
    \label{fig:error_sankey}
\end{figure}

\begin{figure}[t]
    \centering
    \begin{minipage}[t]{0.48\textwidth}
        \centering
        \includegraphics[width=\linewidth, trim=4pt 80pt 8pt 4pt, clip]{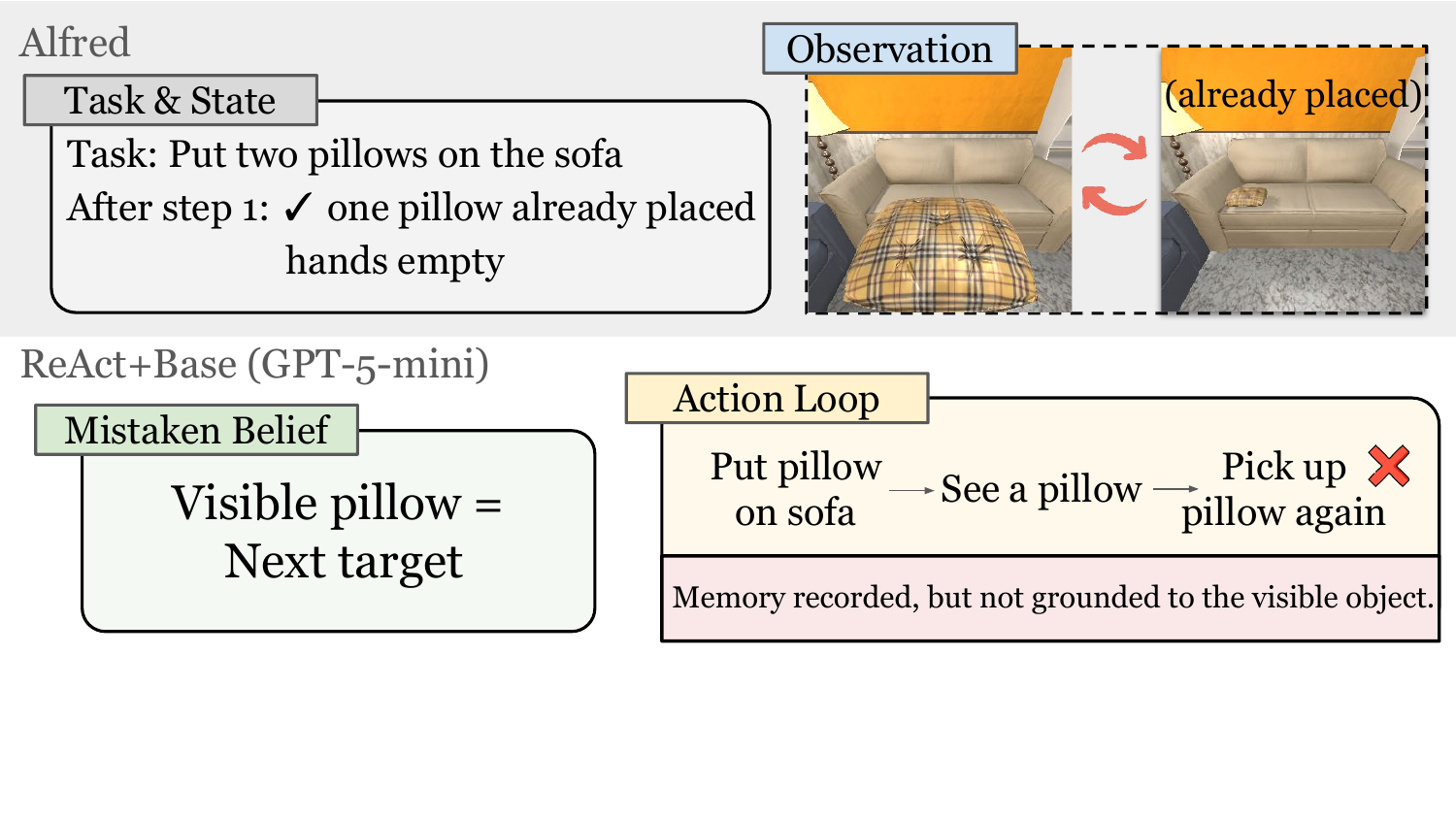}
        \vspace{-8mm}
        \caption{An ALFRED failure case where the agent fails to align recent history with the current observation, leading to a repeated-action loop.}
        \label{fig:alfred_case}
    \end{minipage}
    \hfill
    \begin{minipage}[t]{0.48\textwidth}
        \centering
        \vspace{2mm}
        \includegraphics[width=\linewidth, trim=4pt 40pt 9pt 4pt, clip]{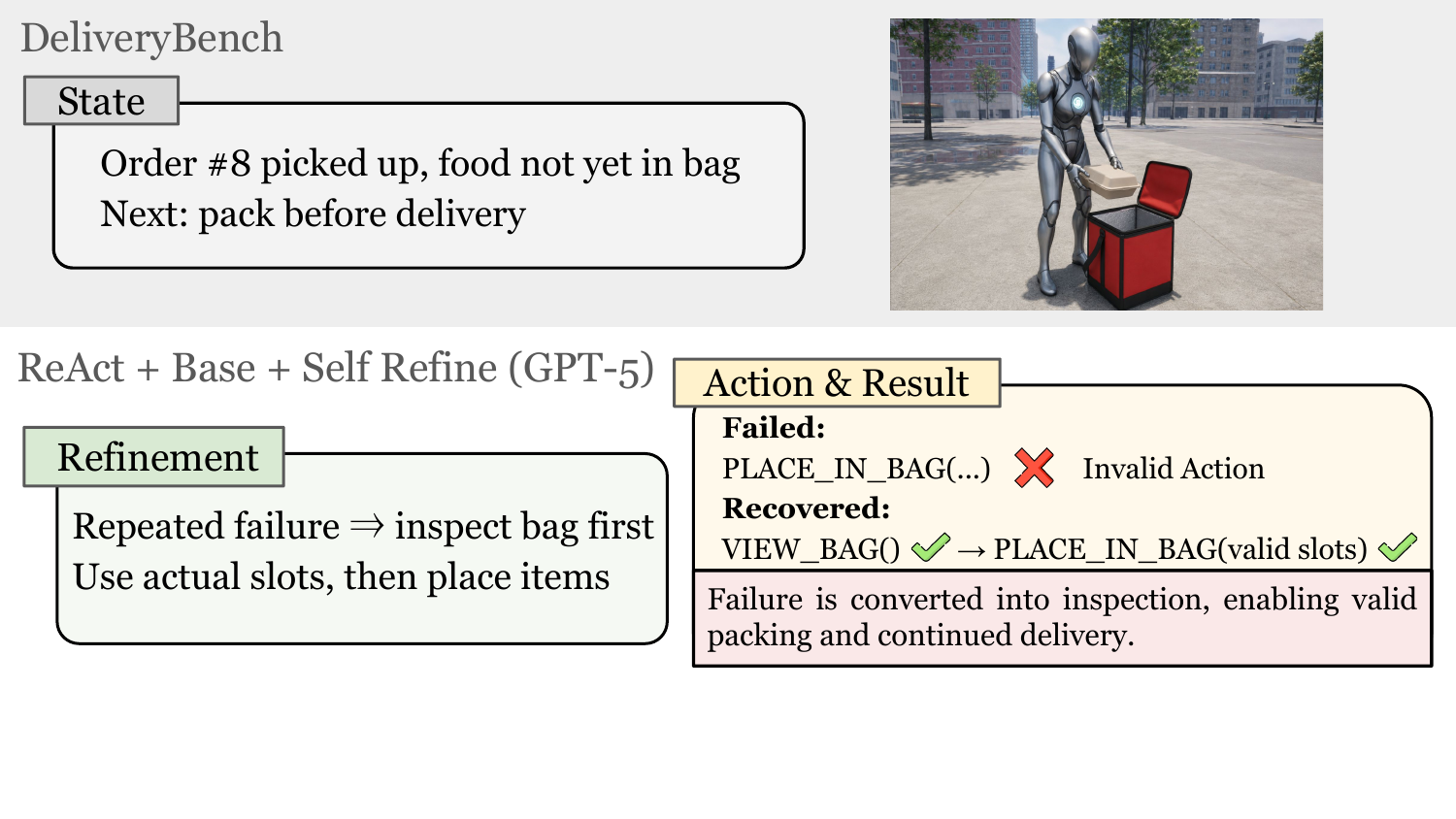}
            \vspace{-11mm}
        \caption{A DeliveryBench recovery case where Self-Refine turns repeated bagging failures into an inspect-and-repair process.}

        \vspace{-0.5cm}
        \label{fig:deliverybench_case}
    \end{minipage}
\end{figure}

\paragraph{Many failures come from losing the task state.}
Figure~\ref{fig:error_sankey} shows that failures across benchmarks often involve history misunderstanding: the agent fails to track the current state from past interactions, mistakes completed progress for the current objective, and falls into repetitive loops. Beyond this shared pattern, RoboTHOR and ALFRED failures concentrate more on visual grounding, while MiniGrid and DeliveryBench errors more often reflect poor adaptation to environment rules, action constraints, and long-horizon strategy, as detailed in Table~\ref{tab:failure_taxonomy_by_benchmark}.

\paragraph{Memory tracks progress; reflection repairs execution.}
Figure~\ref{fig:alfred_case} illustrates a representative ALFRED failure caused by state confusion. Although one pillow has already been placed, the agent treats the visible pillow as the next target and repeatedly alternates between picking and placing actions, forming a loop. This suggests that lightweight episodic memory may not reliably distinguish completed progress from visually similar current observations. DeliveryBench exposes a different challenge: its action interface requires precise parameterization, and agents often fail by issuing invalid or over-compressed commands. Figure~\ref{fig:deliverybench_case} shows that Self-Refine mitigates this by redirecting the agent to inspect the bag state, recover missing information, and resume with valid actions. Together, these cases show that memory is crucial for long-horizon progress tracking, while reflection helps diagnose invalid assumptions and repair failed action plans.

\subsection{Learning with Modular Scaffolds}

\begin{table}[t]
\centering
\resizebox{\columnwidth}{!}{%
\begin{tabular}{@{}lccc@{}}
\toprule
\textbf{Method} & \textbf{Non-RL} & \textbf{GRPO} & \textbf{SUPO} \\
\midrule
Base      & -3.07 & 5.80 & 5.48 \\
MAD+Base  & -3.20 & 7.87 & 6.56 \\
ReAct+DynamicCheatsheet       & -2.89 & 5.02 & \textbf{8.27} \\
ReAct+MemoryBank      & 2.90 & 4.03 & 7.07  \\
ReAct+OpenClaw      & 3.36 & 4.79 & 6.57 \\
ReAct+Base    & 0.00 & 5.62 & 5.83 \\
\bottomrule
\end{tabular}
}
\vspace{-5pt}
\caption{Effect of inference-time modules on different policy backbones. \textsc{SUPO} shows stronger compatibility with modular agent components, likely because its explicit trajectory-summary training exposes the policy to structured history during RL.}
\label{tab:rl_combined}
\vspace{-0.5cm}
\end{table}

\paragraph{Experimental Setup.}
We instantiate the learning module with \textsc{GRPO}~\citep{deepseekmath} on \textsc{Qwen3-4B-Instruct-2502} and train on DeliveryBench using \emph{earning delta} as the reward $\rho_t$. Training uses up to 40 interaction turns per episode, 4 rollouts, and a batch size of 32. \textsc{SUPO}~\citep{supo} follows the same environment, reward, and rollout setup, but augments policy learning with trajectory summarization: every 8 actions, the model summarizes the interaction history and conditions subsequent decisions on the generated summary. We then evaluate each learned policy under the same downstream inference-time modules. This design lets us separate two questions: whether RL improves the underlying policy in isolation, and whether the learned policy remains compatible with modular agent scaffolds such as reasoning and memory.

\paragraph{RL improves the bare policy, but does not automatically solve agent composition.}
Table~\ref{tab:rl_combined} shows that RL substantially improves the unscaffolded or lightly scaffolded policy. For the Base configuration, performance increases from $-3.07$ without RL to $5.80$ with \textsc{GRPO} and $5.48$ with \textsc{SUPO}. Similar gains appear for ReAct+Base, where the non-RL policy obtains $0.00$, while \textsc{GRPO} and \textsc{SUPO} reach $5.62$ and $5.83$, respectively. This indicates that reward optimization teaches the model useful DeliveryBench-specific behaviors, such as respecting action constraints, selecting profitable orders, and avoiding locally invalid decisions. However, these improvements mainly reflect a stronger underlying decision policy. They do not imply that RL-trained policies can automatically exploit richer agent frameworks once reasoning or memory modules are attached at inference time.

\paragraph{Post-hoc scaffolding can be misaligned with standard RL.}
The \textsc{GRPO} results reveal a mismatch between policy learning and inference-time scaffolding. Although \textsc{GRPO} improves the Base policy, adding structured memory on top of the \textsc{GRPO}-trained model does not consistently produce further gains. For example, ReAct+DynamicCheatsheet, ReAct+MemoryBank, and ReAct+OpenClaw obtain $5.02$, $4.03$, and $4.79$, all below or comparable to the simpler \textsc{GRPO} Base and ReAct+Base configurations. This suggests that a policy optimized under the original prompt--observation interface may learn to rely on its training-time input format and action habits. When external memory later changes the context distribution, the learned policy may not know how to use the additional structured information effectively. In this sense, stronger RL on the ``bare'' LLM does not necessarily translate into stronger RL for the full agent.

\paragraph{Summary-based learning improves scaffold compatibility.}
In contrast, \textsc{SUPO} composes more favorably with memory-centric scaffolds. Under the same inference-time modules, \textsc{SUPO} outperforms \textsc{GRPO} on ReAct+DynamicCheatsheet ($8.27$ vs. $5.02$), ReAct+MemoryBank ($7.07$ vs. $4.03$), and ReAct+OpenClaw ($6.57$ vs. $4.79$). The best overall result in Table~\ref{tab:rl_combined} is not the strongest bare RL policy, but \textsc{SUPO} combined with DynamicCheatsheet. This pattern suggests that trajectory-summary training exposes the policy to a compressed, structured form of history during optimization, making it more receptive to structured memory at test time. Rather than treating memory as an external add-on, \textsc{SUPO} partially aligns the learned policy with the kind of context that modular agents provide.

\paragraph{Policies should be optimized with their deployment-time scaffolds.}
These results highlight a distinction between \emph{RL for LLM policies} and \emph{RL for agent frameworks}. Training a bare policy can improve local action selection, but modular agents also require interface compatibility: the policy must learn how to interpret retrieved memory, summarized history, reasoning traces, and reflection feedback. If these signals appear only at inference time, they may act as distribution shifts rather than useful scaffolds. More broadly, reasoning, memory, and reflection should be incorporated into policy optimization instead of being attached only after training. \textsc{SUPO} provides a lightweight step in this direction by training with summary-based context, but the larger implication is that future agent RL should optimize the policy and its modular scaffold jointly.
\section{Conclusion}


We introduced \ours, a modular specification framework for studying LLM-based embodied agents as typed compositions of perception, memory, reasoning, reflection, action execution, and optional learning components. \ours exposes the scaffold as a controlled design space in which modules can be swapped, recombined, and analyzed under shared interfaces rather than treating an agent scaffold as a fixed end-to-end pipeline. Across four embodied benchmarks and multiple model backbones, our experiments show that agent performance is shaped by compatibility among modules, not only by the isolated strength of individual components. Structured multi-granularity memory improves long-horizon state tracking, reasoning and memory interact in environment-dependent ways, reflection is most useful when it repairs local execution errors, and RL-trained policies compose best when their training-time context resembles the scaffolded context used at deployment time. These findings suggest that future LLM agent research should optimize not only the base policy, but also the interfaces and scaffold structures through which the policy perceives, remembers, reasons, and acts.

\bibliography{custom}

\appendix

\section*{Appendix}

\section{Limitations}

\label{app:limitations}

This work has several limitations. Although we evaluate multiple benchmarks, backbones, and module combinations, our study does not cover the full space of agentic tasks, and its embodied and decision-making focus may not directly transfer to domains such as web navigation, software engineering, or open-ended human--AI interaction. We also study representative implementations rather than exhaustively optimizing each module family; reasoning, memory, and reflection can depend on prompt design, retrieval and update policies, context budgets, and environment-specific interfaces. Due to computational constraints, we cannot scale every test set or run repeated trials for all model--module--environment combinations, so some individual results may be affected by sampling variation. Finally, while we analyze cost and qualitative failures, more fine-grained studies of token efficiency, latency, memory growth, retrieval quality, and trajectory-level error propagation are needed. We therefore emphasize consistent cross-setting trends and leave broader task coverage and deeper diagnostic analyses to future work.

\section{Extended Related Work}
\label{app:extended_related_work}

\textbf{LLM Reasoning.}
Beyond the prompting and search-based reasoning methods cited in the main text, a wide range of additional reasoning strategies have been developed. Tool-augmented, logic-aided, and backward-chaining methods ground reasoning in external computation or formal inference~\citep{kazemi2023lambada}, while constrained generation enforces structural validity during decoding~\citep{banerjee2025crane}. Self-correction is a particularly active area: iterative critique-and-revision~\citep{welleck2022generating,havrilla2024glore,gao2024embedding}, self-rewarding correction~\citep{xiong2025self}, cooperative reasoning across models~\citep{yang2025multi}, and adversarial self-play~\citep{li2026learning} all enable models to refine their own outputs, complemented by process-level verification~\citep{lightman2023let}. Multi-agent approaches leverage inter-agent debate and collaborative training~\citep{du2024improving,motwani2024malt,yang2026auditing,yang2025maestro}. Meta-level methods allow models to compose their own reasoning structures~\citep{zhou2024self}, treat reasoning as structure-aware planning with world models~\citep{xiong2025deliberate,xiang2025towards}, incorporate contrastive objectives into Monte Carlo tree search~\citep{gao2024interpretable}, or adapt reasoning depth to task complexity via cognitive-inspired sketching or efficiency analysis~\citep{sprague2024cot,ning2023skeleton,aytes2025sketch}. Latent reasoning methods bypass discrete token generation entirely, operating in continuous latent space~\citep{hao2024training} or via latent diffusion for iterative refinement and diverse solution exploration~\citep{kang2026beyond}, while GFlowNet-based training models reasoning as flow on a DAG to promote diverse generation~\citep{yu2024flow,kang2025gflowvlm}. Reasoning has also been extended to embodied action selection~\citep{li2024embodied}, grounded simulation~\citep{liu2022mind}, deep reasoning for translation~\citep{wang2025drt}, and interleaved reasoning-memorization via self-notes~\citep{lanchantin2023learning}. In \ours, all of these are interchangeable instantiations of a single reasoning interface, enabling controlled comparison within identical agent pipelines.

\textbf{Memory for LLM Agents.}
Beyond the memory categories surveyed in the main text, database-backed approaches store dialogue history for SQL-style retrieval~\citep{hu2023chatdb,modarressi2023ret}, and long-context memory extensions augment or replace the context window itself~\citep{liu2024memlong,xiao2024infllm,yang2024text}. Hierarchical and graph-based methods organize experience into multi-level semantic structures, temporal knowledge graphs, or event-centric logic maps~\citep{li2024graphreader,jiang2026magma,hu2026memory,anokhin2024arigraph}. Procedural memory systems store reusable action sequences, trajectory exemplars, or cross-domain skills for transfer~\citep{fang2025memp,han2025legomem,tang2025agent,xiao2025toolmem,forouzandeh2025learning,zheng2023synapse,wang2024agent_workflow,liu2025contextual,zhang2026memskill}, while self-organizing and evolving designs adaptively restructure memory through cognitive-inspired consolidation, RL-based construction, or meta-evolution~\citep{zhang2025memevolve,tian2025rgmem,qian2025memorag,cao2025remember,wei2025evo,yan2025memory}. Plug-and-play and parameter-efficient memory modules enable domain adaptation without retraining the base model~\citep{cao2025memory,bini2025memlora,packer2023memgpt}. Episodic and social memory grounds agents in persistent interaction histories and simulated human behavior~\citep{park2023generative,li2025metaagents,zhong2024memorybank,zhang2025ella}. Memory maintenance, segment-based organization, and semantic anchoring address staleness, coherence, and linguistic grounding over extended interactions~\citep{chen2025moom,zhang2025multi,chatterjee2025semantic,wang2026comorag,wang2025r3mem}. Multimodal and video-oriented memory extends these ideas to visual and lifelong learning streams~\citep{yeo2025worldmm,latimer2025hindsight,liu2025memverse}. A growing line of work further recognizes that memory and reasoning are mutually dependent: recall-and-post-thinking mechanisms augment reasoning with long-term memory~\citep{liu2023think}, reusable thought templates and evolving knowledge bases bridge past reasoning into current decisions~\citep{yang2024bufferofthoughts,suzgun2026dynamic,qian2025memorag}, hierarchical working memory and structured context manage cognitive load during multi-step reasoning~\citep{hu2025hiagent,wang2025scm}, and reasoning-memory co-evolution allows agents to jointly improve both capabilities~\citep{ouyang2025reasoningbank,ho2025arcmemo}. In \ours, these diverse memory designs are exposed through a shared module interface, enabling systematic evaluation of how storage, retrieval, and update choices---and their interaction with reasoning---shape agent performance.

\textbf{Reflection and Self-Improvement.}
Reflection mechanisms range from step-level critique-and-revision~\citep{paul2024refiner} to trajectory-level self-critique that stores verbal reflections from failed episodes~\citep{yao2023retroformer},reflection through trial-and-error driven test-time planning~\citep{hong2026learning} and experience replay approaches that extract and internalize transferable insights from past trajectories~\citep{zhao2024expel}. Experience-driven lifelong learning enables agents to self-evolve through accumulated interaction~\citep{wu2025evolver,chen2025scaling}. In \ours, reflection is a dedicated module decoupled from reasoning and memory, enabling controlled ablation of its marginal contribution and interaction effects with other components.

\section{Module Implementations}

\label{app:modules}

This section details the module implementations instantiated in \ours. For each module family, we list the representative methods included in our study, summarize their operational characteristics, and provide the corresponding references. All implementations follow the common interfaces introduced in the main paper, enabling controlled comparisons across reasoning, memory, and reflection mechanisms.

\subsection{Reasoning Methods} 
Table~\ref{tab:reasoning-methods} reports the reasoning strategies included in our implementation suite. The selected methods span linear decomposition, explicit planning, search-based exploration, consistency-based aggregation, and multi-agent deliberation, covering the main design choices for structuring intermediate inference in LLM agents.
\begin{table*}[t] \centering \small \begin{tabular}{lp{8cm}l} \toprule \textbf{Method} & \textbf{Description} & \textbf{Reference} \\ \midrule ReAct & Interleaves reasoning and acting over multiple Thought--Action--Observation cycles, grounding decisions in environmental feedback. & \citep{yao2022react} \\ \addlinespace CoT & Generates explicit intermediate reasoning steps before selecting an action, decomposing complex decisions into a chain of thoughts. & \citep{wei2022chain} \\ \addlinespace Plan-and-Solve & Produces a high-level plan before execution, then follows the plan step by step to complete the task. & \citep{wang2023plan} \\ \addlinespace Tree of Thoughts & Expands reasoning into a tree structure, exploring multiple thought branches and evaluating candidates via lookahead search. & \citep{yao2023tree} \\ \addlinespace LATS & Combines language model reasoning with Monte Carlo Tree Search, using execution feedback to guide trajectory exploration. & \citep{zhou2023language} \\ \addlinespace RAP & Treats reasoning as planning via MCTS, building a world model to simulate and evaluate action sequences before committing. & \citep{hao2023reasoning} \\ \addlinespace Self-Consistency & Samples multiple independent reasoning paths and aggregates their outputs to select the most consistent answer. & \citep{wang2022self} \\ \addlinespace MAD & Introduces critique and disagreement among several language agents through structured debate rounds before reaching a final decision. & \citep{du2024improving} \\ \bottomrule \end{tabular} \caption{Reasoning module implementations supported in \ours.} \label{tab:reasoning-methods} \end{table*}

\subsection{Memory Methods}
\label{app:memory_methods}

Table~\ref{tab:memory-methods} summarizes the memory mechanisms implemented in \ours. These methods cover short-context buffers, vector- and database-backed retrieval, hierarchical and graph-structured stores, procedural memory, and adaptive long-term memory systems, providing a broad basis for evaluating how storage, retrieval, and update policies affect agent behavior.

\begin{table*}[t]
\centering
\small
\begin{tabular}{lp{8cm}l}
\toprule
\textbf{Method} & \textbf{Description} & \textbf{Reference} \\
\midrule
Base & Maintains a rolling buffer of recent transitions appended directly to the prompt context. & --- \\
\addlinespace
A-Mem & Maintains memory notes with semantic metadata, LLM-powered content analysis, and relationship management via ChromaDB hybrid search. & \citep{xu2025mem} \\
\addlinespace
ACE & Agentic Context Engine applying iterative reflection, LLM-guided curation, and deduplication to maintain a compact, non-redundant memory store. & \citep{zhang2025agentic} \\
\addlinespace
Buffer of Thought & Maintains a reusable buffer of high-level thought templates distilled from past reasoning traces, enabling retrieval-augmented thought reuse at inference time. & \citep{yang2024bufferofthoughts} \\
\addlinespace
CAM & Organizes experience into a hierarchical semantic structure using incremental overlapping clustering with LLM-guided pruning and coherent summarization. & \citep{li2025cam} \\
\addlinespace
ChatDB & Stores dialogue history in a structured database, enabling SQL-style retrieval of past interactions. & \citep{hu2023chatdb} \\
\addlinespace
DC & Dynamic Cheatsheet that evolves a concise knowledge base via LLM curation and vector-store retrieval at test time. & \citep{suzgun2026dynamic} \\
\addlinespace
GMemory & Three-tier hierarchical graph memory spanning an interaction graph for trajectory condensation, a query graph for task retrieval, and an insight graph for high-level insight management. & \citep{zhang2025g} \\
\addlinespace
LangMem & LangChain-based memory infrastructure supporting episodic, semantic, and procedural memory types with structured storage and retrieval. & \citep{langmem2024} \\
\addlinespace
MemGPT & Manages memory through a paged context window with explicit in-context and archival storage layers, enabling unbounded long-term memory. & \citep{packer2023memgpt} \\
\addlinespace
LightMem & Manages memory through a multi-stage pipeline involving normalization, pre-compression, topic segmentation, LLM-based extraction, embedding, storage in Qdrant, and retrieval. & \citep{fang2025lightmem} \\
\addlinespace
Mem0 & Adaptive personal memory with self-updating storage integrating short-term and long-term retrieval across user sessions. & \citep{chhikara2025mem0} \\
\addlinespace
MemoryBank & Long-term memory bank that stores and retrieves past experiences using similarity-based lookup to inform future decisions. & \citep{zhong2024memorybank} \\
\addlinespace
MIRIX & Multi-agent memory system using event-time storage and context-aware retrieval via a dedicated Mirix client. & \citep{wang2025mirix} \\
\addlinespace
OpenClaw & Context management module that maintains procedural memory comprising learned policies and lessons, and augments agent prompts with relevant skill documents at each step. & \citep{openclaw2026} \\
\addlinespace
Generative Agent Memory & Scores memory entries by combining recency, semantic relevance, and reward utility into a unified ranking function. & \citep{park2023generative} \\
\addlinespace
SimpleMem & Employs a three-stage pipeline: semantic compression, vector storage, and hybrid retrieval across semantic, keyword, and structured indexes. & \citep{liu2026simplemem} \\
\addlinespace
Zep & Models memory as a temporal knowledge graph spanning episodes, entities, and higher-level communities. & \citep{rasmussen2025zep} \\
\bottomrule
\end{tabular}
\caption{Memory module implementations supported in \ours.}
\label{tab:memory-methods}
\end{table*}

\subsection{Reflection Methods}

Table~\ref{tab:reflection-methods} presents the reflection mechanisms included in our framework. These methods operationalize self-improvement at different temporal scales, from step-level critique and revision to episode-level retrospective analysis and reuse of lessons across subsequent trials.

\begin{table*}[!htbp]
\centering
\small
\begin{tabular}{lp{8cm}l}
\toprule
\textbf{Method} & \textbf{Description} & \textbf{Reference} \\
\midrule
\addlinespace
Self-Refine & Performs iterative critique-and-revision at the step level: a critic evaluates the current output and proposes improvements until the answer stabilizes or no further improvement is detected. & \citep{madaan2023selfrefine} \\
\addlinespace
Reflexion & Stores verbal reflections from failed episodes and reuses them in subsequent trials, enabling trajectory-level self-correction across episodes. & \citep{shinn2023reflexion} \\
\addlinespace
Retroformer & Retrospectively analyzes full trajectories after episode completion to distill lessons that improve later planning and decision-making. & \citep{yao2023retroformer} \\

\bottomrule
\end{tabular}
\caption{Reflection module implementations supported in \ours.}
\label{tab:reflection-methods}
\end{table*}

\section{Detailed Evaluation Settings}
\label{app:evaluation_settings}

{\let\section\subsection
\let\subsection\subsubsection
\section{DeliveryBench}
\label{settings:deliverybench}
\subsection{DeliveryBench}
\label{app:deliverybench_setting}

DeliveryBench~\cite{mao2025deliverybench} is a city-scale embodied benchmark in which the agent acts as an autonomous food courier in procedurally generated 3D cities. Unlike short-horizon embodied tasks, DeliveryBench emphasizes long-horizon decision making under realistic operational constraints. The agent must continuously choose profitable orders, travel between restaurants and customers, execute pickup and drop-off operations, and manage limited resources such as time, energy, and transportation cost. The benchmark therefore evaluates whether an agent can sustain constraint-aware planning over extended interactive trajectories rather than merely selecting locally plausible actions.

\paragraph{Task Setting.}
The original DeliveryBench benchmark is built around a profit-earning courier task in urban environments, with diverse road layouts, functional locations, transportation options, and realistic resource dynamics. At each step, the agent must reason over its current operational state, currently active orders, spatial context, and delivery constraints, then choose the next executable delivery action. In our framework, DeliveryBench is instantiated as a long-horizon single-agent environment under the same high-level objective of maximizing delivery profit. For the main study in this paper, we evaluate agent modules under the 1-hour setting, which provides a controlled testbed for comparing reasoning, memory, and reflection designs in a realistic resource-constrained scenario.

\paragraph{Evaluation Protocol.}
Within AGENTFACTORY, DeliveryBench is executed through the unified runner rather than through a benchmark-specific standalone pipeline. Episodes are evaluated over long-horizon interactions and terminate when benchmark lifecycle limits are reached or when the framework step budget is exhausted. This setup preserves the underlying DeliveryBench objective while allowing all agent variants to be compared under the same top-level execution interface. We use this shared framework to isolate the contribution of modular components without changing the surrounding evaluation pipeline.

\paragraph{Metrics.}
Following the benchmark goal, the main metric reported in the paper is \emph{hourly profit}, which directly measures how effectively an agent converts long-horizon decisions into net delivery earnings. In the main DeliveryBench table, we report the mean hourly profit over three independent runs for each agent configuration. In addition to this headline metric, the framework-side evaluator also retains a normalized \emph{score} derived from hourly profit and can record finer-grained delivery diagnostics when available, such as order-quality, time-efficiency, on-time, and resource-related indicators. In the main text, however, we use hourly profit as the primary metric because it best captures the benchmark’s end objective and provides the most interpretable comparison across module combinations.

\paragraph{Unified Agent Input.}
To support cross-benchmark modularity, we expose DeliveryBench to the agent through a unified input interface rather than passing the original benchmark prompt format directly to each reasoning method. Concretely, the perception layer organizes the current observation into four conceptual components: \emph{Image}, \emph{Instruction}, \emph{State Info}, and \emph{Action Schema}. The \emph{Instruction} component encodes the benchmark-level courier objective and task context. The \emph{State Info} component repackages benchmark-native textual context into a structured agent-facing representation, including the current operational status, active orders, map-related context, recent actions, and recent execution feedback. When visual rendering is available, the \emph{Image} component provides environment images through the same shared visual interface used by other embodied benchmarks. The \emph{Action Schema} component normalizes benchmark actions into a standardized framework-facing action list with textual descriptions, so downstream reasoning modules interact with DeliveryBench through the same abstract interface used for other environments.

\paragraph{Input/Memory Interface.}
This unified representation is designed to stay compatible with the original DeliveryBench philosophy, which combines persistent task instructions with dynamic state and map context, while still making benchmark-specific content consumable by shared framework modules. Importantly, in our framework, external memory and cross-step reasoning history are not treated as unconditional parts of the base DeliveryBench observation. Retrieved memory is injected only when a memory module is enabled, and explicit reasoning history is included only when the corresponding history mechanism is turned on. This separation is important for fair modular comparison: it ensures that memory-augmented agents receive additional context because of their selected module design, rather than because the benchmark input itself differs across methods.

\paragraph{Action Representation.}
Finally, although DeliveryBench defines its own native action API, AGENTSPEC re-expresses these actions through a unified action abstraction. This preserves benchmark semantics at the execution level while allowing the same reasoning, memory, and reflection modules to operate across DeliveryBench, ALFRED, MiniGrid, and RoboTHOR under a common interface. As a result, differences in performance can be attributed more directly to module design rather than to environment-specific prompt engineering or ad hoc wrapper logic.}

{\let\section\subsection
\let\subsection\subsubsection
\section{Minigrid}
\label{settings:minigrid}
MiniGrid\cite{chevalier2023minigrid} is a symbolic grid-world benchmark for evaluating basic embodied decision-making. Unlike realistic 3D environments, MiniGrid uses compact grid layouts, discrete object types, and a small action space, making it useful for testing whether an agent can follow instructions, navigate, interact with objects, and solve simple compositional tasks under partial observability.

We construct an evaluation suite in MiniGrid by selecting ten tasks with diverse structures and difficulty levels, covering navigation, object interaction, and compositional reasoning.
The selected tasks are as follows:
(i) \texttt{empty\_6x6\_seed42},
(ii) \texttt{goto\_object\_6x6\_n2\_seed7},
(iii) \texttt{goto\_door\_5x5\_seed123},
(iv) \texttt{fetch\_5x5\_n2\_seed99},
(v) \texttt{simple\_crossing\_s9n1\_seed42},
(vi) \texttt{lava\_crossing\_s9n1\_seed123},
(vii) \texttt{multiroom\_n2s4\_seed42},
(viii) \texttt{lava\_gap\_s5\_seed42},
(ix) \texttt{four\_rooms\_seed42},
and (x) \texttt{simple\_crossing\_s9n2\_seed123}.

The agent receives first-person RGB observations together with a textual environment prompt as input. For all reasoning, memory, and reflection methods, we adopt the default configurations from our codebase without additional tuning. Each episode is executed with a maximum step budget of 100.
}

{\let\section\subsection
\let\subsection\subsubsection
\section{Alfred}
\label{settings:alfred}
ALFRED\cite{shridhar2020alfred} is a simulated household benchmark for long-horizon embodied task completion. It requires an agent to execute natural-language instructions of everyday activities in realistic indoor scenes.

We construct a lightweight evaluation suite from the ALFRED dataset by selecting seven representative tasks, each covering a distinct task type to ensure diversity in object manipulation, multi-object interaction, state-dependent reasoning, and long-horizon planning.

The selected tasks are:
(i) place a bat on the bed,
(ii) put two pillows on the sofa,
(iii) put a heated apple in the fridge,
(iv) cool bread and place it on the countertop,
(v) clean a mug and place it on the coffee machine,
(vi) examine a pencil under a desk lamp,
and (vii) place a box containing keys on a chair.

These tasks collectively cover major ALFRED categories, including pick-and-place, multi-object manipulation, heating, cooling, cleaning, lighting-based inspection, and movable receptacles.

The agent receives first-person RGB observations along with a structured list of visible objects within a 1.5\,m egocentric range to aid perception and grounding. The maximum step budget is 200. All methods adopt default configurations without additional tuning.}

{\let\section\subsection
\let\subsection\subsubsection
\subsection{RoboTHOR}
\label{settings:robothor}

RoboTHOR is a 3D indoor ObjectNav benchmark in which an agent navigates from a first-person viewpoint to a specified target object category. Each episode provides an initial agent pose and a target category, and the agent must navigate using a discrete action space:
\begin{quote}\small\ttfamily
\{\texttt{MoveAhead}, \texttt{RotateLeft}, \texttt{RotateRight}, \texttt{LookUp},
\texttt{LookDown}, \texttt{Stop}\}.
\end{quote}
Following standard ObjectNav evaluation, the task uses stop-based success: the agent must explicitly issue \texttt{Stop}, and the target object must be visible within the configured success threshold. We report success rate (SR) and SPL, where SPL additionally accounts for path efficiency.

For our evaluation, we construct a compact but diverse RoboTHOR test suite by manually selecting 10 episodes from the validation set. The selected episodes cover different target object categories and navigation difficulties, ranging from targets located in the same room as the agent to targets requiring cross-room navigation and broader exploration. This design allows us to evaluate both local navigation behavior and longer-horizon exploration under a controlled number of episodes. We set the maximum episode length to 200 steps, the camera field of view to \(90^\circ\), and the rotation angle for \texttt{RotateLeft} and \texttt{RotateRight} to \(90^\circ\).}

\section{Detailed Experiment Analysis}
\label{sec:detailed_experiment_analysis}

\begin{table*}[t]
\centering
\small
\setlength{\tabcolsep}{3.5pt}
\renewcommand{\arraystretch}{1.15}
\caption{Distribution of failure categories across benchmarks. Values are percentages within each benchmark.}
\label{tab:failure_taxonomy_by_benchmark}
\resizebox{\textwidth}{!}{
\begin{tabular}{l|rrr|rrrrr|rrrr}
\toprule
\multirow{3}{*}{\textbf{Benchmark}}
& \multicolumn{3}{c|}{\textbf{Parse}}
& \multicolumn{5}{c|}{\textbf{Understanding}}
& \multicolumn{4}{c}{\textbf{Visual Grounding}} \\
\cmidrule(lr){2-4}
\cmidrule(lr){5-9}
\cmidrule(lr){10-13}
& \multicolumn{1}{c}{\textbf{Unstr.}}
& \multicolumn{2}{c|}{\textbf{Action Interface}}
& \multicolumn{2}{c}{\textbf{State Confusion}}
& \multicolumn{3}{c|}{\textbf{Knowledge Deficiency}}
& \multicolumn{2}{c}{\textbf{Object Grounding}}
& \multicolumn{2}{c}{\textbf{Spatial Grounding}} \\
\cmidrule(lr){2-2}
\cmidrule(lr){3-4}
\cmidrule(lr){5-6}
\cmidrule(lr){7-9}
\cmidrule(lr){10-11}
\cmidrule(lr){12-13}
& \textbf{Memory}
& \textbf{Halluc.}
& \textbf{Param.}
& \textbf{Missing Hist.}
& \textbf{Hist. Misund.}
& \textbf{Rule Unfam.}
& \textbf{Numerical}
& \textbf{Strategy}
& \textbf{Misclass.}
& \textbf{Halluc.}
& \textbf{Stuck}
& \textbf{Distance} \\
\midrule
RoboTHOR
& 0.0 & 0.0 & 0.0
& 12.5 & 6.2
& 0.0 & 0.0 & 0.0
& 9.4 & 12.5
& 31.2 & 28.1 \\

ALFRED
& 2.1 & 0.0 & 10.4
& 10.4 & 43.8
& 0.0 & 0.0 & 10.4
& 0.0 & 0.0
& 14.6 & 8.3 \\

DeliveryBench
& 7.6 & 3.0 & 22.7
& 0.0 & 10.6
& 7.6 & 21.2 & 27.3
& 0.0 & 0.0
& 0.0 & 0.0 \\

MiniGrid
& 11.1 & 0.0 & 0.0
& 13.3 & 28.9
& 33.3 & 0.0 & 13.3
& 0.0 & 0.0
& 0.0 & 0.0 \\
\bottomrule
\end{tabular}
}
\end{table*}

{\let\section\subsection
\let\subsection\subsubsection
\section{Delivery Bench}
\label{casestudy:deliverybench}
\subsection{Complete Experimental Results}
\label{complete_results_deliverybench}

This subsection provides the detailed experimental results that support the DeliveryBench analysis in Appendix~\ref{casestudy:deliverybench}. Tables~\ref{tab:complete_deliverybench_main_gpt5}--\ref{tab:complete_deliverybench_main_qwen35_9b_nonthinking} report the full configuration-level results for each evaluated backbone, including the effects of reasoning, memory, and reflection modules on hourly profit, token usage, thinking time, steps, and cost. Table~\ref{tab:complete_deliverybench_main_qwen_multiscale} further summarizes the multi-scale Qwen comparison, making it possible to inspect how the same agentic components behave across model sizes.

\begin{table*}[t]
\centering
\resizebox{\textwidth}{!}{%
\begin{tabular}{lll|cccccccc}
\toprule
\textbf{\thead{Reasoning}} & \textbf{\thead{Memory}} & \textbf{\thead{Reflection}} & \textbf{\thead{Mean Hourly\\Profit}} & \textbf{\thead{Input Tokens\\per Step}} & \textbf{\thead{Output Tokens\\per Step}} & \textbf{\thead{Total\\Thinking Time}} & \textbf{\thead{Steps}} & \textbf{\thead{Total Cost\\(unit: US\$)}}\\
\midrule
None        & Base     & None        & 21.54 & 41840.49 &	2016.56	&	9140.89	&	184	&13.33 \\
\midrule
ReAct       & Base     & None        & 30.39 & 40758.07	&1634.20&	7882.73&	193	&12.99\\
ReAct       & ChatDB     & None      & 40.92 &37594.08&	1380.78&	8609.88&	249&	15.14\\
ReAct       & DC         & None      & 53.17& 21554.04	&5524.59&	17349.89	&160	&13.15\\
ReAct       & SimpleMem  & None      & 33.98 & 9716.87&	2622.27&	10904.65&	249&	9.55\\
ReAct       & MemoryBank & None      & 44.47 & 12778.68	&2035.78&	20008.85&	607	&22.05\\
ReAct       & OpenClaw   & None      & 55.19 & 16016.61&	3347.32	&10487.72&	300&	16.05\\
\midrule
CoT         & Base     & None        & 34.74 & 38832.02	&1848.47&	8868.42&	193&	12.94\\
Plan\&Solve & Base     & None        & 41.29 & 38140.68	&2249.70&	11676.04&	234&	16.42\\
MAD         & Base     & None        & 31.12 & 86443.24	&15298.05&	28846.13&	129	&33.54\\
\midrule
ReAct       & Base     & Self-Refine & 32.74 & 68964.44	&2791.49&	12699.32&	201&	22.94\\
ReAct       & Base     & Reflexion   & 40.54    & 41173.95	&1769.14&	8817.86	&176	&12.16\\
\bottomrule
\end{tabular}%
}
\caption{Complete Main results on \textsc{DeliveryBench} using GPT-5 (Default Mode) as backbone.\label{tab:complete_deliverybench_main_gpt5}}
\end{table*}

\begin{table*}[t]
\centering
\resizebox{\textwidth}{!}{%
\begin{tabular}{lll|cccccccc}
\toprule
\textbf{\thead{Reasoning}} & \textbf{\thead{Memory}} & \textbf{\thead{Reflection}} & \textbf{\thead{Mean Hourly\\Profit}} & \textbf{\thead{Input Tokens\\per Step}} & \textbf{\thead{Output Tokens\\per Step}} & \textbf{\thead{Total\\Thinking Time}} & \textbf{\thead{Steps}} & \textbf{\thead{Total Cost\\(unit: US\$)}}\\
\midrule
None        & Base     & None        & 13.73 & 50899.64&	17.73&	3347.50	&92&	2.35 \\
\midrule
ReAct       & Base     & None        & 31.53 & 71241.82&	176.66&	3619.69&	115&	4.16\\
ReAct       & ChatDB     & None      & 10.96 &55040.80&	138.29&	2618.27&	93	&2.60\\
ReAct       & DC         & None      & 28.62& 41466.25&	1340.80	&5329.04&	159&	3.94\\
ReAct       & SimpleMem  & None      & 30.38 & 21916.47	&459.33	&5200.82	&132	&1.63\\
ReAct       & MemoryBank & None      & 22.75 & 50992.29&	296.89	&4101.69	&98	&2.59\\
ReAct       & OpenClaw   & None      & 22.84 & 17496.32	&915.42&	3361.19&	150&	1.72\\
\midrule
CoT         & Base     & None        & 30.35 & 47747.24	&173.15	&3178.05&	126&	3.07\\
Plan\&Solve & Base     & None        & 19.52 & 58301.31&	287.05	&2674.98&	89	&2.67\\
MAD         & Base     & None        & 29.32 & 348472.75	&2648.42&	4546.61&	95	&17.31\\
\midrule
ReAct       & Base     & Self-Refine & 15.87 &147535.58	&699.06	&3223.03&	89&	6.75\\
ReAct       & Base     & Reflexion   & 32.93    &51040.20	&137.45	&1139.87	&172&	4.46\\
\bottomrule
\end{tabular}%
}
\caption{Complete Main results on \textsc{DeliveryBench} using Gemini-3-flash (Default Mode) as backbone.\label{tab:complete_deliverybench_main_gemini_3_flash}}
\end{table*}

\begin{table*}[t]
\centering
\resizebox{\textwidth}{!}{%
\begin{tabular}{lll|cccccccc}
\toprule
\textbf{\thead{Reasoning}} & \textbf{\thead{Memory}} & \textbf{\thead{Reflection}} & \textbf{\thead{Mean Hourly\\Profit}} & \textbf{\thead{Input Tokens\\per Step}} & \textbf{\thead{Output Tokens\\per Step}} & \textbf{\thead{Total\\Thinking Time}} & \textbf{\thead{Steps}} & \textbf{\thead{Total Cost\\(unit: US\$)}}\\
\midrule
None        & Base     & None        & 13.83 & 41664.22&	2039.69	&6831.88&	160&	3.36 \\
\midrule
ReAct       & Base     & None        & 26.97 & 33584.03&	1168.77&	4256.08	&202&	3.20\\
ReAct       & ChatDB     & None      & 17.63 &42070.07&	1406.14&	2955.90	&98&	1.93\\
ReAct       & DC         & None      & 36.83&31378.00	&4596.04	&14664.56	&175&	4.02\\
ReAct       & SimpleMem  & None      & 20.25 & 15405.08	&2670.03&	9979.89&	137&	1.68\\
ReAct       & MemoryBank & None      & 26.69 & 40713.80	&2703.88&	6617.40	&173&	3.84\\
ReAct       & OpenClaw   & None      & 19.56 & 13292.27	&3571.23&	5559.87	&126	&1.71\\
\midrule
CoT         & Base     & None        & 20.70 & 49864.86	&2025.82	&3450.75	&93	&2.25\\
Plan\&Solve & Base     & None        & 20.76 &54603.72	&3400.94&	5382.23	&98	&2.87\\
MAD         & Base     & None        & 28.76 &106888.80	&17538.46&	28492.09&	105&	8.69\\
\midrule
ReAct       & Base     & Self-Refine & 26.44 &110300.93&	5965.66	&11597.13&	115	&6.55\\
ReAct       & Base     & Reflexion   & 33.58 &	47187.86&	1800.58&   3994.62&	164&	3.71\\
\bottomrule
\end{tabular}%
}
\caption{Complete Main results on \textsc{DeliveryBench} using Qwen3.5-397B (Thinking Mode) as backbone.\label{tab:complete_deliverybench_main_qwen35_397b}}
\end{table*}

\begin{table*}[t]
\centering
\resizebox{\textwidth}{!}{%
\begin{tabular}{lll|cccccccc}
\toprule
\textbf{\thead{Reasoning}} & \textbf{\thead{Memory}} & \textbf{\thead{Reflection}} & \textbf{\thead{Mean Hourly\\Profit}} & \textbf{\thead{Input Tokens\\per Step}} & \textbf{\thead{Output Tokens\\per Step}} & \textbf{\thead{Total\\Thinking Time}} & \textbf{\thead{Steps}} & \textbf{\thead{Total Cost\\(unit: US\$)}}\\
\midrule
None        & Base     & None        & 7.13 &	31554.64	&2677.12& 18044.63&	194&	0.38 \\
\midrule
ReAct       & Base     & None        & 12.64 	&28232.46	&483.99	& 2086.66&164&	0.24\\
ReAct       & ChatDB     & None      & 19.84 	&56136.51	&1506.71&3117.69&	86	&0.26\\
ReAct       & DC         & None      & 14.95	&29991.75	&5971.91&10073.67	&108&	0.04\\
ReAct       & SimpleMem  & None      & 9.59 	&9081.73	&2007.32&4198.98	&84	&0.03\\
ReAct       & MemoryBank & None      & 7.49 	&18605.72&	1479.27&3985.16	&169	&0.18\\
ReAct       & OpenClaw   & None      & 18.22 	&16473.94&	10950.63& 8985.50	&110	&0.05\\
\midrule
CoT         & Base     & None        & 12.62 	&50844.34	&1851.30& 4012.57&	92&	0.26\\
Plan\&Solve & Base     & None        & 20.91 	&66762.94	&5704.94&13543.98	&86&	0.36\\
MAD         & Base     & None        & 23.98 	&209013.17	&37683.06&29573.14	&48	&0.77\\
\midrule
ReAct       & Base     & Self-Refine & 18.36 	&125377.87	&8326.56&17113.55&	109&	0.32\\
ReAct       & Base     & Reflexion   & 14.69 	&45238.34	&847.98&2729.8	&89	&0.21\\
\bottomrule
\end{tabular}%
}
\caption{Complete Main results on \textsc{DeliveryBench} using Qwen3.5-9B (Thinking Mode) as backbone.\label{tab:complete_deliverybench_main_qwen35_9b}}
\end{table*}

\begin{table*}[t]
\centering
\resizebox{\textwidth}{!}{%
\begin{tabular}{lll|cccccccc}
\toprule
\textbf{\thead{Reasoning}} & \textbf{\thead{Memory}} & \textbf{\thead{Reflection}} & \textbf{\thead{Mean Hourly\\Profit}} & \textbf{\thead{Input Tokens\\per Step}} & \textbf{\thead{Output Tokens\\per Step}} & \textbf{\thead{Total\\Thinking Time}} & \textbf{\thead{Steps}} & \textbf{\thead{Total Cost\\(unit: US\$)}}\\
\midrule
None        & Base     & None        & 7.13 &40281.28&	709.14	&	2569.25&147	&1.68	 \\
\midrule
ReAct       & Base     & None        & 12.64 &	40734.34&	765.07&6316.57	&95&	1.11	\\
ReAct       & ChatDB     & None      & 19.84 &	41584.49&	863.24&6432.34	&90	&1.09\\
ReAct       & DC         & None      & 14.95 &	35238.95	&5335.59&8067.92	&105	&2.04	\\
ReAct       & SimpleMem  & None      & 9.59 &	15408.48&	1966.28&7463.24&	144	&1.13	\\
ReAct       & MemoryBank & None      & 7.49 	&17436.64	&1436.35&2415.27	&152	&1.10	\\
ReAct       & OpenClaw   & None      & 18.22 &19832.97&	2294.61&3847.01	&	145&	1.38	\\
\midrule
CoT         & Base     & None        & 12.62 	&40559.10	&1147.88&6929.19	&93&	1.15	\\
Plan\&Solve & Base     & None        & 20.91 	&39309.43	&1448.07&5978.33&	70	&0.89	\\
MAD         & Base     & None        & 23.98 	&86828.71&	8970.89&22950.70&	188&	7.46	\\
\midrule
ReAct       & Base     & Self-Refine & 18.36 &80186.41	&2021.08&5999.25		&139	&3.34	\\
ReAct       & Base     & Reflexion   & 14.69 	&40292.11&	1089.82&2310.70&	111	&1.36	\\
\bottomrule
\end{tabular}%
}
\caption{Complete Main results on \textsc{DeliveryBench} using GPT-5 mini (Default Mode) as backbone.\label{tab:complete_deliverybench_main_gpt5_mini}}
\end{table*}

\begin{table*}[t]
\centering
\resizebox{\textwidth}{!}{%
\begin{tabular}{lll|cccccc}
\toprule
\textbf{\thead{Reasoning}} & \textbf{\thead{Memory}} & \textbf{\thead{Reflection}} & \textbf{\thead{Mean Hourly\\Profit}} & \textbf{\thead{Input Tokens\\per Step}} & \textbf{\thead{Output Tokens\\per Step}} & \textbf{\thead{Total\\Thinking Time}} & \textbf{\thead{Steps}} & \textbf{\thead{Total Cost\\(unit: US\$)}}\\
\midrule
None        & Base     & None        & 8.27	&48322.76	&391.29	&2492.18	&		91	&1.80 \\
\midrule
ReAct       & Base     & None        & 9.49&45450.88&	615.67	&	3636.11	&		113	&2.17\\
ReAct       & ChatDB     & None      & 8.21&	47557.91	&241.84	&	4862.25	&		274	&5.24\\
ReAct       & DC         & None      &7.00 & 39119.26 &	2227.61&	8516.99		&	155	&3.17	\\
ReAct       & SimpleMem  & None      & 11.32&17963.88&	767.85	&	3693.18	&		81	&0.71	\\
ReAct       & MemoryBank & None      & 4.07	&19871.15	&522.38	&	3588.51		&	94	&0.84	\\
ReAct       & OpenClaw   & None      & 8.27	&17619.51&	699.93	&	3441.35		&	123	&1.05	\\
\midrule
CoT         & Base     & None        & 22.98	&46730.31&	329.82	&	3309.03		&	111	&2.11	\\
Plan\&Solve & Base     & None        & 10.43	&45745.94	&665.19	&	4740.31		&	176	&3.41\\
MAD         & Base     & None        & 25.57	&102908.56	&2901.98	&	7452.84		&	94&	4.41\\
\midrule
ReAct       & Base     & Self-Refine & 20.21&106277.74	&821.23	&	7213.43		&	220	&9.54	\\
ReAct       & Base     & Reflexion   & 27.88	& 45527.11	&210.46		&1118.7		&	126&	2.30	\\
\bottomrule
\end{tabular}%
}
\caption{Complete Main results on \textsc{DeliveryBench} using Qwen3.5-397B (Non-Thinking Mode) as backbone.\label{tab:complete_deliverybench_main_qwen35_397b_nonthinking}}
\end{table*}

\begin{table*}[t]
\centering
\resizebox{\textwidth}{!}{%
\begin{tabular}{lll|cccccc}
\toprule
\textbf{\thead{Reasoning}} & \textbf{\thead{Memory}} & \textbf{\thead{Reflection}} & \textbf{\thead{Mean Hourly\\Profit}} & \textbf{\thead{Input Tokens\\per Step}} & \textbf{\thead{Output Tokens\\per Step}} & \textbf{\thead{Total\\Thinking Time}} & \textbf{\thead{Steps}} & \textbf{\thead{Total Cost\\(unit: US\$)}}\\
\midrule
None        & Base     & None        & -4.12&	57345.13&	3884.95	&8759.12	&230&	0.79\\
\midrule
ReAct       & Base     & None        & 10.30	&52362.63&	4393.13&	6840.38	&156&	0.51\\
ReAct       & ChatDB     & None      &3.31&	54706.84	&3895.02&	6565.23	&167&	0.55\\
ReAct       & DC         & None      &5.73	&39407.46	&9637.76&	15257.33	&199	&0.68	\\
ReAct       & SimpleMem  & None      &-0.89&	26623.58&	15100.31&	13800.34&	123	&0.44	\\
ReAct       & MemoryBank & None      &8.93	&28000.59	&4492.18&	18146.65	&483&	1.00	\\
ReAct       & OpenClaw   & None      &3.26	&21791.08	&10267.34&	8938.65&	154&	0.40	\\
\midrule
CoT         & Base     & None        & 5.67	&53236.66	&4021.51&	5536.89&	136	&0.44	\\
Plan\&Solve & Base     & None        &3.56&	54375.97	&4434.70&	8459.03	&191&	0.65\\
MAD         & Base     & None        & 17.45	&120098.96&	34562.97	&28549.81	&117	&1.31\\
\midrule
ReAct       & Base     & Self-Refine &-0.09	&130286.06	&14778.62	&21065.27	&173&	1.51	\\
ReAct       & Base     & Reflexion   &14.68	&49368.82	&3132.93	&4240.44	&152	&0.45	\\
\bottomrule
\end{tabular}%
}
\caption{Complete Main results on \textsc{DeliveryBench} using Qwen3.5-9B (Non-Thinking Mode) as backbone.\label{tab:complete_deliverybench_main_qwen35_9b_nonthinking}}
\end{table*}

\begin{table*}[t]
\centering
\resizebox{\textwidth}{!}{%
\begin{tabular}{lll|cccc}
\toprule
\textbf{\thead{Reasoning}} & \textbf{\thead{Memory}} & \textbf{\thead{Reflection}} & \textbf{\thead{Qwen3.5-27B\\Mean Hourly Profit}} & \textbf{\thead{Qwen3.5-9B\\Mean Hourly Profit}} & \textbf{\thead{Qwen3.5-2B\\Mean Hourly Profit}} & \textbf{\thead{Qwen3.5-0.8B\\Mean Hourly Profit}}\\
\midrule
None        & Base       & None        & 28.89 & 7.13 & -8.54 & -2.89\\
\midrule
ReAct       & Base       & None        & 18.57 & 12.64 & -9.91 & -4.55\\
ReAct       & ChatDB     & None        & 28.86 & 19.84 & -9.47 & -7.47\\
ReAct       & DC         & None        & 17.44 & 14.95 & -9.61 & -11.18\\
ReAct       & SimpleMem  & None        & 21.45 & 9.59 & -6.86 & -8.97\\
ReAct       & MemoryBank & None        & 34.70 & 7.49 & -9.81 & -9.09\\
ReAct       & OpenClaw   & None        & 31.12 & 18.22 & -8.72 & -9.99\\
\midrule
CoT         & Base       & None        & 25.62 & 12.62 & -7.63 & -9.65\\
CoT         & ChatDB     & None        & 21.32 & 11.20 & -- & 0\\
CoT         & DC         & None        & 33.64 & 17.17 & -- & -9.64\\
CoT         & SimpleMem  & None        & 26.48 & 3.16 & -8.61 & -2.05\\
CoT         & MemoryBank & None        & 41.71 & 19.48 & -7.47 & -4.15\\
Plan\&Solve & Base      & None        & 19.42 & 20.91 & -4.65 & -7.01\\
Plan\&Solve & ChatDB    & None        & 19.20 & 11.32 & -- & -8.14\\
Plan\&Solve & DC        & None        & 21.78 & 9.43 & -- & -8.92\\
Plan\&Solve & SimpleMem & None        & 24.10 & 8.55 & -7.15 & -4.89\\
Plan\&Solve & MemoryBank & None       & 34.73 & 15.69 & -5.97 & -6.88\\
MAD         & Base       & None        & 26.38 & 23.98 & -9.26 & -9.89\\
MAD         & ChatDB     & None        & 31.16 & 29.80 & -- & -9.52\\
MAD         & DC         & None        & 27.64 & 30.01 & -- & -9.15\\
MAD         & SimpleMem  & None        & 25.91 & 21.67 & -- & -9.41\\
MAD         & MemoryBank & None        & 33.99 & 26.94 & -4.35 & -9.18\\
\midrule
ReAct       & Base       & Self-Refine & 25.05 & 18.36 & -8.04 & 0.00\\
CoT         & Base       & Self-Refine & 23.71 & 11.95 & -- & 0\\
CoT         & SimpleMem  & Self-Refine & 30.07 & 13.43 & -- & 0\\
Plan\&Solve & Base      & Self-Refine & 22.02 & 11.19 & -- & -6.77\\
MAD         & MemoryBank & Self-Refine & 35.29 & 12.18 & -- & -2.5\\
\midrule
ReAct       & Base       & Reflexion   & 27.41 & 14.69 & -20.06 & -4.54\\
ReAct       & DC         & Reflexion   & 31.04 & 17.28 & -9.92 & -9.40\\
ReAct       & MemoryBank & Reflexion   & 29.02 & 12.65 & -8.66 & -9.99\\
CoT         & Base       & Reflexion   & 25.70 & 11.09 & -- & -7.1\\
CoT         & SimpleMem  & Reflexion   & 35.75 & 17.76 & -- & -7.32\\
Plan\&Solve & Base      & Reflexion   & 18.15 & 7.78 & -- & -9.13\\
MAD         & MemoryBank & Reflexion   & 39.45 & 31.04 & -- & -9.98\\
\bottomrule
\end{tabular}%
}
\caption{Complete main results on \textsc{DeliveryBench} for Qwen3.5-27B, Qwen3.5-9B, Qwen3.5-2B, and Qwen3.5-0.8B. All entries report mean hourly profit only. Some method combinations were not evaluated due to time and resource constraints, and their missing results are marked as \texttt{--}.\label{tab:complete_deliverybench_main_qwen_multiscale}}
\end{table*}

\subsection{Case Study: Planning-Based Reasoning Prefers Abstracted Memory}
\label{app:case_planning_memory}

\textbf{Thesis.}
Among all strategy--memory combinations, Plan-and-Solve shows the largest memory-induced performance gap: average profit rises from \$6.18 with \texttt{simple} to \$17.43 with \texttt{simplemem} ($\Delta=+\$11.25$). We trace this mechanism through step~15 in \texttt{0000\_medium\_city\_22roads\_seed42}, where the two memory formats drive qualitatively different reasoning under comparable pressure.

\subsubsection*{Case A: Raw-History Memory Causes Context Pollution (Plan-and-Solve + Simple)}

\textbf{Source run:} \texttt{gpt-5-mini\_plan\_and\_solve\_simple \_none\_1} \\
\textbf{Environment:} \texttt{0000\_medium\_city\_22roads \_seed42} \\
\textbf{Key step:} 15

\paragraph{Task \& State.}
Agent~5 is riding an e-scooter at $(212.00\text{m},\;-362.16\text{m})$ with 73\% energy and 60\% battery (range: $1{,}500\,\text{m}$). It holds one active order: Order~\#0 (pickup at $(-522.59\text{m},\;221.25\text{m})$, drop-off at $(223.09\text{m},\;-421.78\text{m})$, payout \$11.29, status \texttt{Ready for pickup}, and only \textbf{4\,min} remaining). The agent is 65.7\,m from the drop-off but 1{,}323.0\,m from the pickup. It accepted this order at step~12 (when the time limit was 17\,min), then spent intervening steps charging and relocating instead of heading to the restaurant.

\paragraph{Memory.}
\textbf{Raw memory excerpt from \texttt{Memory} at step~15:}
\begin{quote}\small\ttfamily
\{'step':'11', 'state':'\#\# DeliveryBench Instructions ... \#\# map\_snapshot ... [50+ POIs] ... \#\# recent\_actions ...', 'action':'VIEW\_ORDERS()', ...\}\\
\{'step':'12', 'state':'\#\# DeliveryBench Instructions ... [full prompt repeated] ...', 'action':'ACCEPT\_ORDER(0)', ...\}\\
\{'step':'13', 'state':'\#\# DeliveryBench Instructions ... [full prompt repeated again] ...', 'action':'MOVE(-522.59m, 221.25m)', ...\}
\end{quote}

\textbf{Extracted takeaway.}
The memory is not sparse; it is over-complete. Each entry replays large prompt scaffolding (instructions, map, and state context), so the issue is duplicated context rather than missing task signals.

\paragraph{Analysis.}
In the model's ``Solve'' section, it states:
\begin{quote}
``Riding to the pickup (1.323\,km) takes \textasciitilde3.7 minutes on the scooter, which is the fastest feasible path---so I should depart immediately to make pickup and delivery within the time limit.''
\end{quote}
This acknowledges the 3.7\,min ride to pickup but omits the downstream pipeline: \texttt{PICKUP}, \texttt{PLACE\_FOOD\_IN\_BAG}, and a further 1{,}388.7\,m delivery leg ($\approx$3.9\,min at 6\,m/s). The true minimum completion time is $\geq$7.6\,min; with only 4\,min remaining the order is physically undeliverable. Nevertheless, the agent issues \texttt{MOVE(-522.59m,\;221.25m)} and commits to a doomed trajectory.

\subsubsection*{Case B: Narrative Memory Enables Resource-Aware Planning (Plan-and-Solve + Simplemem)}

\textbf{Source run:} \texttt{gpt-5-mini\_plan\_and\_solve \_simplemem\_none\_1} \\
\textbf{Environment:} \texttt{0000\_medium\_city\_22roads \_seed42} \\
\textbf{Key step:} 15

\paragraph{Task \& State.}
Agent~11 is on an e-scooter at $(-48.93\text{m},\;423.87\text{m})$ with 88\% energy but only 2\% battery (range: 47.2\,m). It carries one picked-up order: Order~\#1 (drop-off at $(224.64\text{m},\;-496.02\text{m})$, payout \$15.89, status \texttt{Picked up, waiting for delivery}, with \textbf{16\,min} remaining). The drop-off is 1{,}454.7\,m away, far beyond the current battery range.

\paragraph{Memory.}
\textbf{Raw memory excerpt from \texttt{Memory} at step~15 (5 narrative entries, condensed):}
\begin{quote}\small\ttfamily
- Procedural rule: after PICKUP, use PLACE\_FOOD\_IN\_BAG; temperature constraints may require heat/ice packs.\\
- Procedural rule: hand\_to\_customer may require STEP\_FORWARD/TURN\_AROUND; fallback to door drop may incur penalties.\\
- Timing hint: Order \#1 near pickup, prep ETA about 1 minute.\\
- Historical pool snapshot: prior candidate orders and item-temperature details.\\
- Acceptance record: ACCEPT\_ORDER(1) with pickup/drop-off coordinates and prep-status note.
\end{quote}

\textbf{Extracted takeaway.}
Compared with Case~A, this memory is compact and procedural: no raw JSON replay, no duplicated full prompt, and no full-map restatement.

\paragraph{Analysis.}
The model constructs a quantitative three-step plan:
\begin{enumerate}
  \item Move 15.1\,m to charging\_station~9 at $(-38.79\text{m},\;412.00\text{m})$.
  \item Charge from 2\% to 60\%: needed battery $=58\%$, time $=58/7.5\approx7.73$\,min, cost $=\$0.05\times58=\$2.90$.
  \item Ride 1{,}454.7\,m to drop-off at 6\,m/s ($\approx$4.0\,min).
\end{enumerate}
Total estimated time is about 11.7\,min, well within the 16-minute deadline. The agent executes the first step with \texttt{MOVE(-38.79m,\;412.00m)}.

\subsubsection*{Insight}
The contrast between Case~A and Case~B reveals why planning-based reasoning benefits disproportionately from abstracted memory.

\begin{itemize}
  \item In Case~A, \textbf{simple} memory re-injects full prompt blocks at recall time, consuming context budget and crowding out multi-step feasibility checks.
  \item In Case~B, \textbf{simplemem} compresses history into procedural summaries, preserving budget for battery, time, and deadline arithmetic.
\end{itemize}

The key insight is that summarized memory provides reusable policy-level guidance that better matches plan-then-act reasoning. Planning-oriented strategies like Plan-and-Solve benefit less from exact historical replay and more from distilled experience about what to do in recurring situations. The performance gap (\$6.18 $\to$ \$17.43) is consistent with this mechanism: summarized memory does not make Plan-and-Solve inherently smarter, but it prevents context overload from suppressing its planning capacity.

\subsection{Case Study: DC Memory Refines Batching into Urgency-Aware Triage}
\label{app:case_dc_memory_triage}

\textbf{Source run:} \texttt{gpt-5\_react\_dc\_none\_1} \\
\textbf{Environment:} \texttt{0001\_medium\_city\_22roads\_seed123} \\
\textbf{Key steps:} 4--10

\paragraph{Task \& State.}
This case captures a co-located multi-order state where a generic batching strategy becomes locally insufficient. At step 4, the agent is handling two active orders, \texttt{Order \#19} and \texttt{Order \#25}, from the same restaurant. However, the two orders are asymmetric: \texttt{Order \#25} is closer to readiness and has the tighter remaining time budget, while \texttt{Order \#19} is still farther from execution. The key decision is therefore not simply whether to batch, but whether to keep waiting for a cleaner joint pickup or to split execution and prioritize the more urgent ready-side order.

\paragraph{Observation.}
Across steps 4--7, the environment progressively makes this asymmetry actionable. At steps 4--6, \texttt{Order \#25} is still being prepared for about \texttt{1 min} with only \texttt{2 min} left, whereas \texttt{Order \#19} remains about \texttt{3 min} from readiness with \texttt{4 min} left. By step 7, \texttt{Order \#25} becomes \texttt{Ready for pickup} with only \texttt{1 min} remaining, while \texttt{Order \#19} is still \texttt{Food is still being prepared (~2 min)}. This turns the state into a concrete asymmetric-readiness decision point.

\paragraph{Memory.}
\textbf{Raw memory excerpt from \texttt{Memory (Recent Steps)} at step 7:}
\begin{quote}\small\ttfamily
When multiple accepted orders share a pickup location, include only the ready order IDs in PICKUP and leave not-ready ones for later.\\
When multiple active orders share a pickup location, coordinate to collect them together when both are (nearly) ready, then sequence drop-offs by tighter deadlines and proximity.\\
If co-located orders have mismatched prep ETAs and the sooner-ready order has a tight time limit, then pick up and deliver the ready order first rather than waiting for the slower one.
\end{quote}

\textbf{Extracted takeaway.}
The memory does not simply restate recent observations. Instead, it retains a general co-located batching heuristic while refining it into a sharper exception rule for asymmetric readiness and urgency.

\paragraph{Action \& Result.}
The agent follows this refinement closely. It first switches to the e-scooter at step 4, moves to the pickup door at step 5, waits through the short remaining preparation window at step 6, and then executes \texttt{PICKUP(orders=[25])} at step 7, explicitly excluding the still-unready \texttt{Order \#19}. It then completes the standard pipeline with \texttt{PLACE\_FOOD\_IN\_BAG}, \texttt{MOVE}, and \texttt{DROP\_OFF}. The outcome is not a full rescue: \texttt{task\_report.json} shows that \texttt{Order \#25} is still slightly overdue (\texttt{deadline\_slack\_s = -20.71}), while \texttt{Order \#19} is substantially later (\texttt{deadline\_slack\_s = -118.82}). Still, the agent clearly prioritizes the more urgent ready order rather than waiting for a cleaner two-order batch.

\paragraph{Analysis.}
What makes this case interesting is that the memory refinement unfolds together with the state progression rather than appearing all at once. In steps 4--5, the agent is still in a positioning phase: it switches transport and moves to the pickup door while the memory remains dominated by general heuristics such as waiting briefly for short preparation times and batching co-located orders when both are nearly ready. At step 6, once the asymmetry becomes sufficiently clear, the memory introduces a more specific exception policy for mismatched co-located orders. By step 7, this policy is further sharpened into an executable action-level rule that directly constrains the \texttt{PICKUP} argument to ready order IDs only. The interesting point is therefore not that the agent simply “remembered” to take \texttt{Order \#25} first, but that the memory transformed an initially implicit urgency judgment into a progressively more operational policy representation. This makes the subsequent action sequence (\texttt{SWITCH} $\rightarrow$ \texttt{MOVE} $\rightarrow$ \texttt{WAIT} $\rightarrow$ \texttt{PICKUP([25])}) look less like a one-off choice and more like a policy being concretized online.

\paragraph{Insight.}
This case suggests a plausible mechanism behind DC's strong performance: its memory may help not merely by storing recent events, but by converting general operational heuristics into sharper exception and action-level policies as local states become more concrete. In long-horizon delivery settings, many difficult decisions are not global replanning problems but small local conflicts of readiness, urgency, and execution timing. The evidence here suggests that DC can compress such evolving local structure into reusable procedural guidance. While this single case does not by itself explain DC's aggregate advantage, it illustrates how memory can function as an online policy-refinement mechanism rather than as passive recall alone.

\subsection{Case Study: MemoryBank Converts Past Battery Depletion into Proactive Charging Policy}
\label{app:case_memorybank_charging}

\textbf{Source run:} \texttt{gpt-5-mini\_react\_memorybank\_none\_3} \\
\textbf{Environment:} \texttt{0001\_medium\_city\_22roads\_seed123} \\
\textbf{Key steps:} 9--11

\paragraph{Task \& State.}
At step 9, the agent has just completed its first delivery (\texttt{Order \#17}) and returned to an idle state with no active orders. The critical detail is the e-scooter battery: it has dropped to \textbf{12\%} with only \textbf{310.2,m} of remaining range. The agent's personal energy is healthy (91\%), and it has earned \$112.28 so far with roughly 53 minutes remaining. The decision it faces is whether to immediately accept a new order or first address its depleted battery.

\paragraph{Observation.}
The environment provides the agent's position at \texttt{(-421.96m, 20.20m)} on 8th road (left), with the nearest charging station (\texttt{charging\_station 15}) only \textbf{46.0,m} away. The available order list (shown at step 10 after \texttt{VIEW\_ORDERS}) includes several orders whose pickup-to-dropoff distances far exceed the scooter's remaining 310,m range. The observation alone contains enough raw data for a cautious agent to infer the battery risk, but does not explicitly advise the agent to charge before accepting work.

\paragraph{Memory.}
\textbf{Raw memory excerpt from \texttt{Memory (Recent Steps)} at step 9:}
\begin{quote}\small\ttfamily
[Trajectory Summary] Scooter battery dropped from 50
\end{quote}
\begin{quote}\small\ttfamily
[Environment Insight] Scooter range drains substantially with travel: range 1250m$\to$310m over this route (\textasciitilde940m used). Final battery reached \textasciitilde12
Avoid: accepting long-distance orders without a comfortable scooter-range buffer (don't finish deliveries with battery $\leq$15
Next-episode policy (3 actions):\\
\hspace*{1em}- Before accepting, estimate total travel (move-to-pickup + pickup$\to$dropoff) and require scooter\_range\_left $\geq$ estimated\_distance + 300m buffer.\\
\hspace*{1em}- If order status shows prep\_time $\leq$ time\_left, WAIT at pickup rather than moving away.\\
\hspace*{1em}- Follow explicit delivery notes (e.g., use "call" when requested) to minimize delivery friction.
\end{quote}

\textbf{Extracted takeaway.}
The memory does not merely record what happened in the previous trajectory. It distills the experience of finishing a delivery at dangerously low battery into two concrete operational rules: (1) a 15\% battery floor below which new orders should not be accepted, and (2) a 300,m range buffer requirement over the estimated travel distance. These are not present in the system prompt or environment instructions; they are learned policies derived from the first episode's resource consumption pattern.

\paragraph{Action \& Result.}
The agent follows the memory-derived policy closely across the subsequent steps:
\begin{itemize}
\item \textbf{Step 9:} The agent recognizes its low battery and decides to first check available orders before committing: \texttt{VIEW\_ORDERS()}. Its thought explicitly notes: \emph{My e-scooter battery is low (12\%, \textasciitilde310 m range) \ldots\ I'll choose ones within range or head to a nearby charging station before accepting long trips.}''
\item \textbf{Step 10:} After reviewing the order list, the agent concludes that recharging must come first:\emph{To safely accept and complete profitable nearby orders \ldots\ I need to recharge first. The nearest charging station is charging\_station 15 at (-412.00m, 61.24m) (\textasciitilde46m away).}'' It executes \texttt{MOVE(-412.00m, 61.24m, pace="normal")}.
\item \textbf{Step 11:} Upon arriving at the charging station (battery now 10\%, range \textasciitilde259,m), the agent charges to a deliberate target: ``\emph{I'll charge the scooter to a comfortable level (60\%) to ensure I can take mid-range orders \ldots\ without risking battery depletion. Charging now also avoids wasting time going to pickup only to have to detour for charging later.}'' It executes \texttt{CHARGE(target\_pct=60)}.
\end{itemize}
The outcome is that the agent resumes order acceptance with a fully operational scooter (\textasciitilde60\% battery, \textasciitilde1500,m range), enabling it to handle subsequent deliveries without mobility constraints.

\paragraph{Analysis.}
What makes this case informative is the contrast between the agent's behavior with and without memory. In the first episode (step 1, no memory), the agent begins with 50\% battery and simply executes \texttt{VIEW\_ORDERS()} without any battery-awareness reasoning. It then accepts \texttt{Order \#17} and rides 769.8,m to the dropoff, consuming nearly all remaining range and arriving at 12\% battery---a state that would have been catastrophic had the delivery been slightly longer.

In the second episode, the same 12\% battery state triggers a qualitatively different response. The agent does not blindly accept the next available order. Instead, it explicitly references the low battery threshold, evaluates whether nearby orders are within range, and proactively moves to a charging station before accepting new work. The 60\% charging target is also notable: rather than fully charging (which would waste time at 0.05/\% and 7.5\%/min), the agent selects a target that provides sufficient range for mid-distance orders while minimizing idle time---a cost-benefit tradeoff that aligns with the memory's ``estimated\_distance + 300m buffer'' heuristic.

The key causal link is the \texttt{[Environment Insight]} entry, which converts a single episode's resource-depletion experience into a reusable policy. Without this memory, the agent would need to independently re-derive the battery management strategy from the raw observation (battery percentage, charging station distances, order distances). With it, the policy is pre-computed and directly available in the prompt, reducing the cognitive burden on the LLM and making the correct action sequence more likely.

\paragraph{Insight.}
This case illustrates MemoryBank's core mechanism: converting trajectory-level outcomes into episode-transferable operational policies. The memory system does not simply replay past observations; it generates structured \texttt{Environment Insight} entries that encode quantitative thresholds (battery $\geq$ 15\%), buffer requirements (range $\geq$ distance + 300,m), and procedural guidance (charge before accepting). These insights function as learned heuristics that shift the agent from reactive execution---accepting orders until resources are exhausted---to proactive resource management---maintaining sufficient capacity before committing to new tasks. In long-horizon delivery settings where resource depletion is a common failure mode, this kind of cross-episode policy transfer can meaningfully improve operational robustness, even when the underlying model is a smaller-capacity backbone.
\subsection{Case Study: Dynamic Memory Turns Waiting Time into Actionable Maintenance}
\label{app:case_openclaw_waittime_charging}

\textbf{Source run:} \texttt{agent\_react\_openclaw\_gemini-3-flash} \\
\textbf{Episode:} \texttt{0000\_medium\_city\_22roads\_seed42} \\
\textbf{Key steps:} 4--9

\paragraph{Task \& State.}
At step 4, the agent has already accepted \textbf{Order \#7} and \textbf{Order \#8}, both from \textbf{Restaurant 4}. It has reached the pickup location, but neither order is ready yet: Order \#7 still requires about \textbf{4 minutes}, and Order \#8 about \textbf{7 minutes}. Meanwhile, the e-scooter battery is only \textbf{44\%}. This creates a non-trivial decision point: the agent is not yet able to pick up the food, but it must decide whether to simply wait or use the preparation window for a useful intermediate action.

\paragraph{Observation.}
The environment explicitly exposes the two pending orders, their remaining preparation times, and the fact that they share the same pickup location. It also provides nearby POI information, including a \textbf{charging station} within short travel distance. In other words, the state already contains all the raw ingredients for a good decision, but the observation alone does not specify how waiting time should be operationally exploited.

\paragraph{Memory.}
Below we show a short timeline of the raw memory excerpts that are most relevant to this case.

\begin{quote}\small
\textbf{Step 4} \\
``\textbf{Order Batching}: Prioritize accepting multiple orders from the same pickup location or nearby locations (e.g., restaurant 4, order \#7 and \#8) to minimize travel time and maximize earnings per route.'' \\
``\textbf{Preparation Wait Time}: When batching orders from the same restaurant, be aware of differing food preparation times (e.g., 4 min vs 7 min); use the wait time to plan the optimal delivery sequence based on drop-off locations.''

\vspace{0.4em}
\textbf{Step 5} \\
``\textbf{Wait Time Optimization}: If food preparation time exceeds 3 minutes at a restaurant, use the window to visit nearby POIs (stores for battery packs/energy drinks or charging stations) if they are within a 1--2 minute round-trip distance.'' \\
``\textbf{Wait Time Visibility}: Food prep status (e.g., 4 min vs 7 min) is visible in the active\_orders status; checking this immediately upon arrival allows for better scheduling of mid-route maintenance tasks.''

\vspace{0.4em}
\textbf{Step 6} \\
``\textbf{Charging during Wait Times}: If food prep time is $> 3$ minutes and a charging station is within 100m, prioritize moving to the charging station to top up the e-scooter rather than idling at the restaurant.''

\vspace{0.4em}
\textbf{Step 8} \\
``\textbf{Overdue Priority}: Prioritize orders in `OVERTIME' status for immediate pickup and delivery to minimize further late penalties.'' \\
``\textbf{Pickup Readiness}: Before traveling to a restaurant, check if orders are `Ready for pickup' in the active\_orders status to avoid idling upon arrival.''

\vspace{0.4em}
\textbf{Step 9} \\
``\textbf{Pickup Readiness}: \ldots If an order is already ready while the agent is nearby, proceed to pickup immediately.''
\end{quote}

These excerpts show that OpenClaw does not only preserve static task knowledge; it also maintains and reuses increasingly action-specific guidance as the trajectory unfolds.

\paragraph{Action \& Result.}
The action sequence follows the retrieved memory closely. At \textbf{step 5}, the agent moves to the nearby charging station. At \textbf{step 6}, it executes \texttt{CHARGE(target\_pct=100)}. At \textbf{step 7}, it switches back to the e-scooter, and at \textbf{step 8}, it returns to Restaurant 4. By \textbf{step 9}, both orders are ready and already marked \texttt{OVERTIME}, so the agent immediately executes \texttt{PICKUP(orders=[7,8])}. In effect, the agent converts otherwise idle preparation time into battery maintenance, then resumes the batch with a much stronger mobility state (\textbf{97\%} battery at pickup).

\paragraph{Analysis.}
What makes this case interesting is that the benefit does not come from a single isolated memory rule. Instead, the memory evolves along the trajectory and progressively sharpens the decision process. The earlier excerpts frame the situation as a \emph{batched pickup with asymmetric preparation times}; the middle excerpts re-interpret that waiting period as an \emph{opportunity for maintenance}; and the later excerpts re-focus the agent on \emph{overdue pickup urgency} once the food is ready. In other words, the memory is not merely reminding the agent of generic facts. It is dynamically reorganizing past experience into a step-relevant operational policy.

This matters because the useful intermediate action in this case---charging---does not immediately complete the delivery objective. Its value is delayed and multi-step: the agent sacrifices a short local detour in order to improve downstream execution capacity. Without an external memory that explicitly encodes this kind of reusable operational pattern, the model would need to infer the same strategy from scratch from a noisy observation stream. OpenClaw reduces that burden by surfacing a compact, actionable policy at exactly the moment when the waiting window appears.

\paragraph{Insight.}
This case suggests that OpenClaw performs well not simply because it stores more text, but because it maintains a \textbf{persistent, dynamically updated, and action-oriented external memory store}. The retrieved content is not purely descriptive; it is structured as reusable operational guidance that can be directly mapped onto concrete delivery actions. In DeliveryBench, where strong performance depends on coordinating preparation delays, battery management, and urgency across multiple steps, such memory can turn an otherwise ambiguous waiting state into a well-structured execution plan.

\subsection{Case Study: MAD's Debate Structure Enables Error Correction and Solution Emergence}
\label{app:case_mad_debate}

\textbf{Thesis.} MAD maintains strong performance across memory modules (minimum profit: 16.63). Its debate structure provides two capabilities that single-chain reasoning often lacks: built-in error correction and solution emergence. We illustrate each with a concrete case.

\subsubsection*{Case A: Built-in Error Correction}

\textbf{Source run:} \texttt{gpt-5-mini\_mad\_simplemem\_none\_1} \\
\textbf{Environment:} \texttt{0001\_medium\_city\_22roads\_seed123} \\
\textbf{Key step:} 25

\paragraph{Task \& State.}
The agent has no active orders. It is walking at $(-374.84\text{m},\;-82.56\text{m})$ with 80\% energy. The e-scooter is parked with 0\% battery. Current earnings are \$106.66. A \texttt{recent\_error} indicates that the previous attempt to accept orders \#11 and \#13 failed (``not found or already accepted by others'').

\paragraph{Observation.}
The \texttt{recent\_actions} log shows that \texttt{View available orders} was already executed earlier in the action history. The rules explicitly state: ``DO NOT use VIEW\_ORDERS if the context already includes available order details or your last action is view orders.''

\paragraph{Memory.}
\textbf{Raw memory excerpt from \texttt{Memory} at step 25:}
\begin{quote}\small\ttfamily
- Relevant available orders (latest pool, key tradeoffs):\\
\quad - \#8: \$8.51, 18 min --- mixed hot/cold (Curry 60°C, IceCreamCake -12°C, Milkshake 5°C) → needs both heat \& ice.\\
\quad - \#0: \$8.06, 9 min --- all hot (three 60°C items) → needs heat, tighter time.\\
\quad - \#11: \$7.98, 11 min --- mixed temps (hot + frozen + cold) → mixed packs.\\
\quad - \#13: \$6.04, 8 min --- mostly cold items.\\
- Decision priorities: secure in-hand items into bag now, then accept orders matching available temperature resources and realistic time constraints.
\end{quote}

\paragraph{Debate Trace.}

\begin{itemize}
  \item \textbf{Round 0 (independent analysis).} Agent~0 proposes \texttt{VIEW\_ORDERS()}, reasoning that no active orders means it should refresh the order board. Agent~1 also proposes \texttt{VIEW\_ORDERS()}, citing the need to refresh the list after the failed acceptance of \#11 and \#13. Agent~2, however, notices that \texttt{recent\_actions} already contains a \texttt{View available orders} entry and that the memory retains the order pool; it proposes \texttt{ACCEPT\_ORDER(8)} (\$8.51, 18-minute deadline).
  \item \textbf{Round 1.} Agent~0 reads Agent~2's argument, recognizes that repeating \texttt{VIEW\_ORDERS()} violates the rules and wastes a decision step, and switches to \texttt{ACCEPT\_ORDER(8)}. Agent~1 similarly acknowledges the redundancy, noting that \texttt{recent\_error} only indicates \#11 and \#13 failed---not \#8---and also switches to \texttt{ACCEPT\_ORDER(8)}. Agent~2 maintains \texttt{ACCEPT\_ORDER(8)}.
  \item \textbf{Round 2.} All three agents converge on \texttt{ACCEPT\_ORDER(8)}.
\end{itemize}

\paragraph{Action \& Result.}
The executed action is \texttt{ACCEPT\_ORDER(8)}. The agent secures the highest-value available order without wasting a step on a redundant \texttt{VIEW\_ORDERS()} call.

\paragraph{Analysis.}
Two out of three agents initially proposed a redundant action that would have wasted an entire decision step. In a single-chain reasoning strategy such as CoT, the model has exactly one chance to make this judgment---a single oversight directly costs a step. The debate structure provides a second (and third) opportunity: Agent~2's correct reading of \texttt{recent\_actions} is surfaced to the other agents in Round~1, and the cross-examination process corrects the majority error before any action is executed. Notably, the simplemem memory module---despite its narrative format---preserved the order pool with payouts and deadlines (``\#8: \$8.51, 18 min''), giving Agent~2 sufficient information to propose a concrete alternative rather than merely objecting to the redundancy.

\subsubsection*{Case B: Solution Emergence}

\textbf{Source run:} \texttt{gpt-5-mini\_mad\_simple\_none\_1} \\
\textbf{Environment:} \texttt{0001\_medium\_city\_22roads\_seed123} \\
\textbf{Key step:} 30

\paragraph{Task \& State.}
The agent is on an e-scooter at $(-474.37\text{m},\;224.04\text{m})$ with 74\% energy and 12\% battery. It is carrying \texttt{Order~\#29} (Donut, Curry; \$4.08), which is already 1 minute overtime. The agent has arrived at the drop-off location (building~127, 22nd road). The \texttt{ephemeral\_context} presents the delivery method choice: \texttt{leave\_at\_door}, \texttt{knock}, \texttt{call}, or \texttt{hand\_to\_customer}.

\paragraph{Observation.}
The \texttt{recent\_actions} show the agent has completed the full pipeline: \texttt{MOVE} to pickup $\rightarrow$ \texttt{WAIT} 7\,min $\rightarrow$ \texttt{PICKUP} $\rightarrow$ \texttt{PLACE\_FOOD\_IN\_BAG} $\rightarrow$ \texttt{MOVE} to drop-off ($\sim$427.3\,m). The order is overtime, so the decision is purely about which delivery method balances speed against service quality.

\paragraph{Memory.}
\textbf{Raw memory excerpt from \texttt{Memory} at step 30:}
\begin{quote}\small\ttfamily
 <\#29: picked up and items placed in bag (Donut -> A; Curry -> B) --- currently carrying.\\
- You are at the drop-off (building 127) and order is OVERTIME (deliver ASAP).\\
- Transport: on e-scooter, energy 74\%, low battery (12\%) --- but already at drop-off (7 m).\\
- Must call DROP\_OFF(oid=29, method=...) now. Note: ``hand\_to\_customer'' requires STEP\_FORWARD/TURN\_AROUND/search; leaving at door may incur penalties.
\end{quote}

\paragraph{Debate Trace.}

\begin{itemize}
  \item \textbf{Round 0 (independent analysis).} Agent~0 proposes \texttt{leave\_at\_door}, prioritizing speed under overtime pressure. Agent~1 proposes \texttt{hand\_to\_customer}, prioritizing service quality and complaint avoidance. Agent~2 also proposes \texttt{hand\_to\_customer}.
  \item \textbf{Round 1.} Agent~0, seeing that two other agents chose \texttt{hand\_to\_customer}, is persuaded and switches to \texttt{hand\_to\_customer}. Agent~1, after reading Agent~0's time-pressure argument for \texttt{leave\_at\_door}, synthesizes both positions and first proposes \texttt{knock}---a method absent from all Round~0 proposals. Its reasoning: \texttt{knock} is faster than \texttt{hand\_to\_customer} (no \texttt{STEP\_FORWARD}/\texttt{TURN\_AROUND} search required) but provides better service than \texttt{leave\_at\_door} (the customer has an opportunity to receive the order in person). Agent~2, conversely, is swayed by the speed argument and switches to \texttt{leave\_at\_door}.
  \item \textbf{Round 2.} Agent~0 accepts the compromise logic and switches from \texttt{hand\_to\_customer} to \texttt{knock}. Agent~1 maintains \texttt{knock}. Agent~2 also accepts the compromise and switches from \texttt{leave\_at\_door} to \texttt{knock}. Final vote: 3--0 for \texttt{knock}.
\end{itemize}

\paragraph{Action \& Result.}
The executed action is \texttt{DROP\_OFF(oid=29, method="knock")}. The agent delivers the overtime order using a method that was not proposed by any individual agent in the initial round.

\paragraph{Analysis.}
The key observation is that \texttt{knock} was not in the initial solution set. It emerged in Round~1 as Agent~1 synthesized the competing priorities of speed (\texttt{leave\_at\_door}) and service quality (\texttt{hand\_to\_customer}). This is qualitatively different from simple majority voting or persuasion: the debate produced a novel compromise that Pareto-dominates the initial proposals along the speed--quality trade-off. A single-chain strategy would have committed to one of the two extremes at the first inference pass. The Round~1 dynamics are also noteworthy: the three agents effectively rotated positions (Agent~0 moved toward service quality, Agent~2 moved toward speed), while Agent~1 found the middle ground that ultimately convinced everyone in Round~2.

\subsubsection*{Summary}

These two cases illustrate complementary mechanisms behind MAD's robustness across memory modules. Error correction operates as a \emph{filter}: multiple reasoning paths cross-examine each other and catch individual mistakes before execution. Solution emergence operates as a \emph{generator}: the debate synthesizes novel actions that no single agent proposed initially. Together, they make MAD less sensitive to the quality or format of memory input---even when the memory module provides incomplete information (simplemem omitting resource states) or excessive raw detail (simple injecting unprocessed observations), three independent reasoning paths can each extract different useful signals and reconcile them through debate. The cost is a substantial increase in token consumption, but the result is a performance floor that exceeds the ceiling of most other strategy--memory combinations.

\subsection{Case Study: Self-Refine Converts Repeated Bagging Failures into a Structured Recovery Process}
\label{app:case_selfrefine_bag_recovery}

\textbf{Source run:} \texttt{gpt-5-mini-react-simple-self \_refine.zip/agent\_1} \\
\textbf{Environment:} \texttt{0000\_medium\_city\_22roads\_seed42} \\
\textbf{Key steps:} 42--46

\paragraph{Task \& State.}
The agent has already picked up Orders \#4 and \#5, but delivery cannot proceed because the food has not yet been successfully placed into the insulated bag. This is therefore a post-pickup recovery state rather than a normal planning state. The bag already contains a heat pack in compartment A and an ice pack in compartment B, and the agent is acting after repeated failed bagging attempts.

\paragraph{Observation.}
The prompt explicitly shows that both orders are \emph{picked up, waiting for delivery}, while the food remains unplaced. The environment also reports another failed bagging attempt:
\begin{quote}
\small
\texttt{place\_food\_in\_bag failed: invalid compartment: A, 15,16 -> B}
\end{quote}
At the same time, the bag hint lists the unresolved assignment problem: Order \#4 contains \texttt{Curry}, and Order \#5 contains \texttt{Pizza}, \texttt{PokeBowl}, and \texttt{Milkshake}, with compartment A heated and compartment B cooled.

\paragraph{Reflection.}
\textbf{Step 42 (initial correction, still fragile).}
\begin{quote}
\small
\texttt{The agent correctly diagnosed the previous failure (formatting/parsing issue) and adjusted the bag\_cmd to list each order separately. The placement respects compartment temperature and capacity.} \\
\texttt{But the chosen action are incorrect.}
\end{quote}

\textbf{Step 43 (turning point: inspect before retry).}
\begin{quote}
\small
\texttt{VIEW\_BAG() is the right next action. It will reveal the actual valid compartment IDs/slot info so the agent can craft a valid single combined bag\_cmd and avoid repeating the same failing PLACE\_FOOD\_IN\_BAG call.} \\
\texttt{... inspect bag to resolve the error, then issue a corrected PLACE\_FOOD\_IN\_BAG.}
\end{quote}

\textbf{Step 44 (decomposed recovery).}
\begin{quote}
\small
\texttt{Placing order 4 item 1 (Curry) into A is the right immediate step.} \\
\texttt{It avoids the earlier multi-order/formatting errors by issuing a simple single-order command.}
\end{quote}

\paragraph{Action \& Result.}
The agent first executes \texttt{VIEW\_BAG()} instead of retrying another complex bagging command, then decomposes the repair into two simpler actions:
\begin{quote}
\small
\texttt{Step 44: PLACE\_FOOD\_IN\_BAG(bag\_cmd="order 4: 1 -> A")} \\
\texttt{Step 45: PLACE\_FOOD\_IN\_BAG(bag\_cmd="order 5: 1 -> A; 2,3 -> B")}
\end{quote}
By step 46, both orders are successfully bagged and the agent resumes delivery toward the nearer dropoff.

\paragraph{Analysis.}
What makes this case informative is that the benefit of self-refine does not come from a single perfect correction. In step 42, the reflection already identifies the previous failure as a formatting issue and approves a revised bagging command. However, the subsequent environment state in step 43 shows that this repair is still insufficient, since the action fails again with another invalid-compartment error. This makes the case more revealing than a clean one-shot success: the model is not simply solving the problem immediately, but improving through iterative recovery.

The real turning point appears in step 43, where self-refine changes the recovery strategy itself. Instead of continuing to optimize another all-in-one bagging command, the reflection explicitly recommends an inspection step, \texttt{VIEW\_BAG()}, before retrying. This matters because the failure is no longer treated as only a textual formatting issue; it is reinterpreted as a state-alignment problem between the agent's command and the simulator's actual bag representation. In other words, self-refine helps the model realize that the next useful action is not to ``try harder,'' but to look again.

Step 44 then shows a second layer of improvement. After inspection, the agent does not return to a large combined command. Instead, it adopts a decomposed repair policy, placing one order first and postponing the second. This decomposition is mechanically valuable: it reduces parser fragility, lowers action complexity, and creates an intermediate success state that makes the next action easier. The success at steps 44--45 therefore comes not from more elaborate verbal reasoning alone, but from a better control strategy for recovery: inspect the real state, simplify the action, and restore forward progress incrementally.

\paragraph{Insight.}
This case suggests a broader explanation for why small models with self-refine can begin to approach stronger backbones in DeliveryBench. The gain is not necessarily that self-refine gives them uniformly better high-level planning. Rather, it provides an explicit error-correction layer that helps convert repeated local failures into structured recovery procedures. When tasks involve simulator-specific action syntax, hidden state details, or brittle execution interfaces, such a correction layer can substantially reduce wasted steps and prevent the agent from getting stuck in unproductive retries.

More generally, the case shows that self-refine is especially useful when the model's first attempt is locally plausible but operationally fragile. In such situations, the main advantage is not producing a more sophisticated initial plan, but detecting that the current strategy is unstable and replacing it with a more reliable one. Here, that replacement takes the form of an inspect-then-decompose policy: first recover the true state, then split a brittle composite action into simpler, more robust sub-actions. This kind of repair mechanism is a realistic and practically important route by which self-refine can narrow the performance gap between smaller and larger models.

}

{\let\section\subsection
\let\subsection\subsubsection
\section{Minigrid}
\label{casestudy:minigrid}

\subsection{Complete Experimental Results}
\label{complete_results_minigrid}

Table~\ref{tab:minigrid_suite_ablation} summarizes the complete ablation results on the MiniGrid task suite with \texttt{GPT-5-mini}. We report task-level outcomes across different combinations of reasoning strategies, memory modules, and reflection settings. Table~\ref{tab:minigrid_main_qwen_gpt5mini_multiscale} further reports success rates across model scales and backbones.

\begin{table*}[t]
\centering
\caption{
Ablation results on the MiniGrid task suite. We evaluate different combinations of reasoning modules, memory components, and reflection settings with \texttt{GPT-5-mini} as base model. Each row corresponds to one method configuration. A green checkmark indicates successful task completion, while a red cross indicates failure. Values in parentheses denote the steps used.
}
\label{tab:minigrid_suite_ablation}

\begin{adjustbox}{width=\textwidth}
\begin{tabular}{@{}l c c c c c c c c c c@{}}
\toprule
\textbf{Method} &
\makecell{\textbf{Empty}} &
\makecell{\textbf{Fetch}} &
\makecell{\textbf{FourRooms}} &
\makecell{\textbf{GoToDoor}} &
\makecell{\textbf{GoToObj}} &
\makecell{\textbf{LavaCross}} &
\makecell{\textbf{LavaGap}} &
\makecell{\textbf{MultiRoom}} &
\makecell{\textbf{SimpleCross1}} &
\makecell{\textbf{SimpleCross2}} \\
\midrule

CoT + ChatDB &
\cmark (17) & \xmark & \xmark & \xmark & \cmark (17) & \xmark & \xmark & \xmark & \xmark & \xmark \\

CoT + MemoryBank &
\cmark (19) & \xmark & \xmark & \xmark & \cmark (5) & \xmark & \cmark (7) & \xmark & \xmark & \xmark \\

CoT + OpenClaw &
\cmark (15) & \xmark & \xmark & \cmark (67) & \cmark (5) & \xmark & \cmark (35) & \xmark & \xmark & \xmark \\

CoT + Simple &
\cmark (31) & \cmark (7) & \xmark & \xmark & \cmark (9) & \xmark & \cmark (5) & \cmark (54) & \xmark & \xmark \\

CoT + SimpleMem &
\cmark (42) & \xmark & \xmark & \xmark & \cmark (73) & \xmark & \cmark (9) & \xmark & \xmark & \xmark \\

Direct + Simple &
\cmark (94) & \xmark & \xmark & \xmark & \xmark & \xmark & \cmark (5) & \cmark (47) & \xmark & \xmark \\

MAD + Simple &
\cmark (85) & \cmark (7) & \xmark & \xmark & \cmark (73) & \xmark & \cmark (5) & \cmark (97) & \cmark (100) & \xmark \\

MAD + Simple + Self-Refine &
\cmark (7) & \xmark & \xmark & \xmark & \xmark & \xmark & \xmark & \cmark (8) & \cmark (28) & \xmark \\

Plan\&Solve + Simple &
\cmark (19) & \xmark & \xmark & \xmark & \cmark (20) & \xmark & \cmark (58) & \cmark (8) & \xmark & \xmark \\

Plan\&Solve + SimpleMem &
\xmark & \xmark & \xmark & \xmark & \xmark & \xmark & \cmark (14) & \xmark & \xmark & \xmark \\

ReAct + ChatDB &
\cmark (25) & \xmark & \xmark & \xmark & \cmark (23) & \xmark & \xmark & \xmark & \xmark & \xmark \\

ReAct + DC &
\cmark (7) & \cmark (85) & \xmark & \xmark & \cmark (15) & \xmark & \cmark (5) & \cmark (10) & \xmark & \xmark \\

ReAct + DC + Self-Refine &
\cmark (7) & \xmark & \xmark & \xmark & \xmark & \xmark & \cmark (9) & \xmark & \xmark & \xmark \\

ReAct + MemoryBank &
\cmark (20) & \xmark & \xmark & \xmark & \cmark (5) & \xmark & \cmark (5) & \cmark (17) & \xmark & \xmark \\

ReAct + OpenClaw &
\cmark (49) & \xmark & \xmark & \xmark & \cmark (49) & \xmark & \cmark (29) & \xmark & \xmark & \xmark \\

ReAct + Simple &
\cmark (9) & \xmark & \xmark & \xmark & \cmark (19) & \xmark & \cmark (5) & \cmark (22) & \xmark & \xmark \\

ReAct + Simple + Self-Refine &
\cmark (7) & \xmark & \xmark & \xmark & \cmark (9) & \xmark & \cmark (7) & \cmark (10) & \xmark & \xmark \\

ReAct + SimpleMem &
\cmark (17) & \xmark & \xmark & \xmark & \cmark (11) & \xmark & \cmark (5) & \xmark & \cmark (22) & \xmark \\

\bottomrule
\end{tabular}
\end{adjustbox}
\end{table*}

\begin{table*}[t]
\centering
\resizebox{\textwidth}{!}{%
\begin{tabular}{lll|ccccc}
\toprule
\textbf{\thead{Reasoning}} & \textbf{\thead{Memory}} & \textbf{\thead{Reflection}} & \textbf{\thead{Qwen3.5-27B\\Success Rate (SR)}} & \textbf{\thead{Qwen3.5-9B\\Success Rate (SR)}} & \textbf{\thead{Qwen3.5-2B\\Success Rate (SR)}} & \textbf{\thead{Qwen3.5-0.8B\\Success Rate (SR)}} & \textbf{\thead{GPT-5 mini\\Success Rate (SR)}}\\
\midrule
None        & Base       & None        & 0.4 & 0 & 0.1 & 0.1 & 0.3\\
\midrule
ReAct       & Base       & None        & 0.5 & 0.4 & 0.4 & 0.4 & 0.4\\
ReAct       & ChatDB     & None        & 0.6 & 0.5 & 0.2 & 0.1 & 0.2\\
ReAct       & DC         & None        & 0.5 & 0.5 & 0.2 & 0.2 & 0.5\\
ReAct       & SimpleMem  & None        & 0.5 & 0.4 & 0.3 & 0.2 & 0.4\\
ReAct       & MemoryBank & None        & 0.4 & 0.3 & 0.3 & 0.3 & 0.4\\
ReAct       & OpenClaw   & None        & 0.3 & 0.3 & 0.3 & 0.3 & 0.3\\
\midrule
CoT         & Base       & None        & 0.5 & 0.4 & 0.4 & 0.2 & 0.5\\
CoT         & ChatDB     & None        & 0.6 & 0.4 & 0.4 & 0.2 & --\\
CoT         & DC         & None        & 0.6 & 0.4 & -- & 0.3 & --\\
CoT         & SimpleMem  & None        & 0.5 & 0.2 & 0.4 & 0.1 & --\\
CoT         & MemoryBank & None        & 0.2 & 0.2 & 0.4 & 0.2 & --\\
Plan\&Solve & Base      & None        & 0.5 & 0.2 & 0.3 & 0.1 & 0.4\\
Plan\&Solve & ChatDB    & None        & 0.5 & 0.3 & 0 & 0.1 & --\\
Plan\&Solve & DC        & None        & 0.3 & 0.3 & -- & 0 & --\\
Plan\&Solve & SimpleMem & None        & 0.5 & 0.2 & 0.1 & 0 & --\\
Plan\&Solve & MemoryBank & None       & 0.1 & 0.1 & 0.1 & 0.1 & --\\
MAD         & Base       & None        & 0.5 & 0.1 & 0.6 & 0.1 & 0.6\\
MAD         & ChatDB     & None        & 0.4 & 0.3 & 0.4 & 0.1 & --\\
MAD         & DC         & None        & 0.5 & 0.3 & -- & 0.1 & --\\
MAD         & SimpleMem  & None        & 0.4 & 0.3 & 0.4 & 0.1 & --\\
MAD         & MemoryBank & None        & 0.3 & 0.3 & 0.4 & 0.3 & --\\
\midrule
ReAct       & Base       & Self-Refine & 0.5 & 0.6 & -- & 0.1 & 0.4\\
CoT         & Base       & Self-Refine & 0.3 & 0.3 & -- & 0.3 & --\\
CoT         & SimpleMem  & Self-Refine & 0.4 & 0.3 & -- & 0.3 & --\\
Plan\&Solve & Base      & Self-Refine & 0.6 & 0.1 & -- & 0 & --\\
MAD         & MemoryBank & Self-Refine & 0.4 & 0.3 & -- & 0.2 & --\\
\midrule
ReAct       & Base       & Reflexion   & 0.5 & 0.3 & -- & 0.2 & 0.1\\
ReAct       & DC         & Reflexion   & 0.5 & 0.4 & -- & 0.2 & --\\
ReAct       & MemoryBank & Reflexion   & 0.3 & 0.1 & 0.5 & 0.3 & --\\
CoT         & Base       & Reflexion   & 0.4 & 0.6 & -- & 0.2 & --\\
CoT         & SimpleMem  & Reflexion   & 0.5 & 0.4 & -- & 0.4 & --\\
Plan\&Solve & Base      & Reflexion   & 0.5 & 0.2 & -- & 0 & --\\
MAD         & MemoryBank & Reflexion   & 0.6 & 0.3 & -- & 0.2 & --\\
\bottomrule
\end{tabular}%
}
\caption{Complete main results on \textsc{MiniGrid} for Qwen3.5-27B, Qwen3.5-9B, Qwen3.5-2B, and Qwen3.5-0.8B, and GPT-5 mini. All entries report success rate (SR) only. Some method combinations were not evaluated due to time and resource constraints, and their missing results are marked as \texttt{--}.}

\label{tab:minigrid_main_qwen_gpt5mini_multiscale}
\end{table*}

\subsection{Case A}
\vspace{-2mm}
\begin{figure}[h]
    \centering
    \includegraphics[width=0.8\linewidth,page=1]{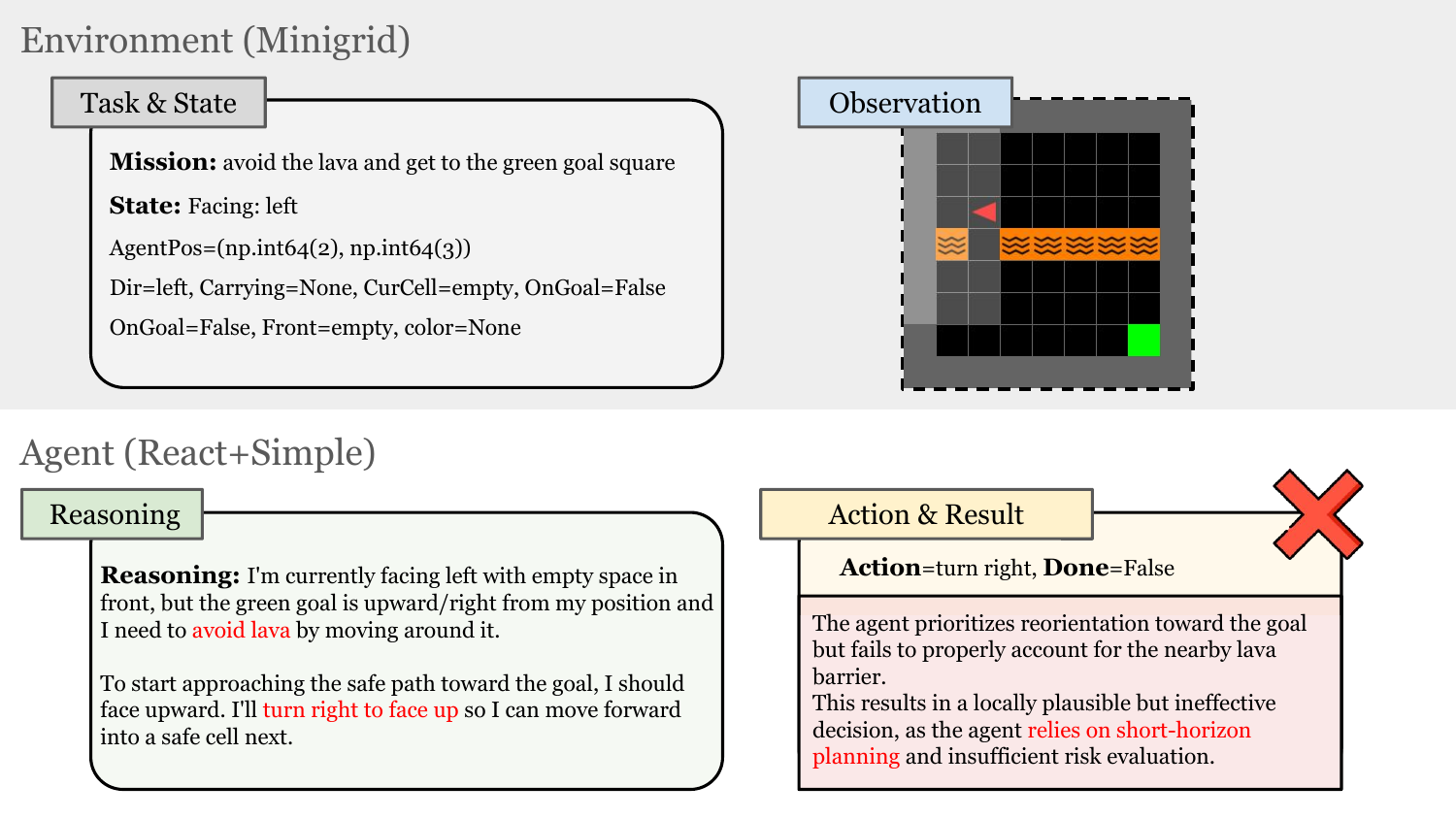}
    \caption{\textbf{Minigrid: ReAct+Simple+None (Failure).}}
    \label{fig:minigrid_react_failure}
\end{figure}

In Figure~\ref{fig:minigrid_react_failure}, the agent correctly infers the goal direction and turns right to align with it, but fails to account for the nearby lava barrier. While the action is locally plausible, it does not lead to a valid path. By prioritizing goal alignment over environmental constraints, the agent produces an ineffective decision, reflecting short-horizon planning and insufficient risk evaluation.

\begin{figure}[h]
    \centering
    \includegraphics[width=0.8\linewidth,page=2]{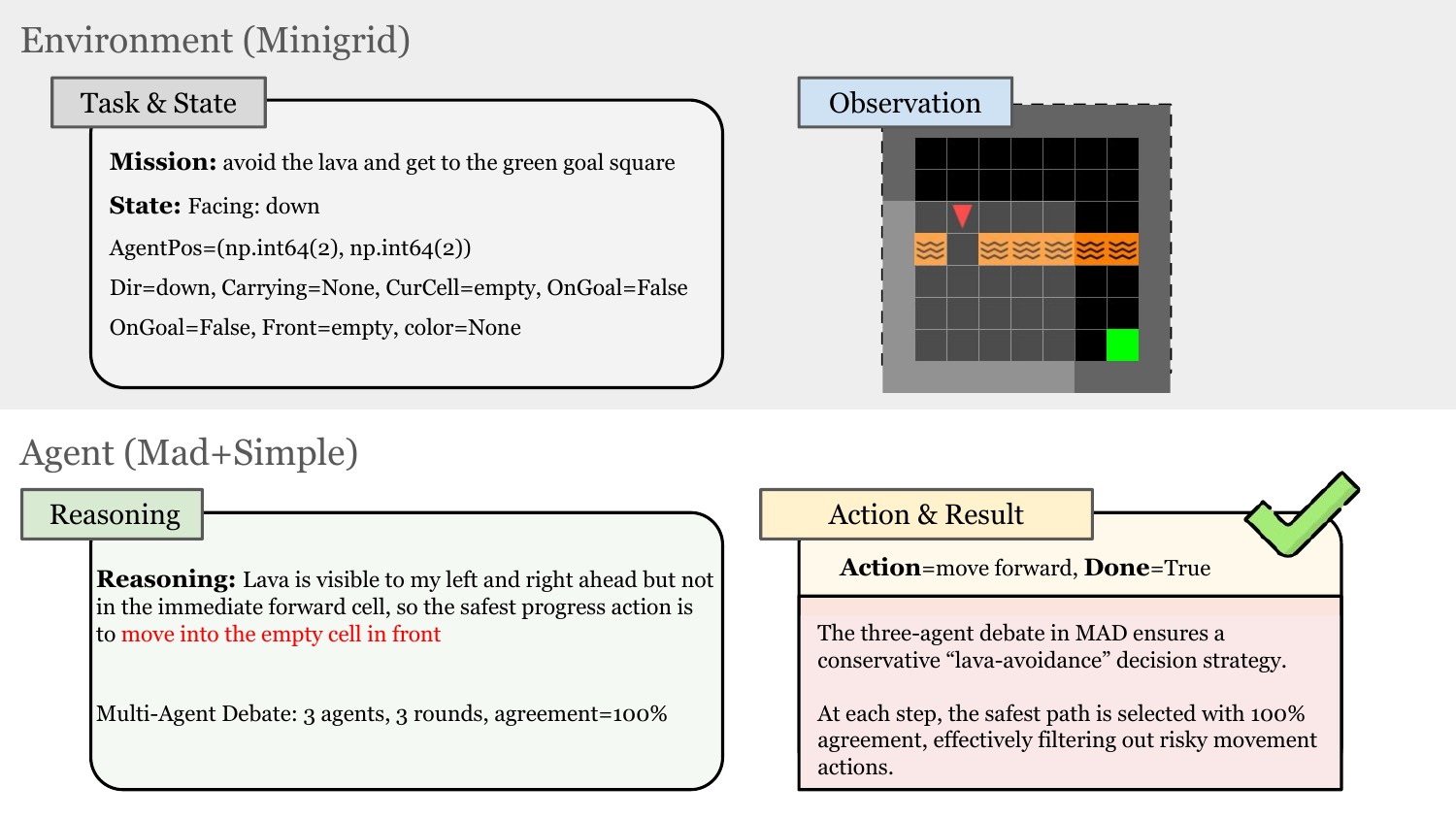}
    \caption{\textbf{Minigrid: MAD+Simple+None (Success).}}
    \label{fig:minigrid_mad_success}
\end{figure}
In Figure~\ref{fig:minigrid_mad_success}, the agents identify a safe passage through the lava and move forward into the empty cell, enabling correct progression. MAD employs a three-agent, three-round debate process that encourages a conservative, lava-avoidance strategy. At each step, the agents converge on a safe action with consistent agreement, effectively filtering out risky movements.

\subsection{Case B}

\begin{figure}[h]
    \centering
    \includegraphics[width=0.8\linewidth,page=4]{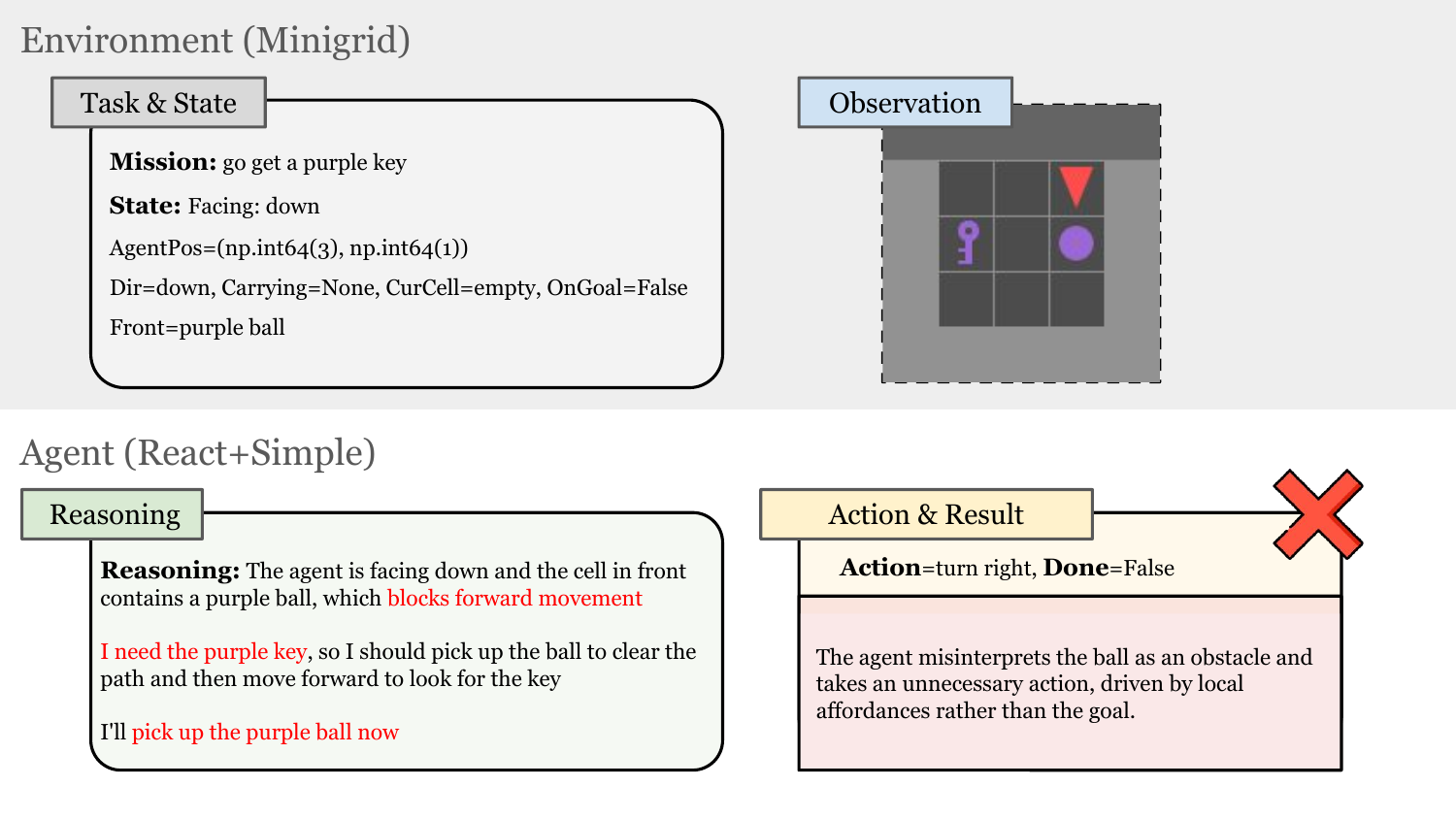}
    \caption{\textbf{Minigrid: React+Simple+None (False).}}
    \label{fig:minigrid_react_simple_false}
\end{figure}

In Figure~\ref{fig:minigrid_react_simple_false}, the agent observes a purple ball directly in front and incorrectly treats it as an obstacle that blocks forward movement. It decides to pick up the ball in order to clear the path, despite the mission being to retrieve the purple key. This action is driven by local affordances rather than task relevance, resulting in a redundant and improper decision.

\begin{figure}[h]
    \centering
    \includegraphics[width=0.8\linewidth,page=3]{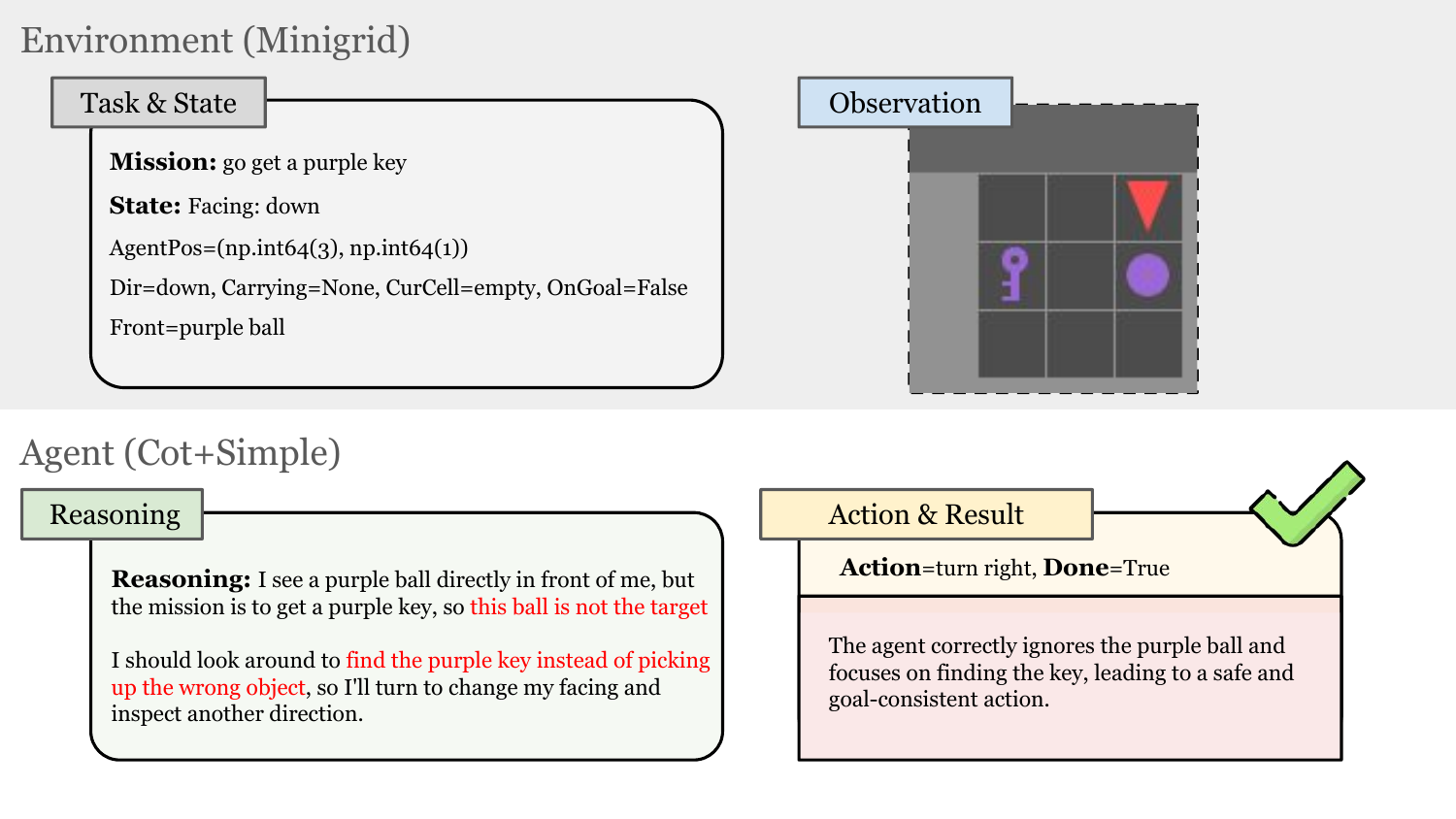}
    \caption{\textbf{Minigrid: Cot+Simple+None (Success).}}
    \label{fig:minigrid_cot_success}
\end{figure}

In Figure~\ref{fig:minigrid_cot_success}, the agent recognizes that the purple ball is not the target object and instead focuses on locating the purple key. It chooses to turn and explore alternative directions rather than interacting with the irrelevant object. This leads to a goal-consistent action, correctly avoiding distraction from non-essential elements in the environment.

\subsection{Case C}
\begin{figure}[h]
    \centering
    \includegraphics[width=0.8\linewidth,page=5]{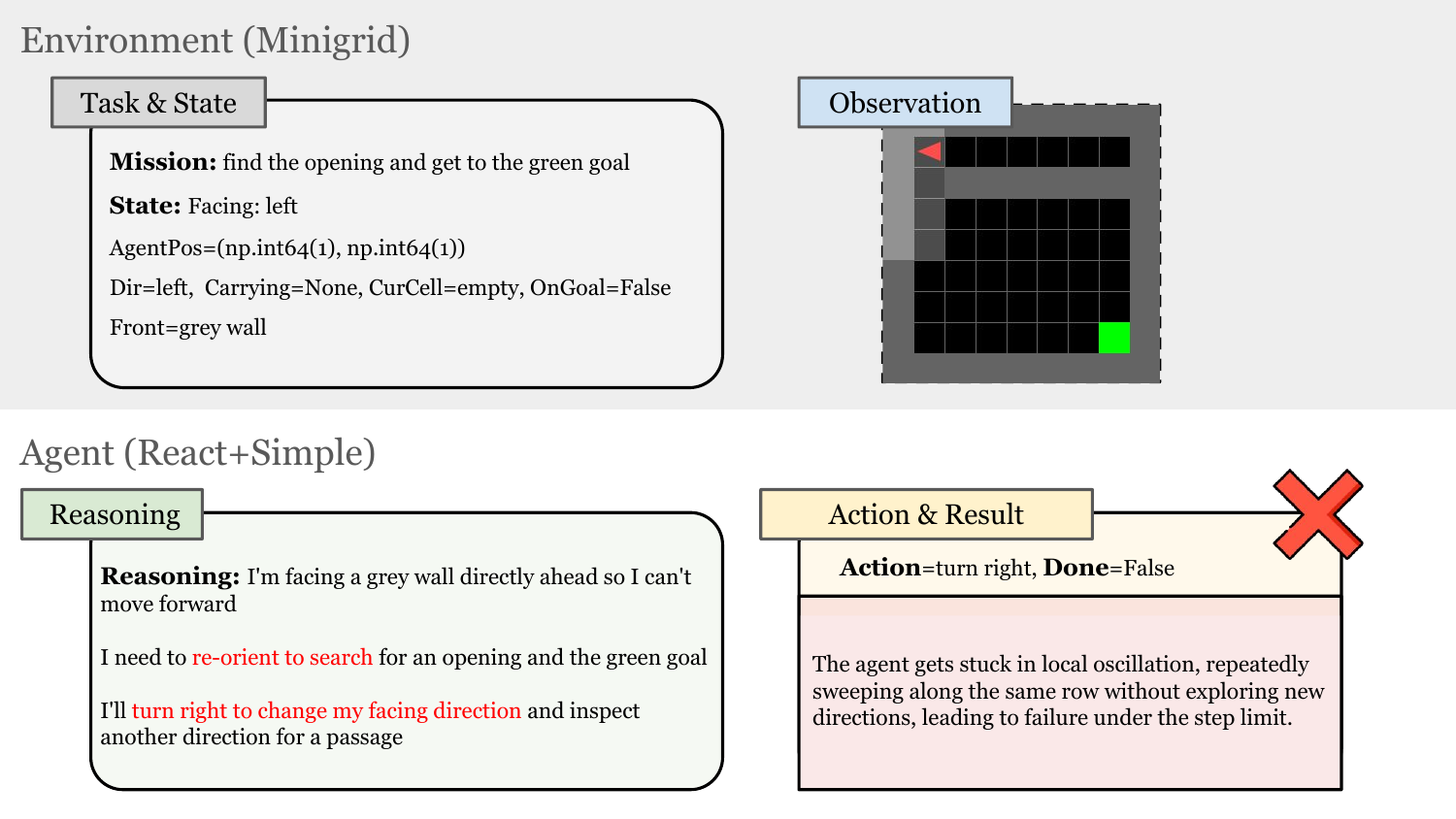}
    \caption{\textbf{Minigrid: React+Simple+None (False).}}
    \label{fig:minigrid_react_simple_false}
\end{figure}

In Figure~\ref{fig:minigrid_react_simple_false}, the agent observes a wall directly ahead and decides to turn right to search for an opening. However, this behavior leads to repeated local exploration along the same row without entering new regions. The agent becomes trapped in a local oscillation pattern, failing to effectively explore alternative directions and ultimately exceeding the step limit.

\begin{figure}[h]
    \centering
    \includegraphics[width=0.8\linewidth,page=6]{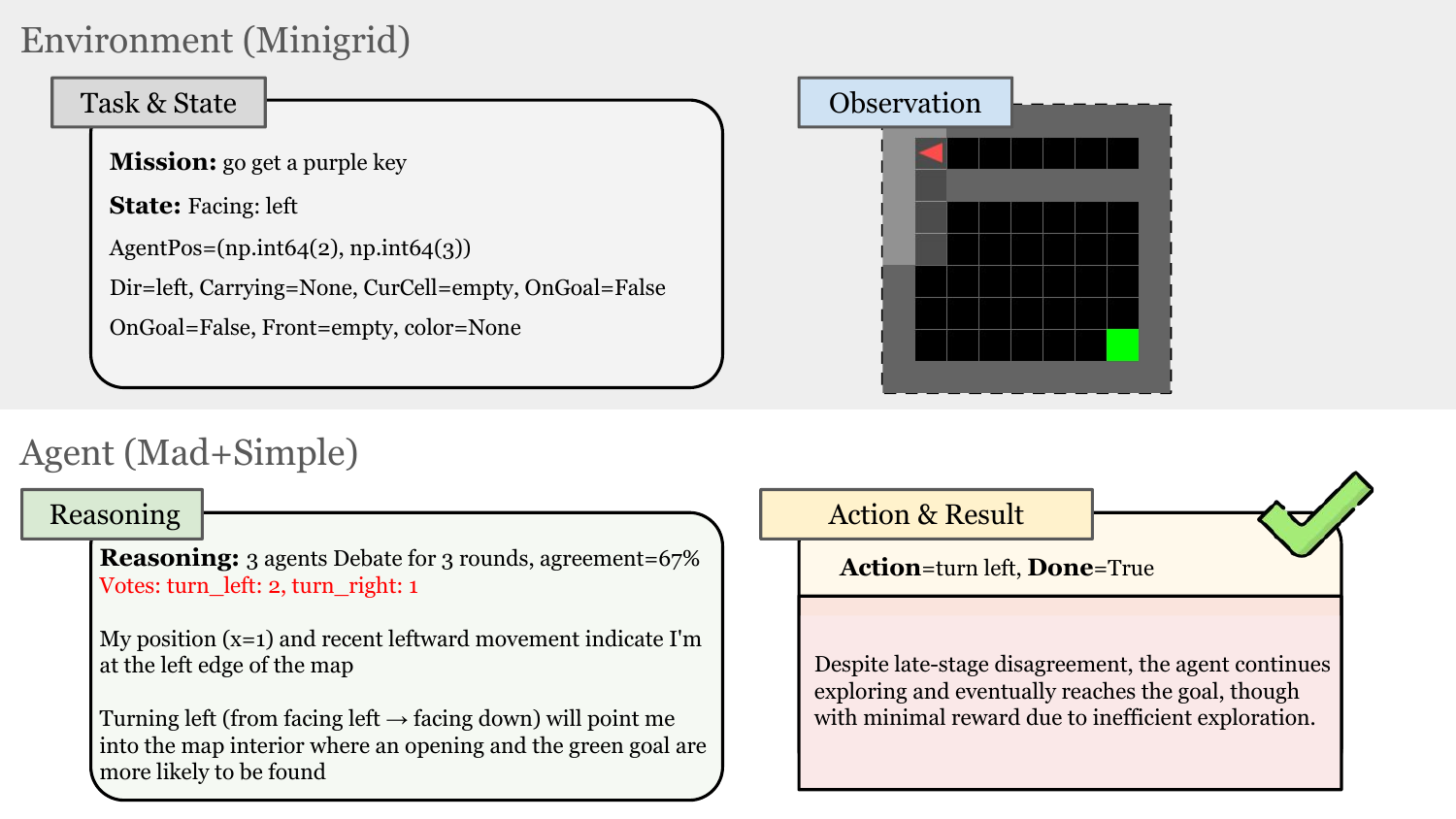}
    \caption{\textbf{Minigrid: Mad+Simple+None (Success).}}
    \label{fig:minigrid_mad_simple_success}
\end{figure}

In Figure~\ref{fig:minigrid_mad_simple_success}, the agents infer that they are near the left boundary and choose to turn left to move toward the map interior. Despite partial disagreement during the debate, the selected action enables continued exploration into new regions. This avoids local oscillation and allows the agent to reach the goal, albeit with some inefficiency eventually.}

{\let\section\subsection
\let\subsection\subsubsection
\section{Alfred}
\label{casestudy:alfred}
\subsection{Complete Experimental Results}
\label{complete_results_alfred}
Table~\ref{tab:alfred_suite_ablation} summarizes the complete results on the ALFRED task suite. We report performance across different combinations of reasoning strategies, memory modules, and reflection settings. 

\begin{table*}[t]
\centering
\scriptsize
\setlength{\tabcolsep}{5pt}
\renewcommand{\arraystretch}{1.15}

\caption{
Ablation results on the ALFRED task suite. We evaluate different combinations of reasoning modules (CoT, Direct, MAD, ReAct, and Plan-and-Solve), memory components (ChatDB, DC, Memory Bank, Simple Memory, or none), and reflection settings with \texttt{gpt-5-mini} as backbone model. Each row corresponds to one method configuration. For readability, method names are written without underscores. Unless a method name explicitly includes \texttt{self-refine}, the reflection module is set to none. A green checkmark indicates successful task completion, while a red cross indicates failure. Values in parentheses denote subgoal completion rates.
}
\label{tab:alfred_suite_ablation}

\begin{adjustbox}{width=\textwidth}
\begin{tabular}{@{}l c c c c c c c@{}}
\toprule
\textbf{Method} &
\makecell{\textbf{1}\\Pick} &
\makecell{\textbf{2}\\Pick2} &
\makecell{\textbf{3}\\Heat} &
\makecell{\textbf{4}\\Cool} &
\makecell{\textbf{5}\\Clean} &
\makecell{\textbf{6}\\Light} &
\makecell{\textbf{7}\\Mobile} \\
\midrule

CoT + ChatDB &
\cmark &
\xmark (0/2) &
\xmark (0/3) &
\cmark &
\xmark (0/3) &
\xmark (0/2) &
\xmark (1/3) \\

CoT + DC &
\cmark &
\xmark (1/2) &
\xmark (1/3) &
\xmark (1/3) &
\xmark (0/3) &
\xmark (0/2) &
\cmark \\


CoT + SimpleMem &
\xmark (0/1) &
\xmark (1/2) &
\xmark (0/3) &
\xmark (1/3) &
\xmark (0/3) &
\xmark (0/2) &
\xmark (0/3) \\

CoT + Simple &
\cmark &
\xmark (0/2) &
\xmark (1/3) &
\xmark (1/3) &
\xmark (0/3) &
\xmark (1/2) &
\xmark (0/3) \\

Direct + Simple &
\xmark (0/1) &
\xmark (0/2) &
\xmark (0/3) &
\xmark (1/3) &
\xmark (0/3) &
\xmark (0/2) &
\xmark (0/3) \\

MAD + Simple &
\xmark (0/1) &
\xmark (0/2) &
\xmark (0/3) &
\xmark (1/3) &
\xmark (1/3) &
\xmark (1/2) &
\xmark (0/3) \\

Plan\&Solve + Simple&
\xmark (0/1) &
\xmark (1/2) &
\xmark (1/3) &
\xmark (0/3) &
\xmark (0/3) &
\xmark (0/2) &
\xmark (0/3) \\

ReAct + ChatDB &
\xmark (0/1) &
\xmark (0/2) &
\cmark &
\cmark &
\xmark (0/3) &
\xmark (1/2) &
\xmark (0/3) \\

ReAct + DC &
\cmark &
\xmark (0/2) &
\xmark (0/3) &
\xmark (1/3) &
\xmark (0/3) &
\xmark (1/2) &
\xmark (0/3) \\

ReAct + Memory Bank &
\cmark &
\xmark (0/2) &
\xmark (1/3) &
\cmark &
\xmark (0/3) &
\xmark (0/2) &
\xmark (0/3) \\


ReAct + SimpleMem &
\cmark &
\xmark (0/2) &
\xmark (0/3) &
\xmark (1/3) &
\xmark (0/3) &
\xmark (0/2) &
\xmark (0/3) \\

ReAct + Simple &
\cmark &
\xmark (1/2) &
\xmark (1/3) &
\xmark (1/3) &
\xmark (0/3) &
\xmark (1/2) &
\xmark (0/3) \\

ReAct + Simple + Self-Refine &
\cmark &
\xmark (1/2) &
\xmark (1/3) &
\xmark (1/3) &
\xmark (0/3) &
\xmark (1/2) &
\xmark (0/3) \\

\midrule
\textbf{Success Rate} &
\textbf{10/17} &
\textbf{0/17} &
\textbf{1/17} &
\textbf{3/17} &
\textbf{0/17} &
\textbf{0/17} &
\textbf{1/17} \\
\bottomrule
\end{tabular}
\end{adjustbox}
\end{table*}

\begin{table*}[t]
\centering
\resizebox{\textwidth}{!}{%
\begin{tabular}{lll|ccccc}
\toprule
\textbf{\thead{Reasoning}} & \textbf{\thead{Memory}} & \textbf{\thead{Reflection}} & \textbf{\thead{Qwen3.5-27B\\Success Rate (SR)}} & \textbf{\thead{Qwen3.5-9B\\Success Rate (SR)}} & \textbf{\thead{Qwen3.5-2B\\Success Rate (SR)}} & \textbf{\thead{Qwen3.5-0.8B\\Success Rate (SR)}} & \textbf{\thead{GPT-5 mini\\Success Rate (SR)}}\\
\midrule
None        & Base       & None        & 0 & 0 & 0 & 0 & 0\\
\midrule
ReAct       & Base       & None        & 0.4 & 0.5 & 0.2 & 0 & 0.4\\
ReAct       & ChatDB     & None        & 0.5 & 0.5 & 0.1 & 0 & 0.4\\
ReAct       & DC         & None        & 0.5 & 0.2 & 0.2 & 0 & 0.4\\
ReAct       & SimpleMem  & None        & 0.6 & 0.4 & 0.1 & 0 & 0.2\\
ReAct       & MemoryBank & None        & 0.3 & 0.3 & 0.2 & 0 & 0.1\\
ReAct       & OpenClaw   & None        & 0.3 & 0.2 & 0.1 & 0 & 0.1\\
\midrule
CoT         & Base       & None        & 0.3 & 0.4 & 0.1 & 0 & 0.2\\
CoT         & ChatDB     & None        & 0.6 & 0.3 & 0.1 & 0 & --\\
CoT         & DC         & None        & 0.4 & 0.4 & 0 & 0 & --\\
CoT         & SimpleMem  & None        & 0.5 & 0.3 & 0.1 & 0 & --\\
CoT         & MemoryBank & None        & 0.4 & 0.3 & 0.1 & 0 & --\\
Plan\&Solve & Base      & None        & 0 & 0.1 & 0 & 0 & 0.2\\
Plan\&Solve & ChatDB    & None        & 0.1 & 0.3 & 0 & 0 & --\\
Plan\&Solve & DC        & None        & 0.1 & 0 & 0 & 0 & --\\
Plan\&Solve & SimpleMem & None        & 0.2 & 0 & 0 & 0 & --\\
Plan\&Solve & MemoryBank & None       & 0.3 & 0 & 0 & 0 & --\\
MAD         & Base       & None        & 0.4 & 0.3 & 0.2 & 0 & 0\\
MAD         & ChatDB     & None        & 0.1 & 0.3 & 0.1 & 0 & --\\
MAD         & DC         & None        & 0.1 & 0.3 & 0.1 & 0.1 & --\\
MAD         & SimpleMem  & None        & 0 & 0 & 0 & 0.1 & --\\
MAD         & MemoryBank & None        & 0 & 0 & 0 & 0 & --\\
\midrule
ReAct       & Base       & Self-Refine & 0.1 & 0.3 & 0.1 & 0 & 0.5\\
CoT         & Base       & Self-Refine & 0.2 & 0.4 & -- & -- & --\\
CoT         & SimpleMem  & Self-Refine & 0.1 & 0.3 & -- & -- & --\\
Plan\&Solve & Base      & Self-Refine & 0.1 & 0.1 & -- & -- & --\\
MAD         & MemoryBank & Self-Refine & 0.1 & 0.4 & -- & -- & --\\
\midrule
ReAct       & Base       & Reflexion   & 0.5 & 0.3 & 0.1 & 0 & 0.3\\
ReAct       & DC         & Reflexion   & 0.2 & 0.3 & 0.2 & 0 & --\\
ReAct       & MemoryBank & Reflexion   & 0.4 & 0.1 & 0 & 0 & --\\
CoT         & Base       & Reflexion   & 0.4 & 0.2 & -- & -- & --\\
CoT         & SimpleMem  & Reflexion   & 0.7 & 0.4 & -- & -- & --\\
Plan\&Solve & Base      & Reflexion   & 0 & 0.1 & -- & -- & --\\
MAD         & MemoryBank & Reflexion   & 0.2 & -- & -- & -- & --\\
\bottomrule
\end{tabular}%
}
\caption{Complete main results on \textsc{ALFRED} for Qwen3.5-27B, Qwen3.5-9B, Qwen3.5-2B, and Qwen3.5-0.8B, and GPT-5 mini. All entries report success rate (SR) only. Some method combinations were not evaluated due to time and resource constraints, and their missing results are marked as \texttt{--}.}

\label{tab:alfred_main_qwen_gpt5mini_multiscale}
\end{table*}

\subsection{Case A}

\begin{figure}[h]
    \centering
    \includegraphics[width=0.8\linewidth,page=9]{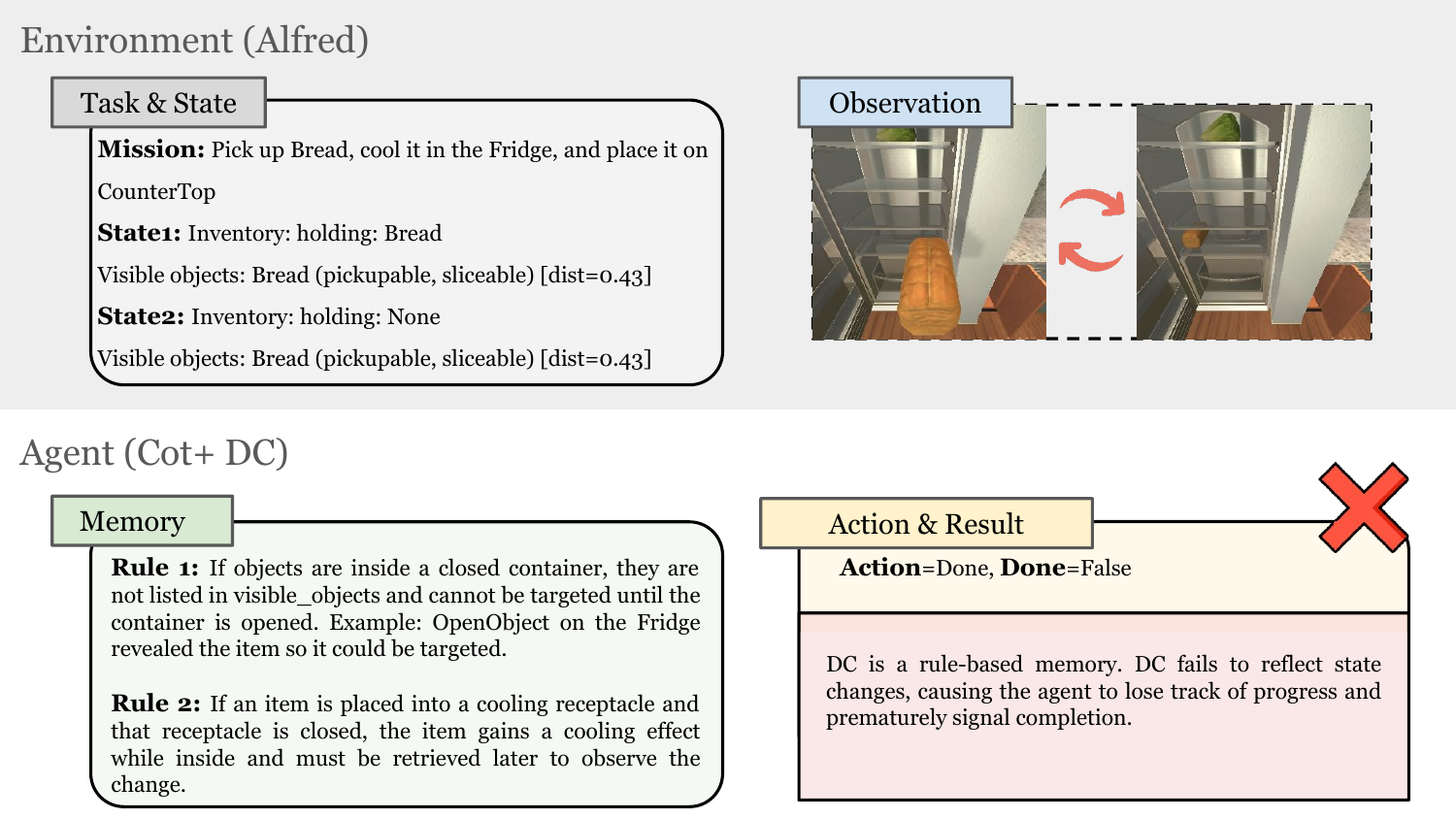}
    \caption{\textbf{Alfred: Cot+DC+None (False).}}
    \label{fig:alfred_react_simple_false}
\end{figure}

In Figure~\ref{fig:alfred_react_simple_false}, the agent relies on rule-based memory but fails to update it according to state changes. After placing the bread into the fridge, it loses track of the object state and incorrectly assumes the task is complete. As a result, the agent prematurely signals completion without retrieving the cooled bread, reflecting a failure to track task progress.

\begin{figure}[h]
    \centering
    \includegraphics[width=0.8\linewidth,page=10]{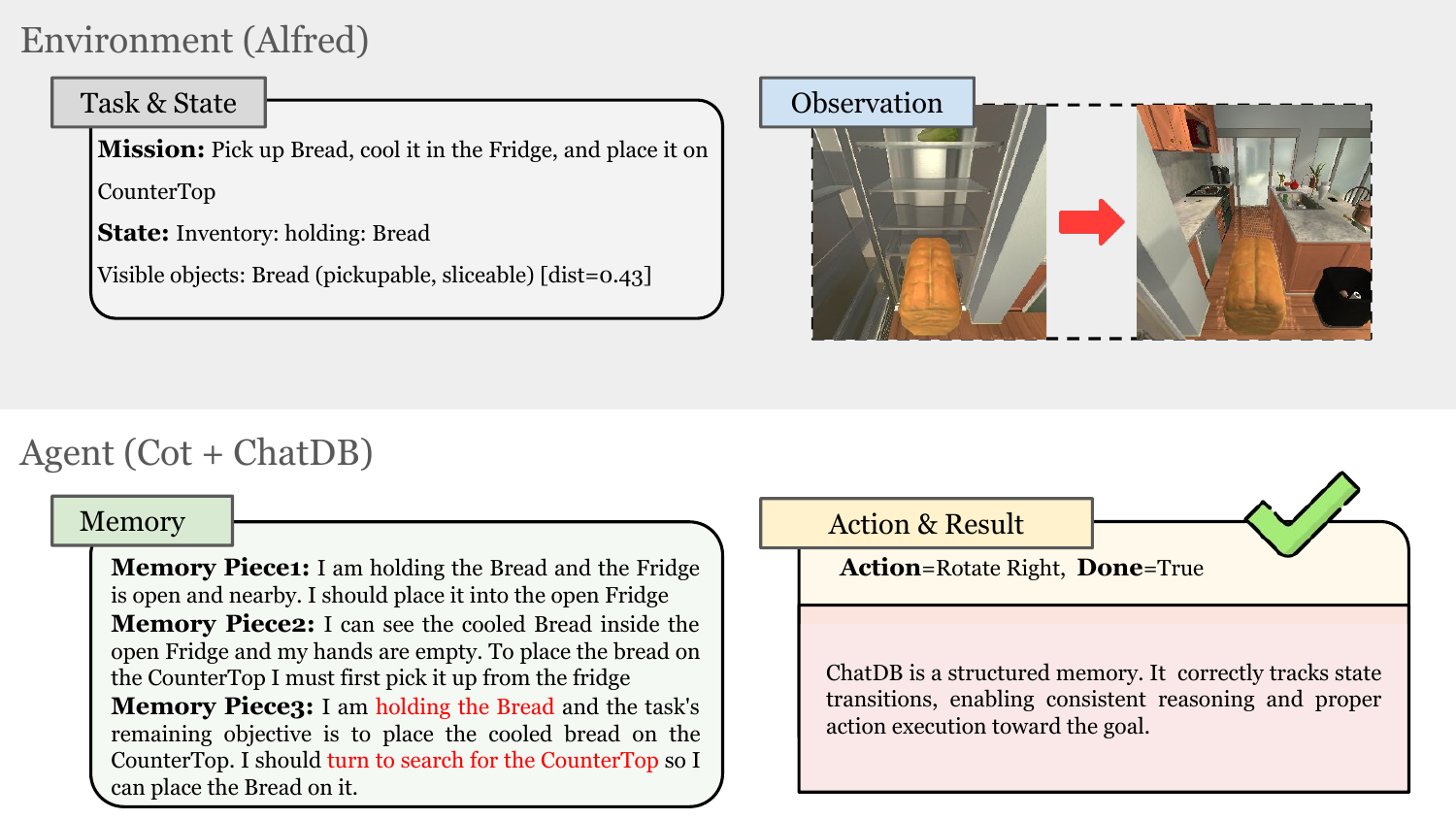}
    \caption{\textbf{Alfred: Cot+ChatDB+None (Success).}}
    \label{fig:alfred_mad_simple_success}
\end{figure}
In Figure~\ref{fig:alfred_mad_simple_success}, the agent maintains structured memory that explicitly records state transitions across steps. It correctly tracks that the bread has been cooled and is currently inside the fridge, and infers the remaining objective. Based on this, it retrieves the bread and proceeds toward the countertop, leading to a consistent and goal-aligned action sequence.

\subsection{Case B}
\begin{figure}[h]
    \centering
    \includegraphics[width=0.8\linewidth,page=11]{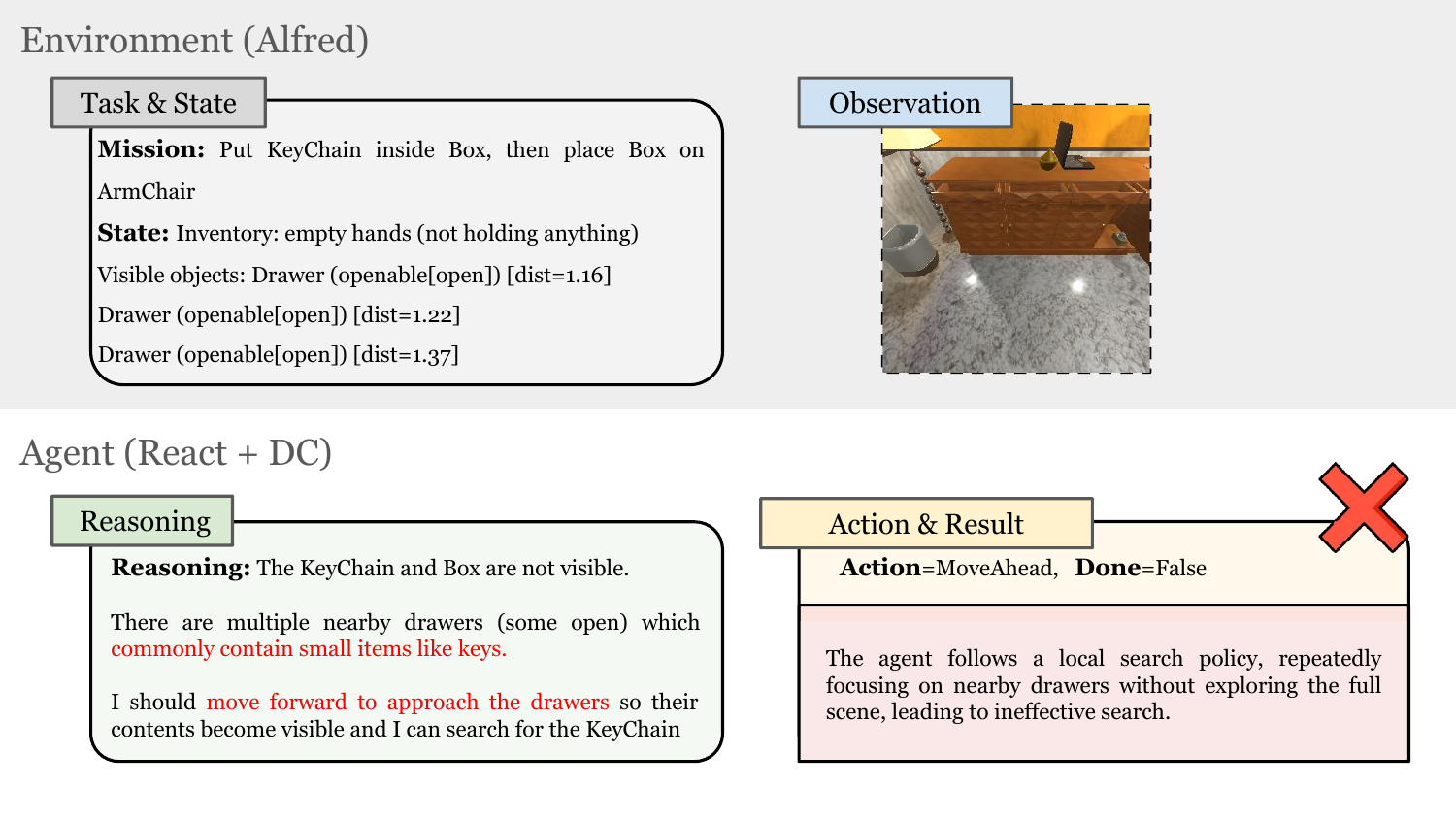}
    \caption{\textbf{Alfred: React+DC+None (False).}}
    \label{fig:alfred_react_dc_false}
\end{figure}

In Figure~\ref{fig:alfred_react_dc_false}, the agent focuses on nearby drawers and decides to move forward to inspect them, assuming they are likely to contain the target object. However, this results in a local search policy that repeatedly prioritizes nearby containers without exploring the broader scene. The agent fails to expand its search space, leading to inefficient exploration and an inability to locate the key object.

\begin{figure}[h]
    \centering
    \includegraphics[width=0.8\linewidth,page=12]{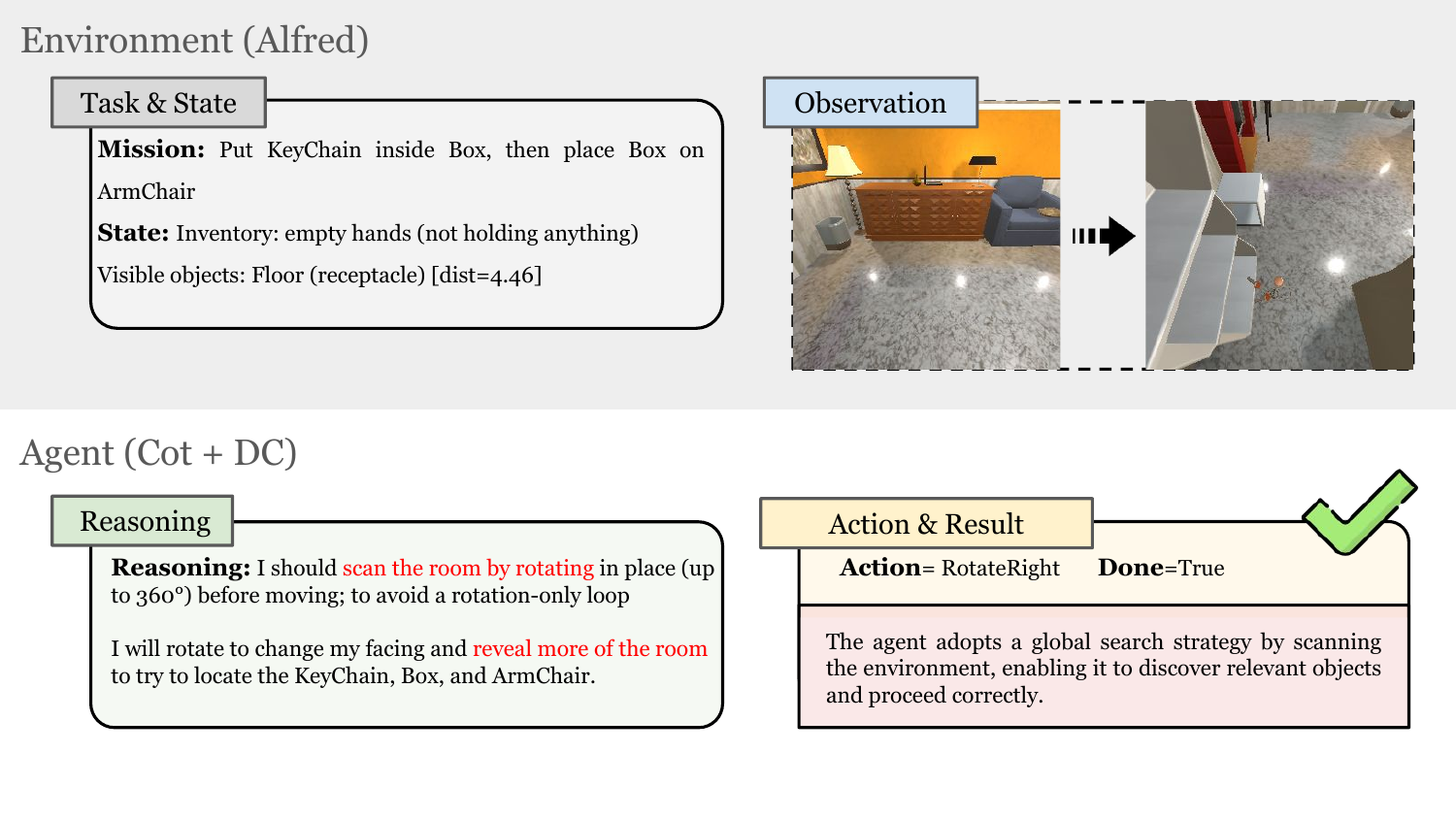}
    \caption{\textbf{Alfred: Cot+DC+None (Success).}}
    \label{fig:alfred_cot_dc_success}
\end{figure}

In Figure~\ref{fig:alfred_cot_dc_success}, the agent explicitly adopts a global search strategy by rotating in place to scan the environment. This allows it to reveal a wider view of the scene and identify relevant objects beyond the immediate vicinity. By expanding its observation space before acting, the agent avoids local search bias and proceeds with a more effective exploration strategy.}

{\let\section\subsection
\let\subsection\subsubsection
\section{RoboTHOR}
\label{casestudy:robothor}

\begin{table*}[t]
\centering
\resizebox{\textwidth}{!}{%
\begin{tabular}{lll|ccccc}
\toprule
\textbf{\thead{Reasoning}} & \textbf{\thead{Memory}} & \textbf{\thead{Reflection}} & \textbf{\thead{Qwen3.5-27B\\Success Rate (SR)}} & \textbf{\thead{Qwen3.5-9B\\Success Rate (SR)}} & \textbf{\thead{Qwen3.5-2B\\Success Rate (SR)}} & \textbf{\thead{Qwen3.5-0.8B\\Success Rate (SR)}} & \textbf{\thead{GPT-5 mini\\Success Rate (SR)}}\\
\midrule
None        & Base       & None        & 0.1 & 0 & 0 & 0 & 0.1\\
\midrule
ReAct       & Base       & None        & 0.1 & 0 & 0 & 0 & 0.2\\
ReAct       & ChatDB     & None        & 0.2 & 0 & 0 & 0.1 & 0.2\\
ReAct       & DC         & None        & 0.1 & 0.1 & 0 & 0.1 & 0.1\\
ReAct       & SimpleMem  & None        & 0.1 & 0.1 & 0 & 0 & 0.1\\
ReAct       & MemoryBank & None        & 0.2 & 0.2 & 0.1 & 0 & 0.3\\
ReAct       & OpenClaw   & None        & 0.2 & 0.2 & 0 & 0 & 0.1\\
\midrule
CoT         & Base       & None        & 0 & 0 & 0 & 0 & 0\\
CoT         & ChatDB     & None        & 0.2 & 0 & -- & 0.1 & --\\
CoT         & DC         & None        & 0 & 0 & -- & 0.1 & --\\
CoT         & SimpleMem  & None        & 0.1 & 0.1 & 0 & 0.1 & --\\
CoT         & MemoryBank & None        & 0.2 & 0.1 & 0.1 & 0.1 & --\\
Plan\&Solve & Base      & None        & 0.1 & 0 & 0 & 0 & 0.1\\
Plan\&Solve & ChatDB    & None        & 0 & 0.2 & -- & 0 & --\\
Plan\&Solve & DC        & None        & 0 & 0.1 & -- & 0.1 & --\\
Plan\&Solve & SimpleMem & None        & 0 & 0.2 & 0 & 0.1 & --\\
Plan\&Solve & MemoryBank & None       & 0.2 & 0.2 & 0 & 0 & --\\
MAD         & Base       & None        & 0.2 & 0.1 & 0 & 0 & 0.1\\
MAD         & ChatDB     & None        & 0.2 & 0.2 & -- & 0 & --\\
MAD         & DC         & None        & 0 & 0 & -- & 0.1 & --\\
MAD         & SimpleMem  & None        & 0 & 0.1 & -- & 0.2 & --\\
MAD         & MemoryBank & None        & 0.1 & -- & -- & 0.1 & --\\
\midrule
ReAct       & Base       & Self-Refine & 0.1 & 0.1 & 0 & 0.1 & 0.2\\
CoT         & Base       & Self-Refine & 0.1 & 0 & -- & 0 & --\\
CoT         & SimpleMem  & Self-Refine & 0 & 0.1 & -- & 0.1 & --\\
Plan\&Solve & Base      & Self-Refine & 0.1 & 0.1 & -- & 0 & --\\
MAD         & MemoryBank & Self-Refine & 0.1 & 0.1 & -- & 0.1 & --\\
\midrule
ReAct       & Base       & Reflexion   & 0.1 & 0.2 & 0.1 & 0 & 0.4\\
ReAct       & DC         & Reflexion   & 0.1 & 0.2 & 0 & 0.1 & --\\
ReAct       & MemoryBank & Reflexion   & 0.2 & 0.3 & 0.1 & 0 & --\\
CoT         & Base       & Reflexion   & 0.1 & 0.1 & -- & 0.1 & --\\
CoT         & SimpleMem  & Reflexion   & 0 & 0.1 & -- & 0 & --\\
Plan\&Solve & Base      & Reflexion   & 0 & 0.1 & -- & 0 & --\\
MAD         & MemoryBank & Reflexion   & 0.2 & 0.1 & -- & 0.1 & --\\
\bottomrule
\end{tabular}%
}
\caption{Complete main results on \textsc{RoboTHOR}. All entries report success rate (SR) only. Some method combinations were not evaluated due to time and resource constraints, and their missing results are marked as \texttt{--}.}

\label{tab:robothor_main_qwen_gpt5mini_multiscale}
\end{table*}

\subsection{Case 1: MemoryBank helps breaking the loop}
\label{casestudy:mb}
\paragraph{Memory}
\textit{Failure Episode (Television Task):} Episode 2, Steps 1--5. Agent position: $x=2.50, y=0.90, z=-1.50$ (static). Target: Television (success $\leq 1.50$m).
\begin{itemize}
  \item Step 1: MoveAhead succeeded (agent pose moved from $x=3.00$ to $x=2.50$).
  \item Steps 2--5: Four consecutive RotateRight actions. Agent heading changes: yaw $270^{\circ} \to 315^{\circ} \to 0^{\circ} \to 44.6^{\circ}$.
  \item Reasoning trace: Steps 2--5 all report ``I don't see television in current view,'' followed by ``I'll rotate right to scan a new direction.'' No MoveAhead follows any rotation; no new spatial exploration occurs.
\end{itemize}
\textit{Success Episode (Basketball Task):} Episode 1, Steps 186--187. Agent position: $x=8.00, y=0.90, z=-2.50$. Target: BasketBall.
\begin{itemize}
  \item Step 186: MoveAhead succeeded. Agent moves within goal threshold ($\leq 1.50$m).
  \item Step 187: Stop executed. Episode terminated with success=True.
  \item Environment diagnostics: front\_goal\_no\_forward=0, front\_key\_no\_pick\_up=0, front\_door\_no\_toggle=0 (path fully clear).
\end{itemize}
\paragraph{insight} The root cause is \textit{rotation-exploration conflation}: the agent treats heading changes (yaw-only) identically to spatial moves ($(x,z)$ changes), ignoring that rotations at static position yield no new observables. The television task exemplifies this---four consecutive rotations at position $(2.50, 0.90, -1.50)$ generate identical diagnostics, yet the agent re-evaluates each as information-generating. 

Three correctives: (1) \textbf{No rotation without spatial progress}: if position unchanged and target unseen, execute MoveAhead instead of rotating again. (2) \textbf{Headed commitment}: commit to one orientation, then execute 2--3 MoveAhead before re-scanning. (3) \textbf{Distance trigger}: if distance $\leq$ goal\_radius, execute Stop immediately. These mechanisms prevent the rotate-in-place dead-loop and enforce the spatial-movement discipline demonstrated in the successful basketball episode.}

\section{Web Interface for Interactive Module Configuration}
\label{app:webui}

To improve usability and lower the barrier to experimentation, \ours provides a lightweight web-based interface for interactively configuring agent pipelines. The interface exposes the major components of the framework, including the environment, perception adapter, reasoning method, memory backend, and reflection module. Users can choose different implementations for each component through dropdown menus and directly generate runnable Python setup code from the selected configuration.

\begin{figure}[t]
    \centering
    \begin{minipage}[t]{0.49\textwidth}
        \centering
        \includegraphics[width=\textwidth]{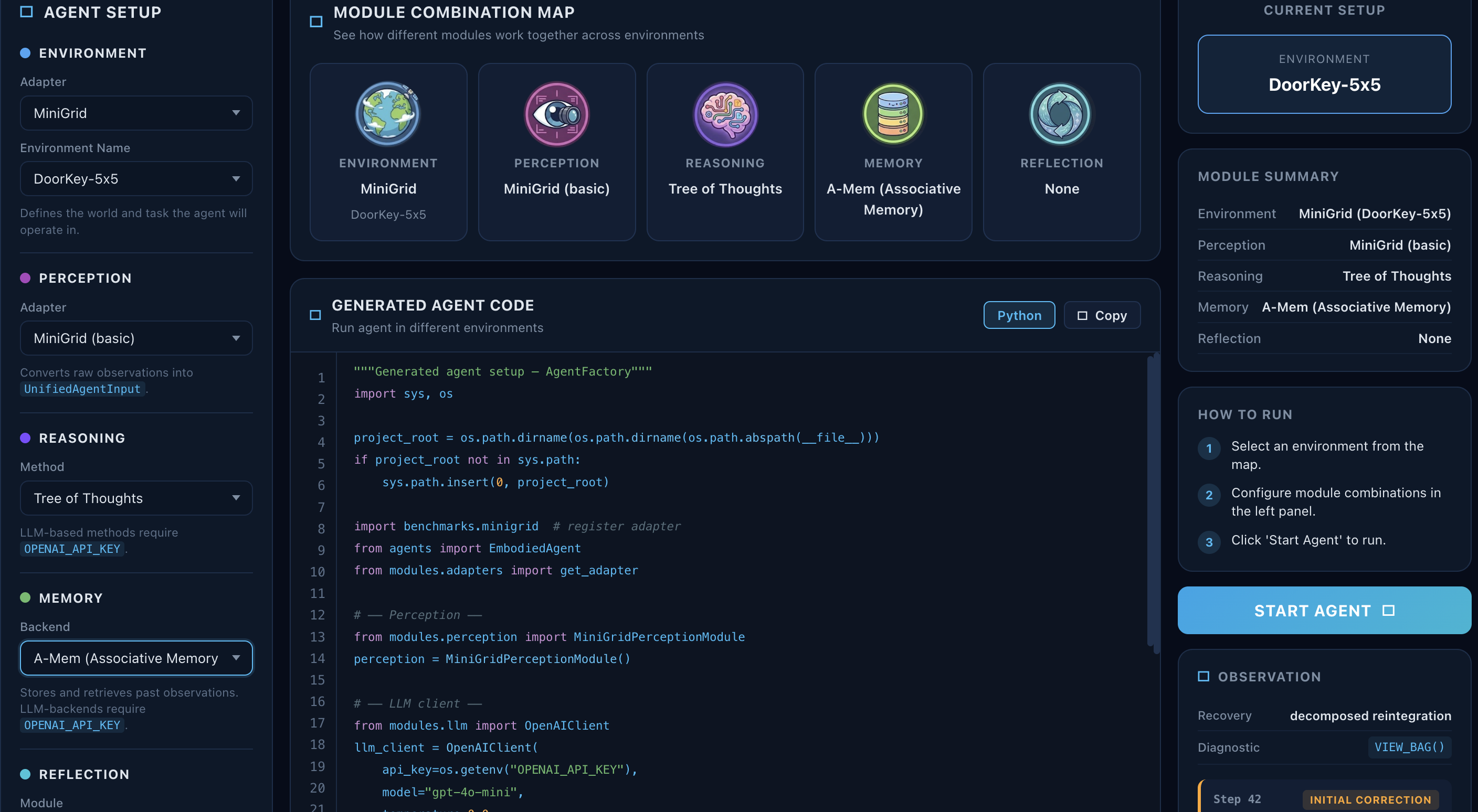}

    \end{minipage}
    \hfill
    \begin{minipage}[t]{0.49\textwidth}
        \centering
        \includegraphics[width=\textwidth]{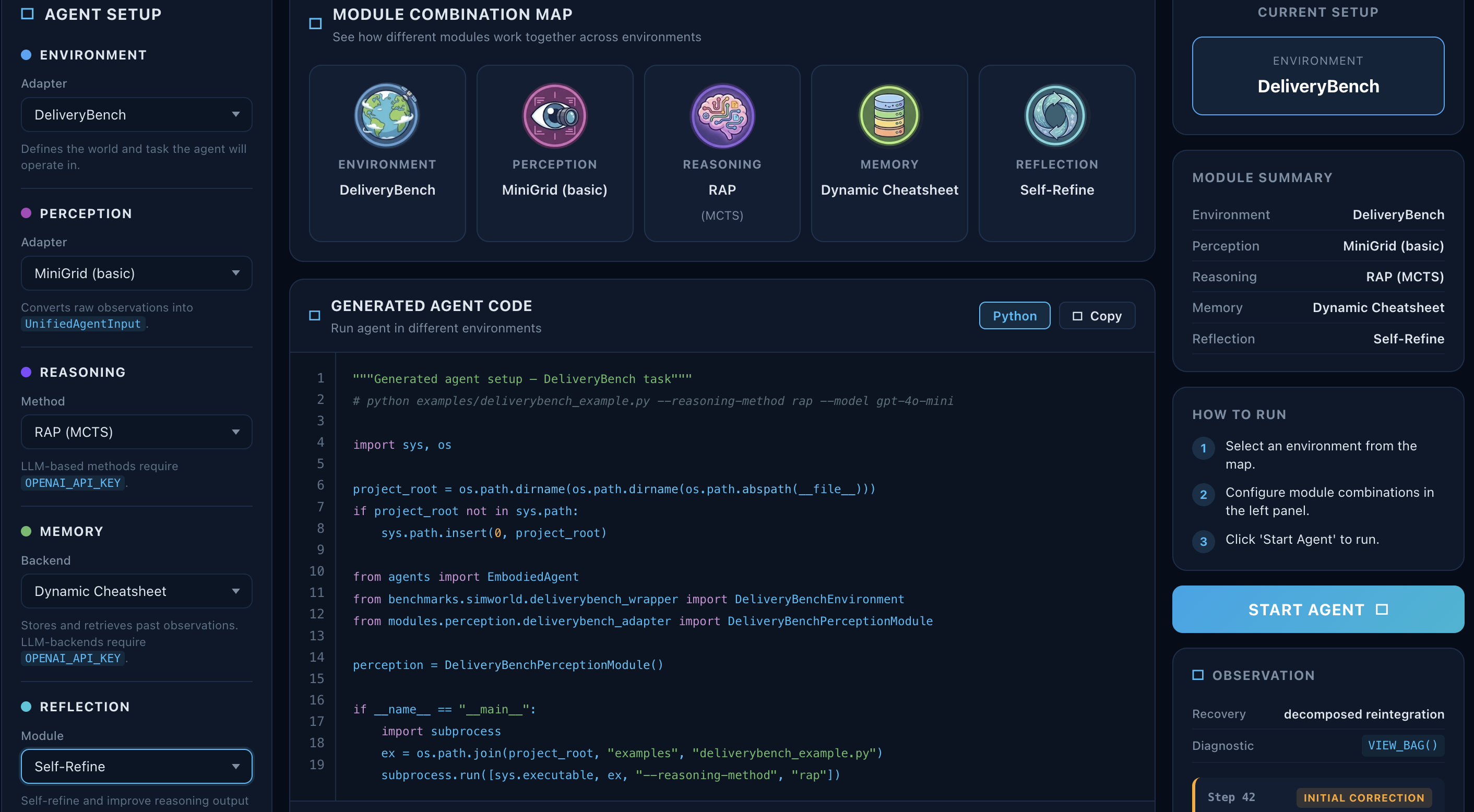}

    \end{minipage}
    \caption{Web interface of \ours{} for interactive module configuration. Users can select implementations for environment, perception, reasoning, memory, and reflection, and then generate runnable Python setup code.}
    \label{fig:webui_examples}
\end{figure}

Figure~\ref{fig:webui_examples} shows two example configurations of the interface. The first example illustrates a MiniGrid setup with Tree of Thoughts reasoning, A-MEM (Associative memory), and Self-Refine reflection. The second example shows a DeliveryBench setup with RAP(Reasoning via Planning) reasoning, Dynamic Cheatsheet (DC) Memory, and Self-Refine reflection. These examples highlight the modular design of \ours: the same interface can be used to compose different agent systems across environments and experimental settings without modifying the underlying framework code manually.

This interface is intended as a practical entry point for both research and prototyping. It allows users to quickly instantiate different combinations of modules, inspect how design choices map to executable configurations, and reproduce experimental settings more conveniently.

\end{document}